\documentclass[letterpaper]{article} 
\usepackage{aaai25}  
\usepackage{times}  
\usepackage{helvet}  
\usepackage{courier}  
\usepackage[hyphens]{url}  
\usepackage{graphicx} 
\usepackage{natbib}  
\usepackage{caption} 
\usepackage{subcaption}
\usepackage{algorithm}
\usepackage{algorithmic}
\usepackage{comment}

\usepackage{newfloat}
\usepackage{listings}
\DeclareCaptionStyle{ruled}{labelfont=normalfont,labelsep=colon,strut=off} 
\floatstyle{ruled}
\newfloat{listing}{tb}{lst}{}
\floatname{listing}{Listing}
\title{Norm Growth and Stability Challenges in Localized \\Sequential Knowledge Editing}
\author{
    Akshat Gupta\textsuperscript{\rm 1}\footnote{Correspondence to: akshat.gupta@berkeley.edu}, Christine Fang\textsuperscript{\rm 1}, Atahan Ozdemir\textsuperscript{\rm 1}, Maochuan Lu\textsuperscript{\rm 1}, \\Ahmed Alaa\textsuperscript{\rm 1}, Thomas Hartvigsen\textsuperscript{\rm 2}, Gopala Anumanchipalli\textsuperscript{\rm 1}
}
\affiliations{
    \textsuperscript{\rm 1}University of California Berkeley, \textsuperscript{\rm 2}University of Virginia\\
}

\usepackage{bibentry}

\begin{document}

\maketitle

\begin{abstract}
This study investigates the impact of localized updates to large language models (LLMs), specifically in the context of knowledge editing - a task aimed at incorporating or modifying specific facts without altering broader model capabilities. We first show that across different post-training interventions like continuous pre-training, full fine-tuning and LORA-based fine-tuning, the Frobenius norm of the updated matrices always increases. This increasing norm is especially detrimental for localized knowledge editing, where only a subset of matrices are updated in a model . We reveal a consistent phenomenon across various editing techniques, including fine-tuning, hypernetwork-based approaches, and locate-and-edit methods: the norm of the updated matrix invariably increases with successive updates. Such growth disrupts model balance, particularly when isolated matrices are updated while the rest of the model remains static, leading to potential instability and degradation of downstream performance. Upon deeper investigations of the intermediate activation vectors, we find that the norm of internal activations decreases and is accompanied by shifts in the subspaces occupied by these activations, which shows that these activation vectors now occupy completely different regions in the representation space compared to the unedited model. With our paper, we highlight the technical challenges with continuous and localized sequential knowledge editing and their implications for maintaining model stability and utility.

\end{abstract}

%

\section{Introduction}
Recent advances in model interpretability have led to methods that perform localized updates to large language models by intervening at very specific locations \cite{ROME, MEMIT, hernandez2023inspecting}. A popular domain in which such methods are employed is knowledge editing \cite{editing-survey} - a task where singular facts are added or updated inside of a model in a data and compute efficient manner. This paper starts with a simple yet powerful observation that sequential knowledge updates made to a model always leads to an increase in the norm of the matrix being updated. We first ask the question - is the increase in norm with continous updates specific to localized knowledge editing methods or is this a general phenomenon? 

Our experiments with numerous post-training interventions, including continual pretraining, full fine-tuning and LORA-based fine-tuning, present a very surprising result - \textbf{the norm of the weight matrices being updated always increases for all these post-training interventions}. While there has been dispersed work showing norm growth \cite{norm-growth}, to the best of our knowledge, our study is the first to emprically evaluate this comprehensively for large number of important post-training interventions.

\begin{figure*}[h]
    \centering
    \begin{subfigure}{0.22\textwidth}
        \includegraphics[width=\linewidth]{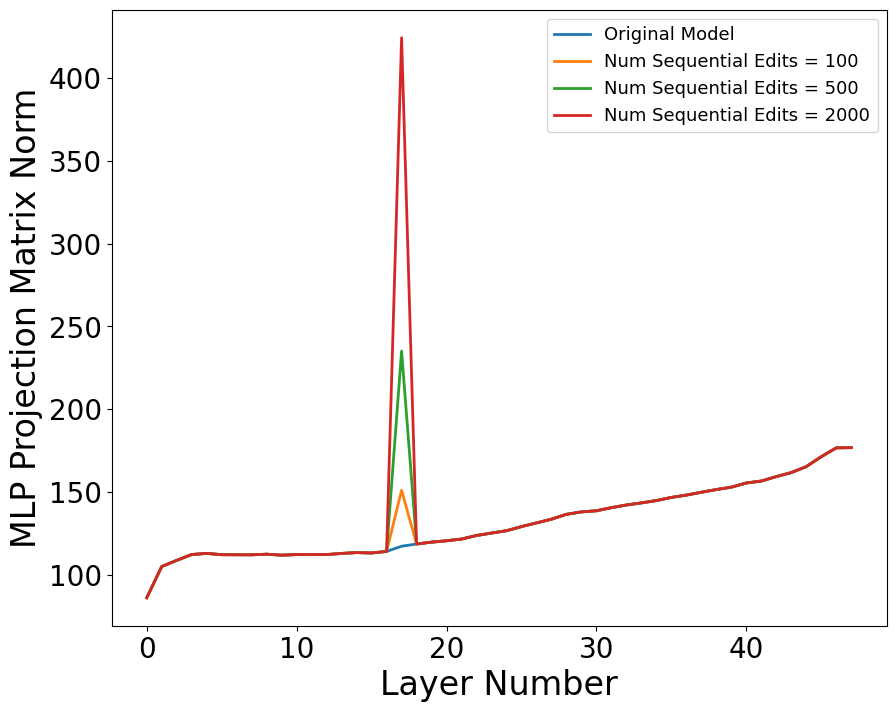}
        \caption{ROME}
    \end{subfigure}
    \begin{subfigure}{0.22\textwidth}
        \includegraphics[width=\linewidth]{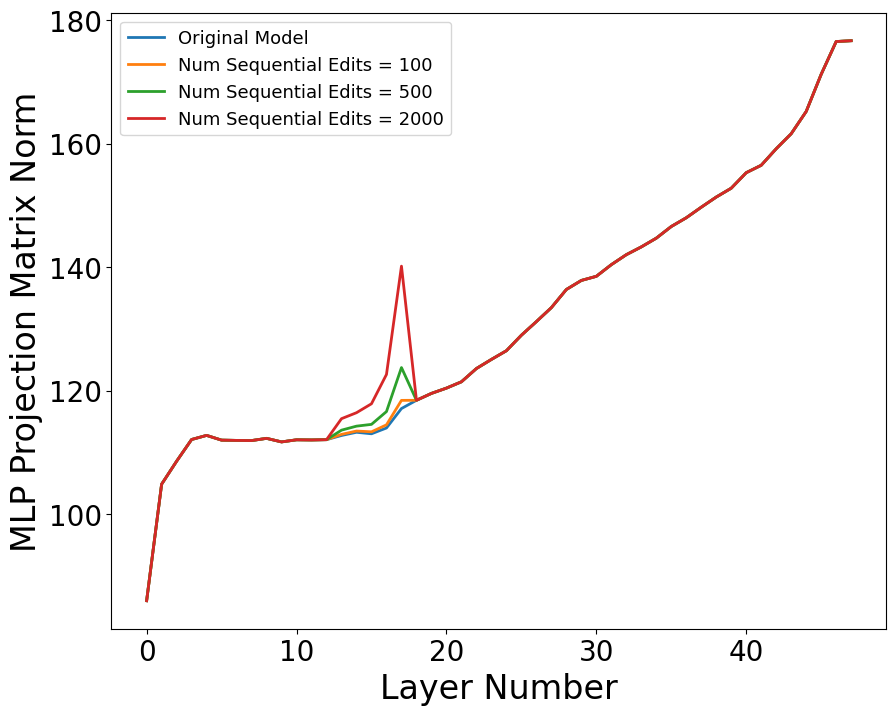}
        \caption{MEMIT}
    \end{subfigure}
    \begin{subfigure}{0.22\textwidth}
        \includegraphics[width=\linewidth]{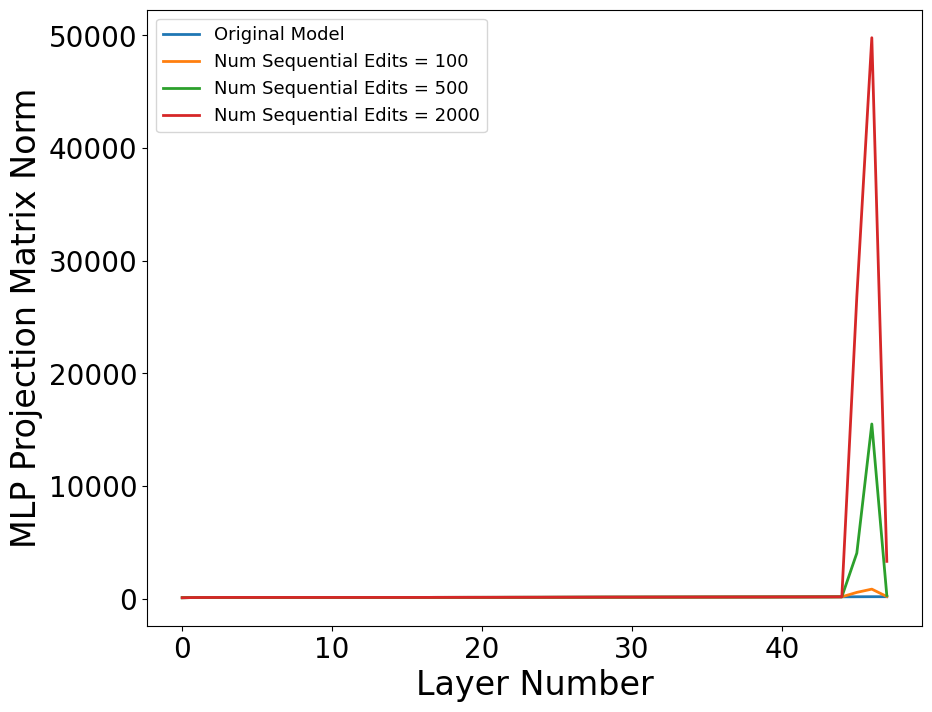}
        \caption{MEND}
    \end{subfigure}
        \begin{subfigure}{0.22\textwidth}
        \includegraphics[width=\linewidth]{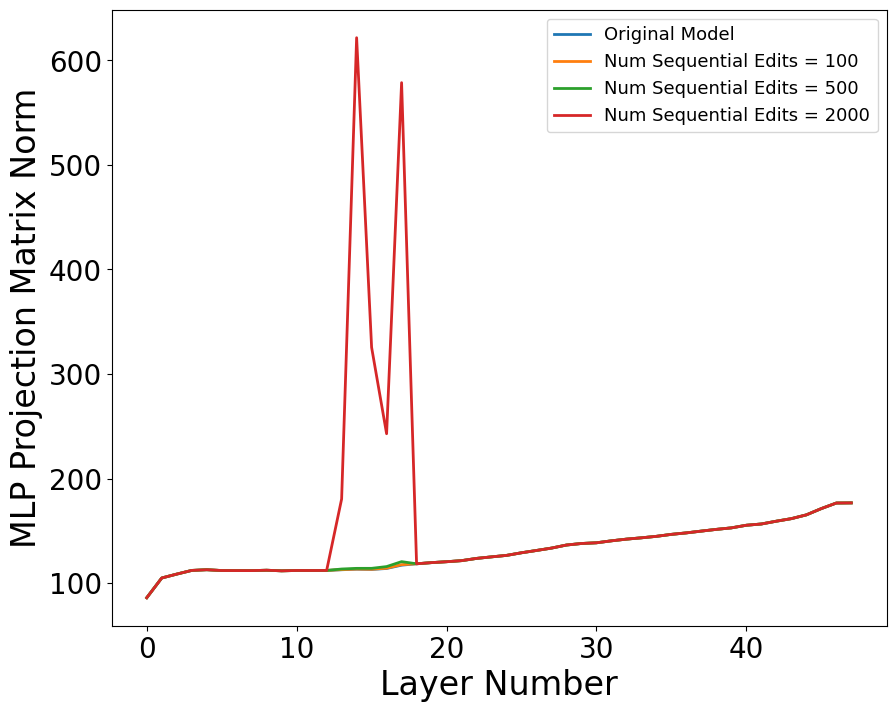}
        \caption{PMET}
    \end{subfigure}


    \caption{Norm growth for edits 100, 500, 2000 for GPT2-XL. }
    \label{fig:norm-growth}
\end{figure*}

We next study the presence and implications of this phenomenon when performing localized updates. To do so we study the task of knowledge editing where localized updates are very common. Knowledge editing methods usually update specific parts of the model, for example the MLP sub-modules of certain layers, to add or update new information. This allows data and compute efficient updates to be made to a model. We perform continuous sequential knowledge edits to a model using various parameter modifying knowledge editing methods along with localized fine-tuning, and find that for all these scenarios, the norm of the edited matrix always increases with the number of updates. While the increasing norm may not be concering in general, it is especially detrimental for performing localized updates. This is because disproportionate and contionuous growth in the norm of one or few layers of a model, while the rest of the model remains frozen, will compromise the balance and stability of the entire system, eventually leading to a breaking point as observed in prior work \cite{akshat-catastrophic, akshat-rebuilding}. This disprortionate growth is shown in Figure \ref{fig:norm-growth}. 

We further analyze the effects of performing continous localized knowledge updates to a model by studying how the hidden activations of the model changes. We find that contrary to the increasing norm of edited matrix, the norm of the activation vectors generated after the edited layers continuously decreases. We also show that these activation begin to occupy different regions in space when compared to the original. A follow-up to our work by \citet{encore} resolve the problems of disproportionate norm growth by using regularization methods and propose a more robust knowledge editing method called ENCORE. 


To summarize, we make the following contributions in this paper:
\begin{enumerate}
    \item We show that the frobenius norm of the updated weight matrices always increases during post training interventions.
    \item The norm of edited matrix increases disproportionately for localized knowledge updates, leading to model collapse. 
    \item This collapse is accompanied by a change in the norm and orientations of the resultant hidden activations, showing that the activations of the edited models now lie in a different region of representation space.
\end{enumerate}

\section{Norm Growth During Post-Training Interventions}
We focus on the following common interventions that are applied on a model after the pretraining step - continual pretraining, full fine-tuning, LORA based fine-tuning \cite{hu2021lora}. We discuss our experiment settings for each of the following below:

\begin{enumerate}
    \item \textbf{Continued pretraining (CPT)} - We consider continued pretraining as as separate case from full fine-tuning although in both cases all the weights of the model are updated using a next-token prediction loss. We define continued pretraining as a process where the foundational knowledge of a model is extended by training on a large domain-specific corpora. In our experiments, we present the results for performing CPT on 20 billion tokens for Python programming \cite{li2023starcoder} on Llama-2 (7B) \cite{llama2}.
    
    \item \textbf{Full Fine-Tuning (FFT)} - We define full fine-tuning as task specific next-token prediction training of a model to optimize the model's parameters on a particular task. We present the results for fine-tuning Llama-2 (7B) model on 110k question answer pairs for programming \cite{wei2024magicoder}. 

    \item \textbf{LORA based full fine-tuning (LFFT)} - Here we use LORA \cite{hu2021lora} to fine-tune all the model weights in the same setting as FFT. 


\end{enumerate}

\begin{figure*}[h]
    \centering
    \begin{subfigure}{0.22\textwidth}
        \includegraphics[width=\linewidth]{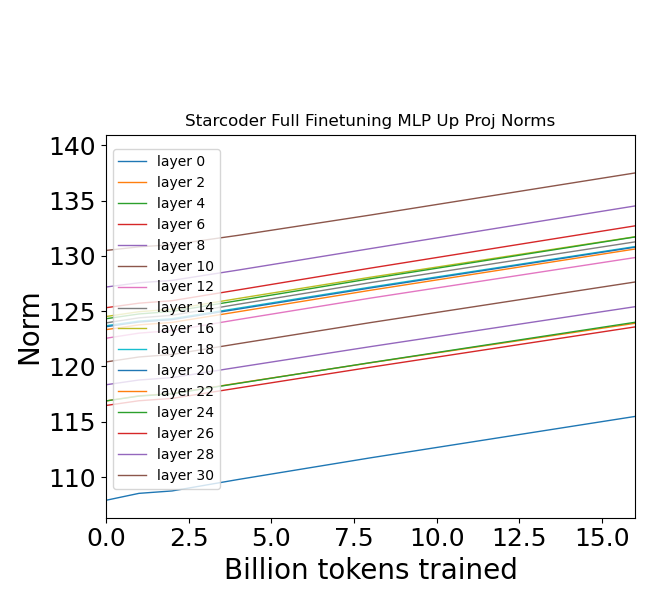}
        \caption{CPT MLP-Up}
    \end{subfigure}
    \begin{subfigure}{0.22\textwidth}
        \includegraphics[width=\linewidth]{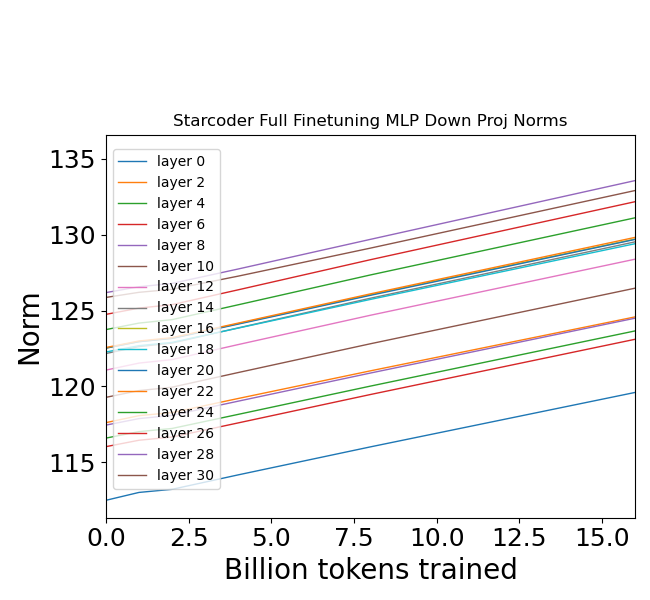}
        \caption{CPT MLP-Down}
    \end{subfigure}
    \begin{subfigure}{0.22\textwidth}
        \includegraphics[width=\linewidth]{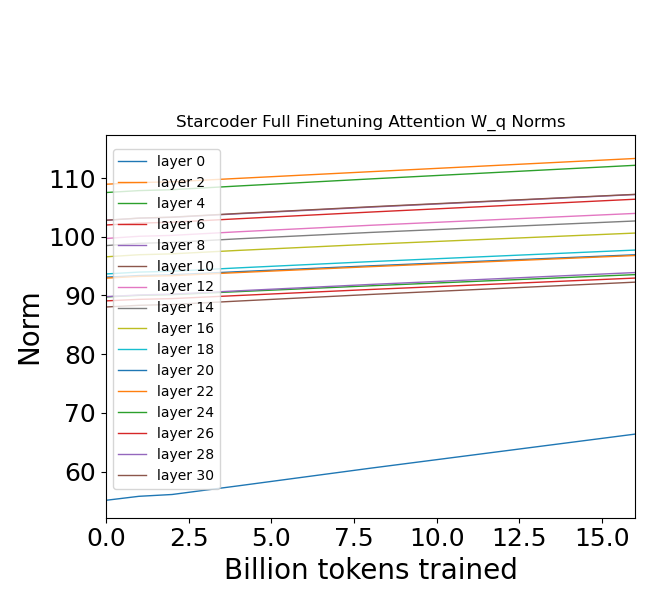}
        \caption{CPT Attention-Key}
    \end{subfigure}
        \begin{subfigure}{0.22\textwidth}
        \includegraphics[width=\linewidth]{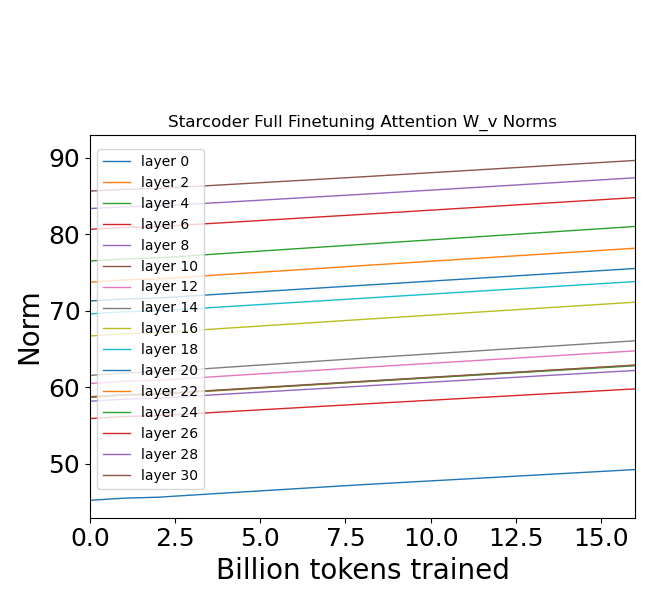}
        \caption{CPT Attention-Value}
    \end{subfigure}

    \vspace{0.5cm} 

    \begin{subfigure}{0.22\textwidth}
        \includegraphics[width=\linewidth]{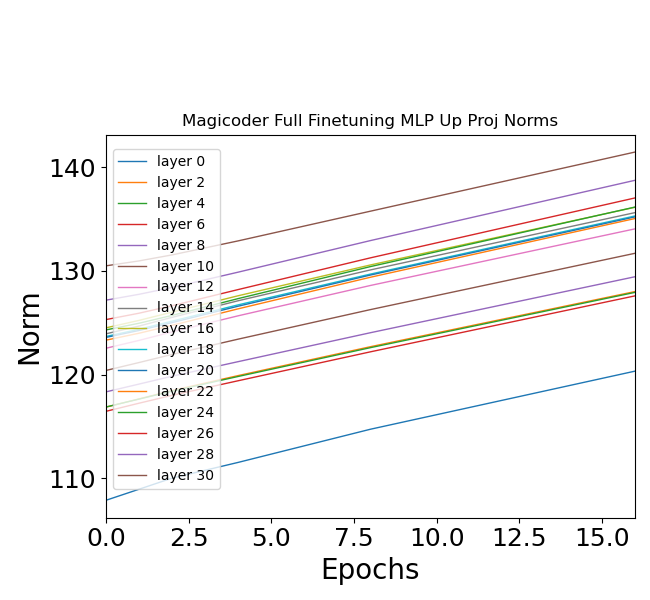}
        \caption{FFT MLP-Up}
    \end{subfigure}
    \begin{subfigure}{0.22\textwidth}
        \includegraphics[width=\linewidth]{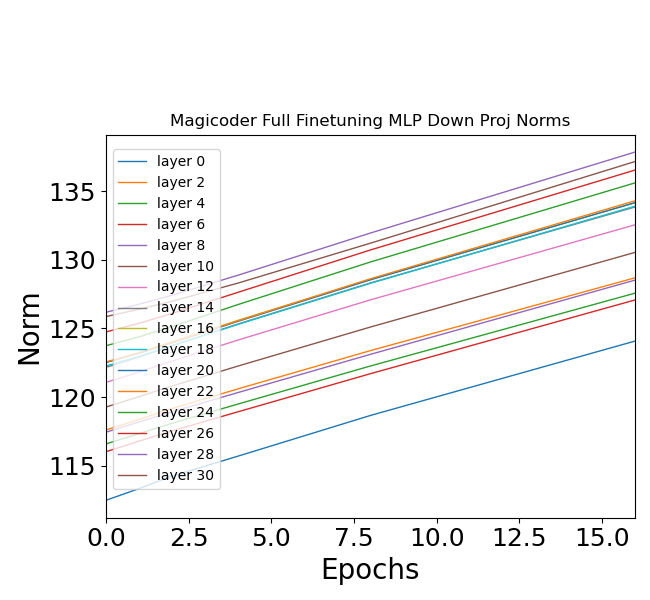}
        \caption{FFT MLP-Down}
    \end{subfigure}
    \begin{subfigure}{0.22\textwidth}
        \includegraphics[width=\linewidth]{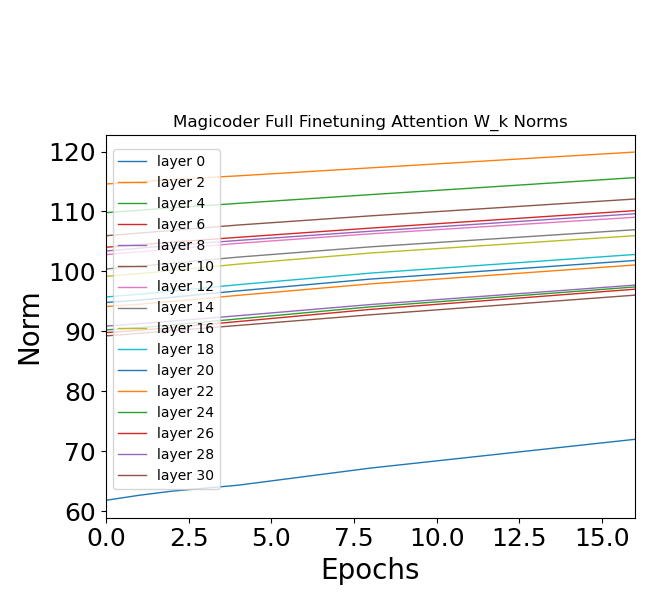}
        \caption{FFT Attention-Key}
    \end{subfigure}
    \begin{subfigure}{0.22\textwidth}
        \includegraphics[width=\linewidth]{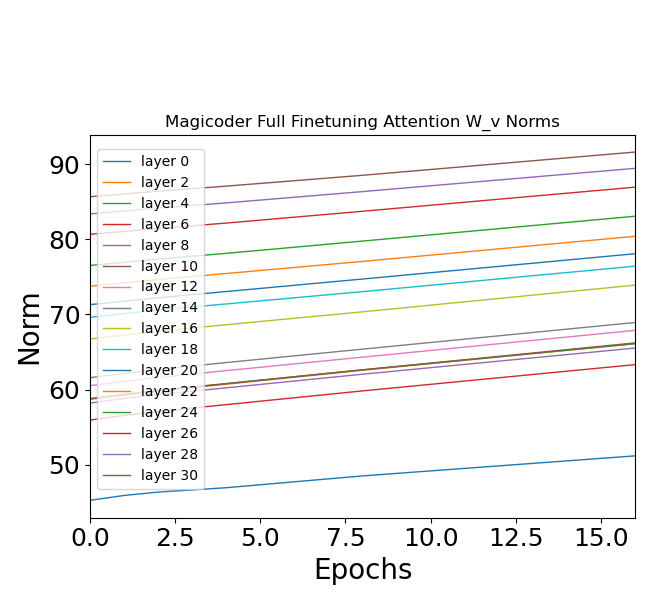}
        \caption{FFT Attention-Value}
    \end{subfigure}

    \vspace{0.5cm} 

    \begin{subfigure}{0.22\textwidth}
        \includegraphics[width=\linewidth]{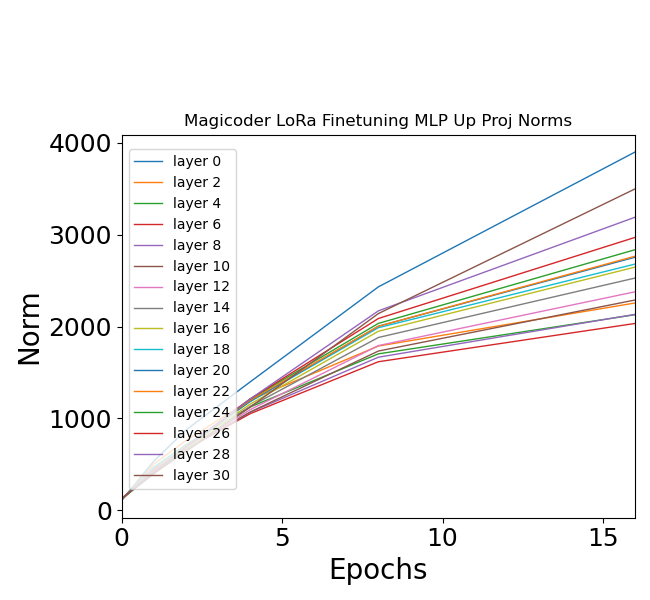}
        \caption{LFFT MLP-Up}
    \end{subfigure}
    \begin{subfigure}{0.22\textwidth}
        \includegraphics[width=\linewidth]{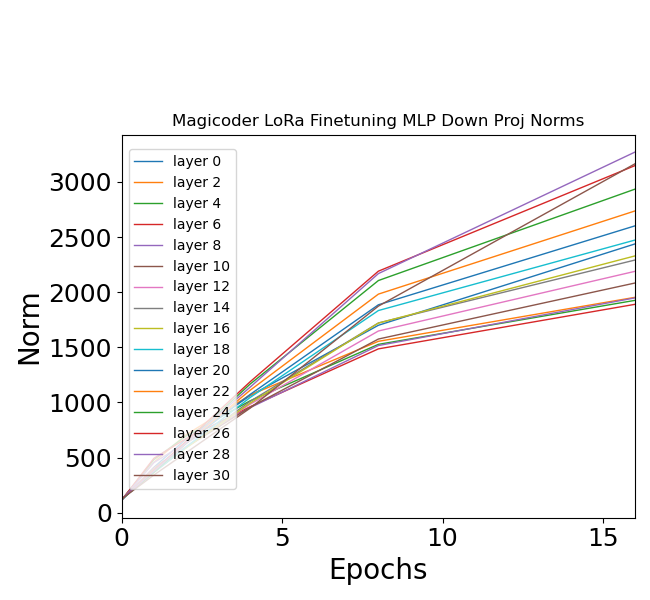}
        \caption{LFFT MLP-Down}
    \end{subfigure}
    \begin{subfigure}{0.22\textwidth}
        \includegraphics[width=\linewidth]{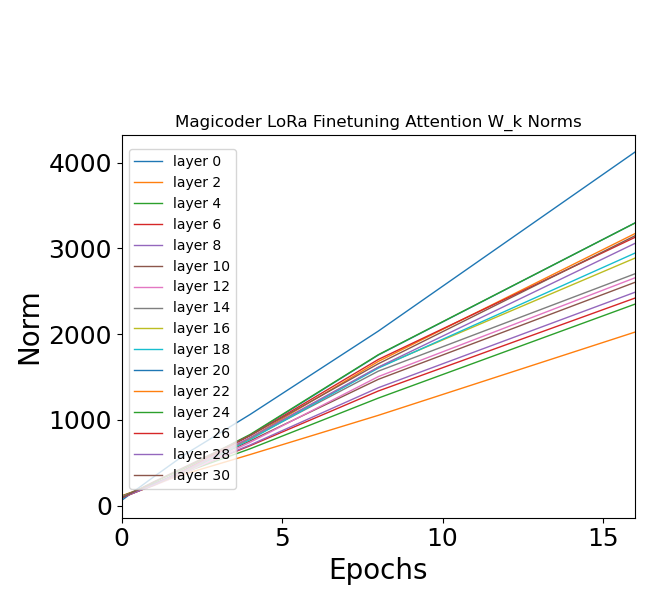}
        \caption{LFFT Attention-Key}
    \end{subfigure}
    \begin{subfigure}{0.22\textwidth}
        \includegraphics[width=\linewidth]{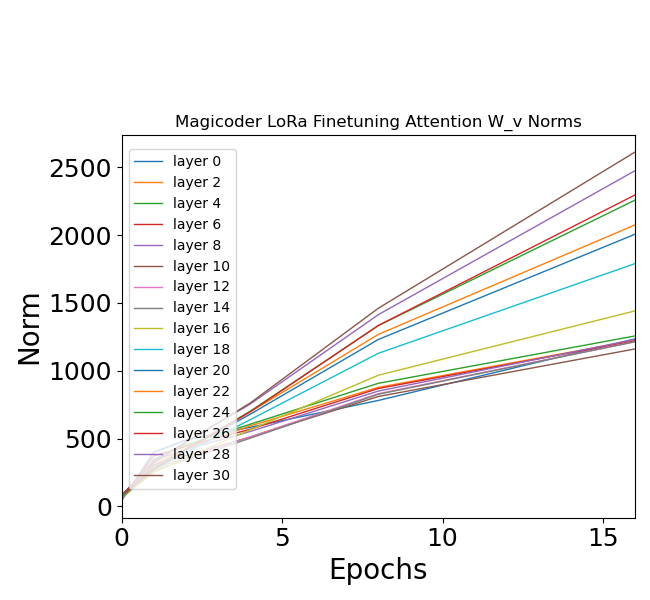}
        \caption{LFFT Attention-Value}
    \end{subfigure}

    \caption{Norm growth during post-training interventions }
    \label{fig:post-training}
\end{figure*}

For each of the above, we present post-training intervention results using the checkpoints provided in the study by \citet{lora-learns-less}. In each of the above cases, the model update equation can be written as:

\begin{equation}
    W_{new} = W_{old} + \Delta W
\end{equation}

Thus, the norm of the new weight matrix does not neccessarily have to decrease and can lie in the range as shown below using the triangle inequality:

\begin{equation}
   | |W_{old}|_F - |\Delta W|_F | \leq  |W_{new}|_F \leq |W_{old}|_F + |\Delta W|_F
\end{equation}

which means that after each update, the norm of the matrix being updated can both decrease or increase. Yet we find that the norm of the updated matrix in each of these interventions always increases. This can be seen in Figure \ref{fig:post-training}. The frobenius norm of all three MLP\footnote{Note that the Llama architecture has three MLP matrices instead of the common practice of two.} and attention matrices in Llama-2 (7B) can be seen to increase during CPT, FFT and LFFT. While this increase in norm is not detrimental to model performance, we believe this is because the norm of all the weight matrices involved increases in conjunction with each other. 

While our study is not exhaustive across different types of models or datasets used, we use the above findings to set the stage for the coming sections and to motivate studying norms of edited matrices when performing localized updates to a model. As only a few weight matrices are updated during parameter modifying knowledge editing methods, this phenomenon can have adverse consequences when the norm of some intermediate matrices grow while the remaining model remains frozen.



\begin{figure*}[h]
    \centering
    \begin{subfigure}{0.22\textwidth}
        \includegraphics[width=\linewidth]{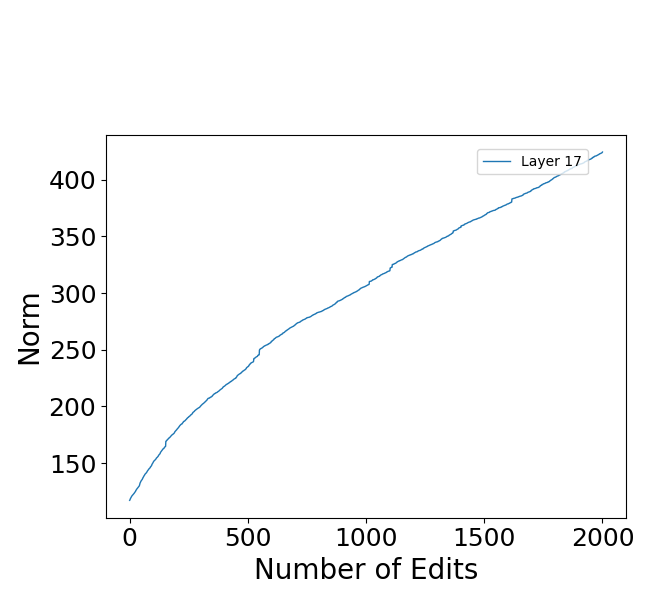}
        \caption{ROME-norm}
    \end{subfigure}
    \begin{subfigure}{0.22\textwidth}
        \includegraphics[width=\linewidth]{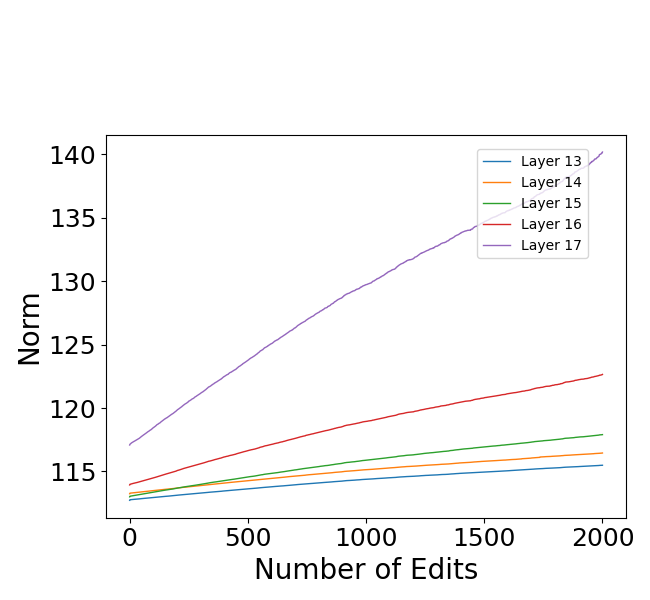}
        \caption{MEMIT-norm}
    \end{subfigure}
    \begin{subfigure}{0.22\textwidth}
        \includegraphics[width=\linewidth]{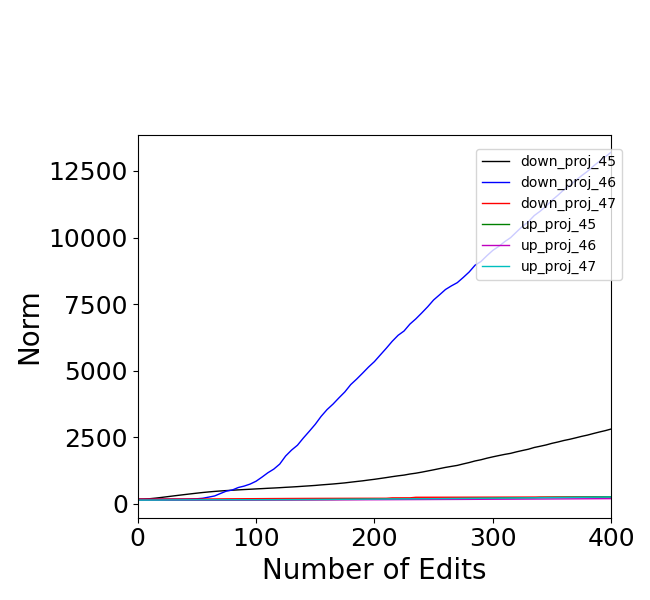}
        \caption{MEND-norm}
    \end{subfigure}
        \begin{subfigure}{0.22\textwidth}
        \includegraphics[width=\linewidth]{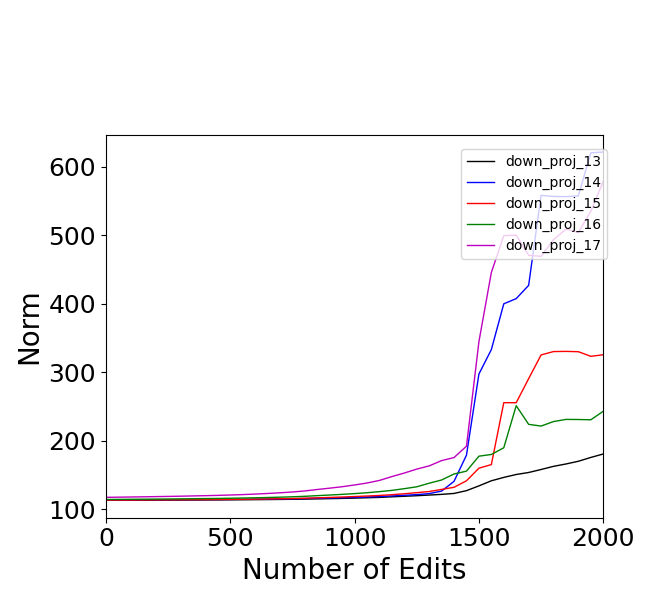}
        \caption{PMET-norm}
    \end{subfigure}

    \vspace{0.5cm} 

    \begin{subfigure}{0.22\textwidth}
        \includegraphics[width=\linewidth]{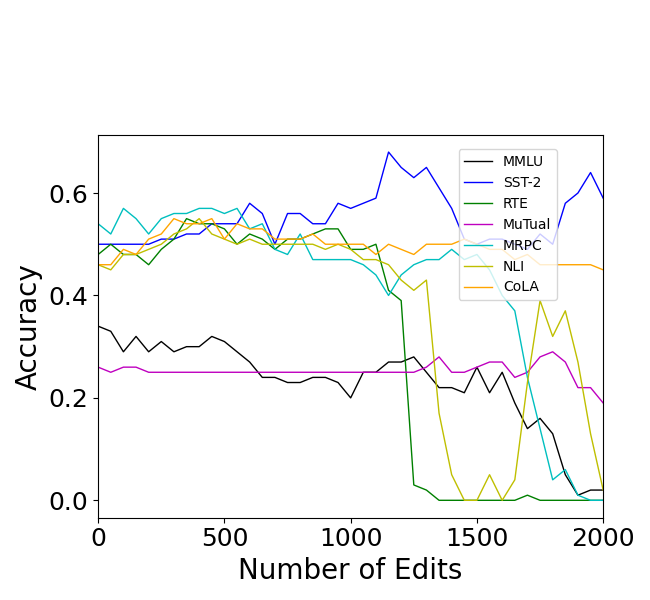}
        \caption{ROME-downstream}
    \end{subfigure}
    \begin{subfigure}{0.22\textwidth}
        \includegraphics[width=\linewidth]{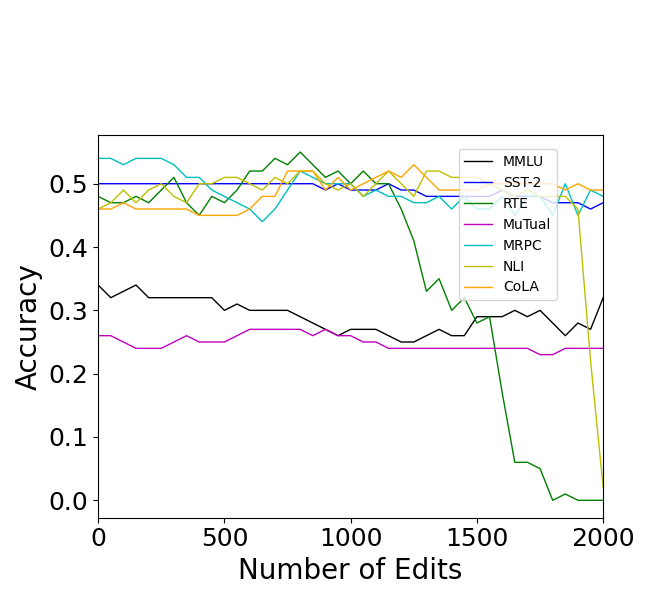}
        \caption{MEMIT-downstream}
    \end{subfigure}
    \begin{subfigure}{0.22\textwidth}
        \includegraphics[width=\linewidth]{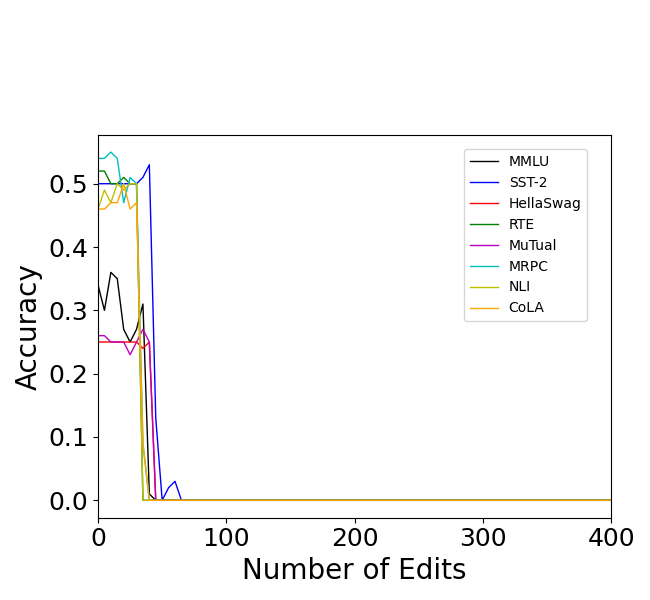}
        \caption{MEND-downstream}
    \end{subfigure}
        \begin{subfigure}{0.22\textwidth}
        \includegraphics[width=\linewidth]{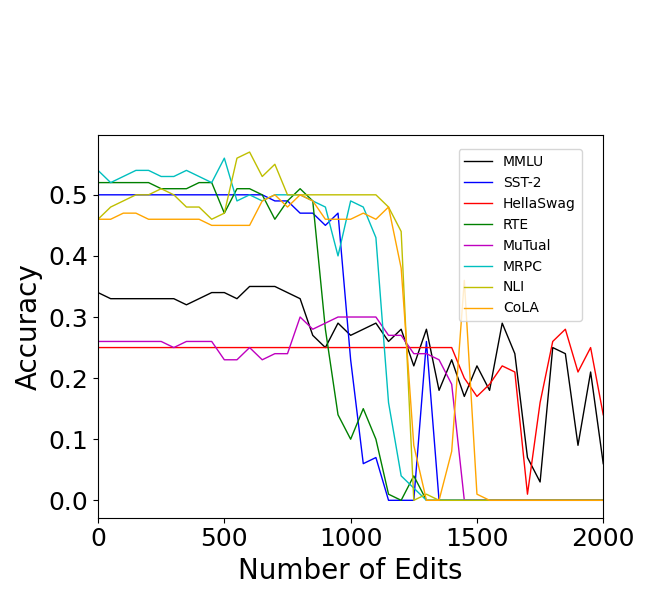}
        \caption{PMET-downstream}
    \end{subfigure}


    \caption{Norm growth and downstream performance during knowledge editing on GPT2-XL for different methods.}
    \label{fig:editing}
\end{figure*}

\section{Localized Updates during Knowledge Editing}
Knowledge editing is defined as the task of making data and compute efficient knowledge updates to large language model without compromising their general ability \cite{editing-survey, composable-inteventions}. In this paper, we focus on parameter-modifying knowledge editing methods, where knowledge is updated by changing the weights of the model \cite{ROME, MEMIT, akshat-unified}. The compute efficient component of knowledge editing comes from the fact that usually only one or a few layers of a model are updated when incorporating new knowledge. In this paper we focus on four popular knowledge editing methods - \textbf{ROME} \cite{ROME}, \textbf{MEMIT} \cite{MEMIT}, \textbf{MEND} \cite{MEND} and \textbf{PMET} \cite{PMET}. We perform sequential edits using these methods on GPT2-XL (1.5B) \cite{gpt-2} and GPT-J (6B) \cite{gpt-j} for 2000 edits and analyze the different properties of the updated matrices. The list of layers updated for the different model editing algorithms can be seen in Table \ref{tab:layers-edited}.

\begin{table}[h]
    \centering
    \begin{tabular}{c|c|c}
        \textbf{Algorithm} & \textbf{GPT2-XL} & \textbf{GPT-J} \\ \hline
        ROME & 17 & 5 \\ \hline
        MEMIT & 13-17 & 3-8 \\ \hline
        MEND & 45-47 & 25-27 \\ \hline
        PMET & 13-17 & 3-8 \\ 
    \end{tabular}
    \caption{List of layers edited when using each of the above algorithms for GPT2-XL and GPT-J.}
    \label{tab:layers-edited}
\end{table}

As can be seen in Figure \ref{fig:editing}, the norm of the edited matrices always increases for all four types of model editing methods used. Figure \ref{fig:editing} also shows evidences of model degradation as a function of sequential model editing evaluated on 8 downstream tasks including MMLU \cite{MMLU} and a few tasks from the GLUE benchmark \cite{glue}. A more detailed account of downstream measurement is provided in appendix \ref{sec:app:downstream}. To put the norm growth in perspective of the rest of the model, we plot the norm of the edited matrix after 100, 500 and 2000 edits along with the norm of other matrices present inside the model in Figure \ref{fig:norm-growth}. We clearly see the anomolous growth in norm for the edited layers, while the norm of the remaining layers remains constant. 

\begin{figure*}[h]
    \centering
    \begin{subfigure}{0.22\textwidth}
        \includegraphics[width=\linewidth]{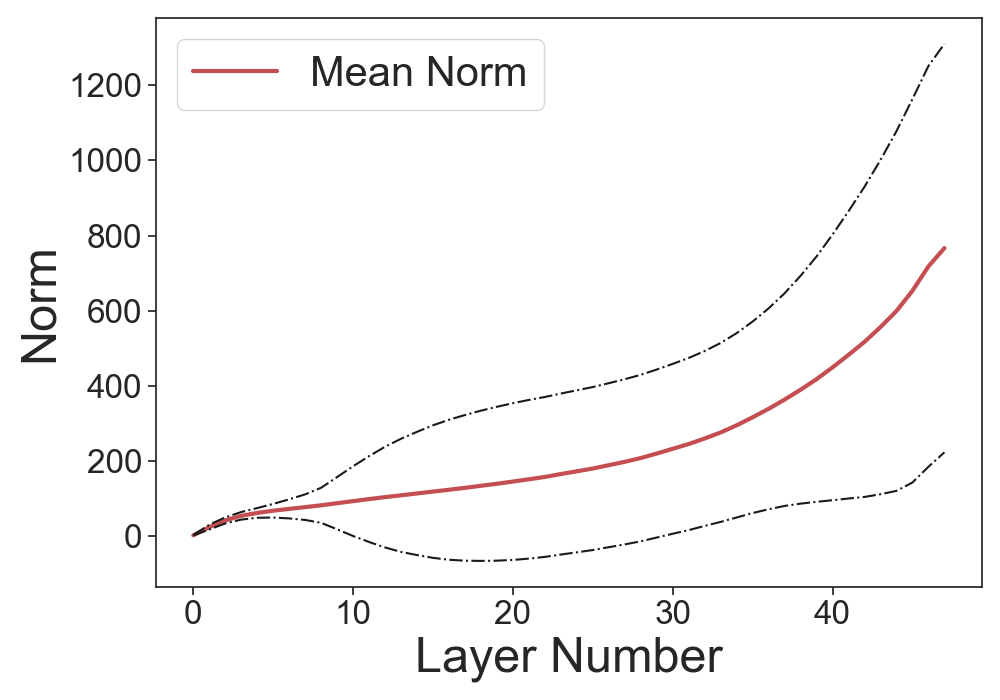}
        \caption{Unedited Model}
    \end{subfigure}
    \begin{subfigure}{0.22\textwidth}
        \includegraphics[width=\linewidth]{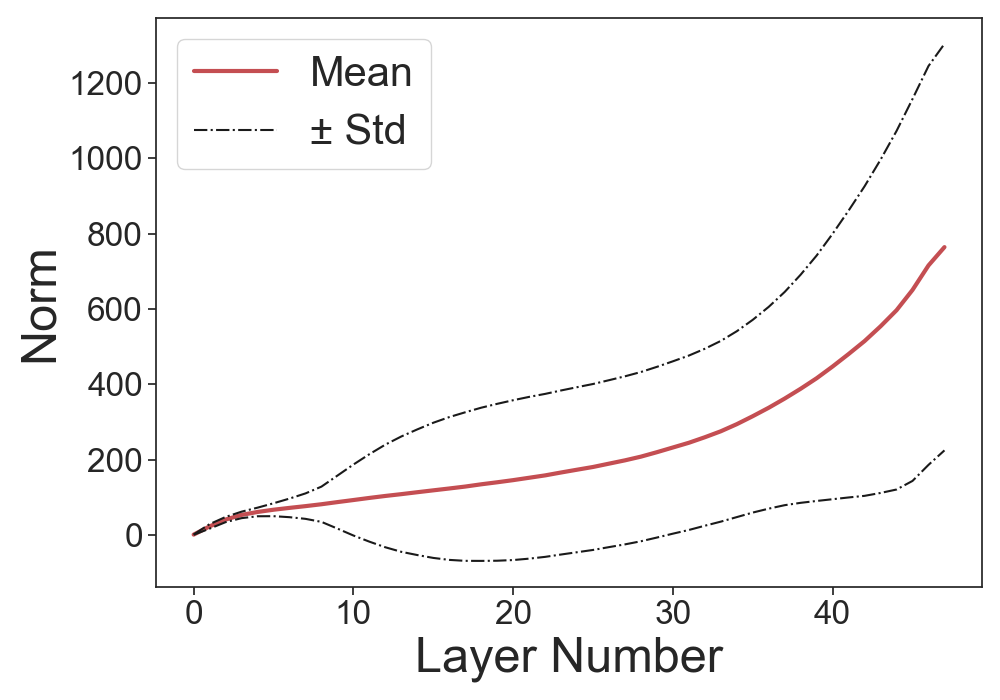}
        \caption{After 100 edits}
    \end{subfigure}
    \begin{subfigure}{0.22\textwidth}
        \includegraphics[width=\linewidth]{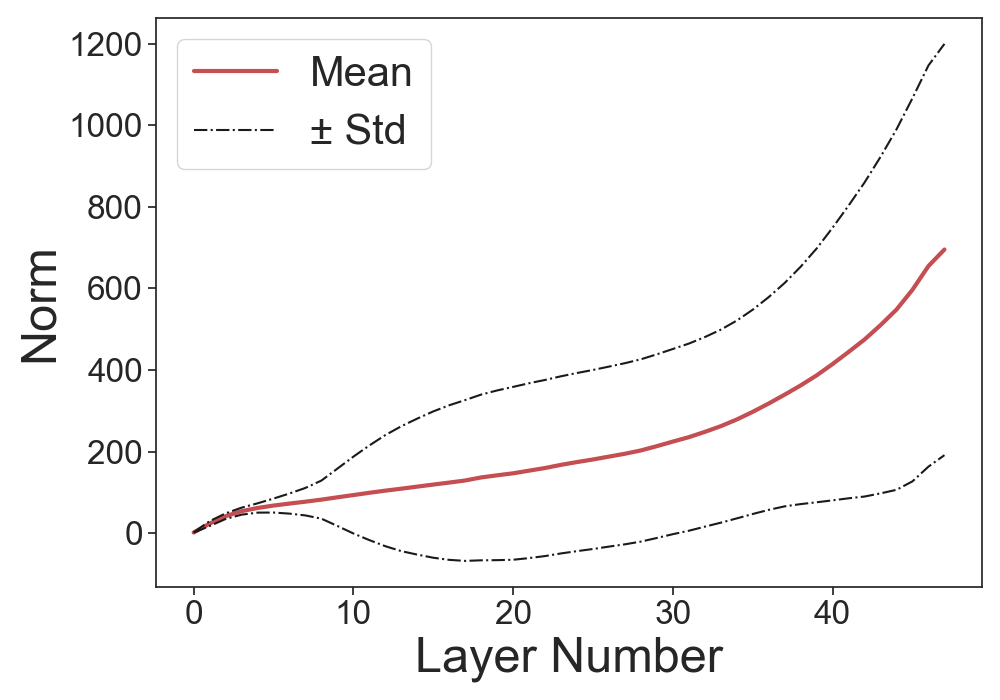}
        \caption{After 500 edits}
    \end{subfigure}
    \begin{subfigure}{0.22\textwidth}
        \includegraphics[width=\linewidth]{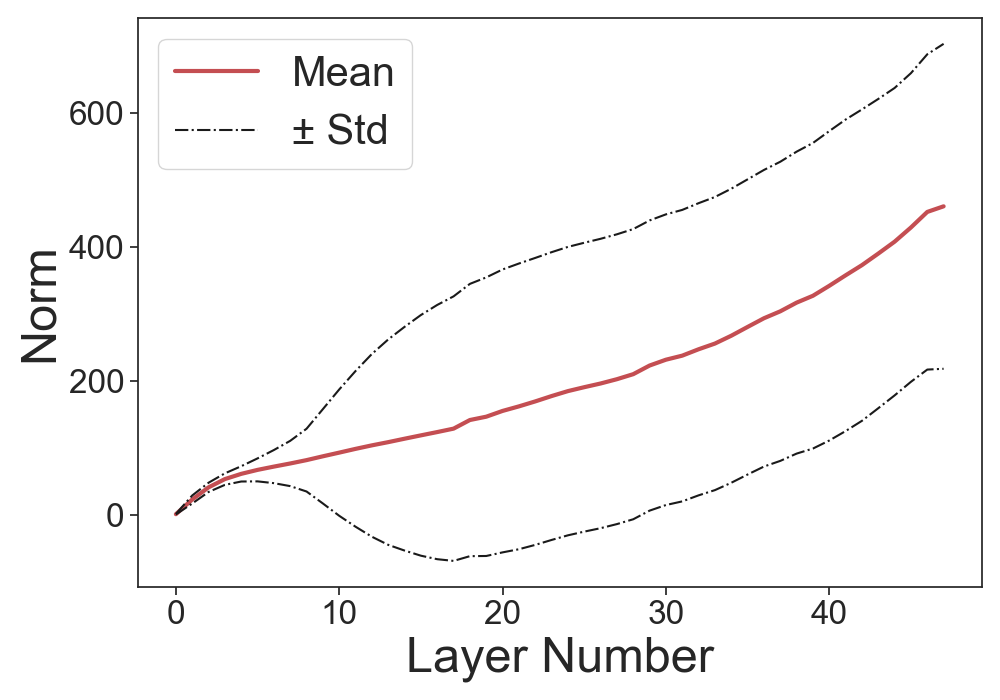}
        \caption{Adter 2000 edits}
    \end{subfigure}

    \vspace{0.5cm} 

    \caption{Activation Norms at different layers for unedited, edits 100, 500, 2000 for GPT2-XL/ROME.}
    \label{fig:activation-norm-growth}
\end{figure*}

\paragraph{Effect on Norm of Internal Activations.} To understand further effects of the increase norm of the edited matrix, we look at its effect on the corresponding activations that are output from that layer. To do so, we send 1 million tokens of wikipedia articles through both GPT2-XL and GPT-J, and study the norm and orientations of the internal activations of the model before and after editing. Specifically, we are looking at the residual stream vectors after each layers inside an LLM. The norm of the activation before and after editing can be seen in Figure \ref{fig:activation-norm-growth}. In this figure, we present the norm of the activation vectors for GPT2-XL when editing the model using ROME, comparing the unedited model with the model after 100, 500 and 2000 edits. We see that the activation norms remains the same until layer 17 for all cases, but slowly starts to decrease as we edit the model more. After 2000 edits, the norm of the activation vectors at layer 40 is much almost have that of the original enedited model, showing that the decrease in norm compounds as the activation vectors pass through the model. In contrast to the increasing norm of the edited weight matrices, the average norm of the activations decreases post editing. While in this paper we do not analyze the implications of this, it is studied in more detail in \citet{encore}. They show that the norms of activation vectors generated from the edited matrices increase while the norms of activations from the subsequent layers decreases, resulting in an increased importance of vectors generated from those layers.

\begin{figure*}[h]
    \centering
    \begin{subfigure}{0.24\textwidth}
        \includegraphics[width=\linewidth]{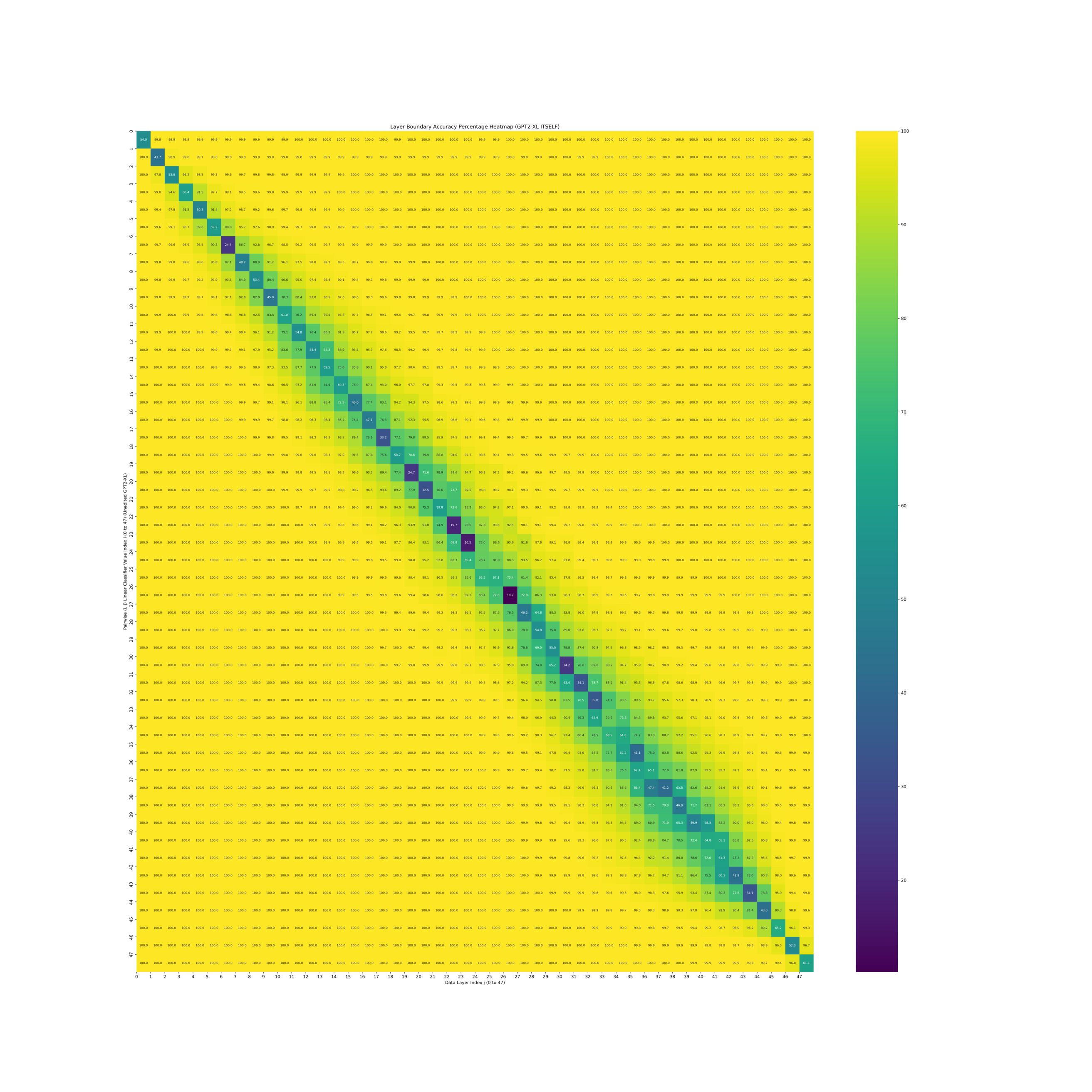}
        \caption{GPT2-XL unedited}
    \end{subfigure}
    \begin{subfigure}{0.24\textwidth}
        \includegraphics[width=\linewidth]{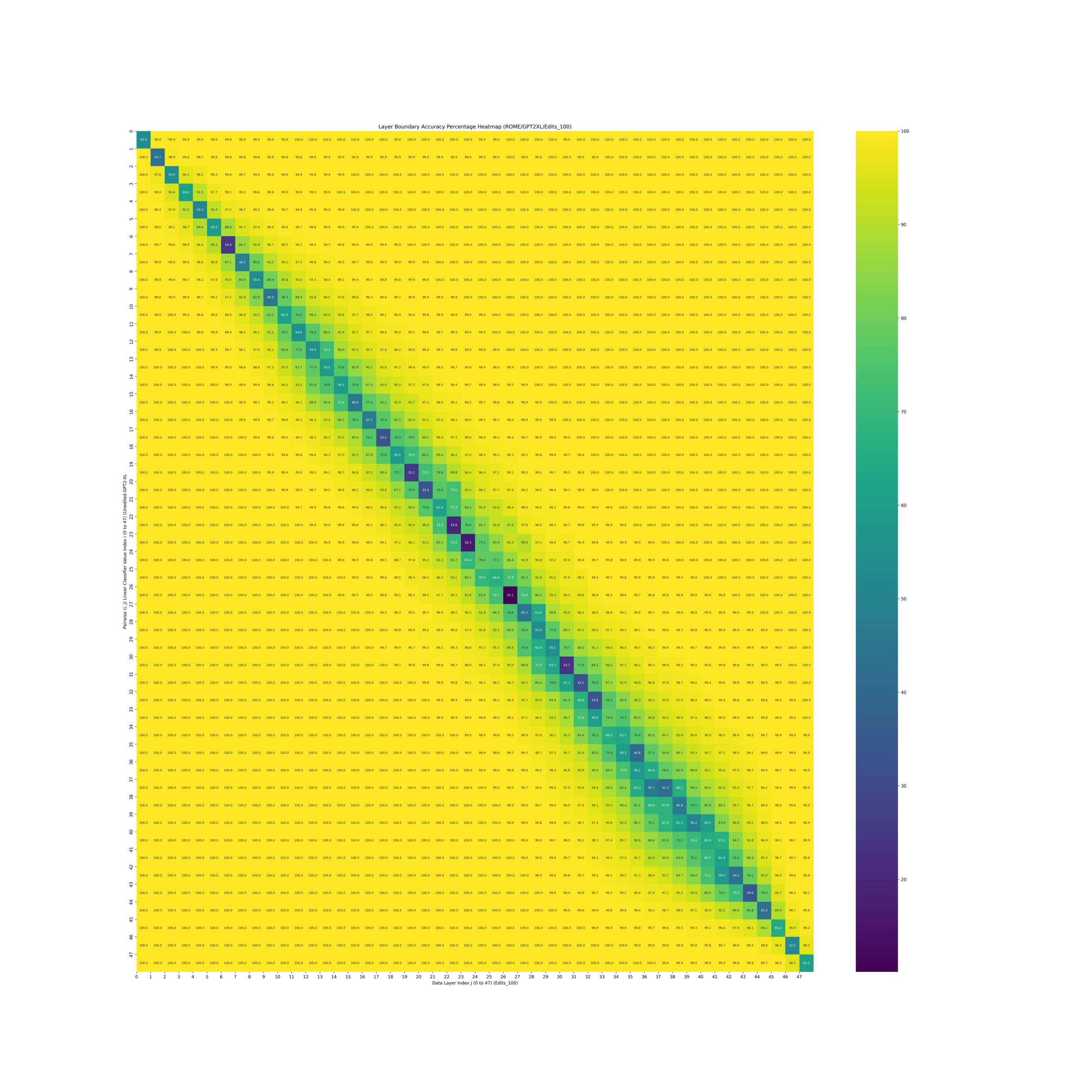}
        \caption{100 edits}
    \end{subfigure}
    \begin{subfigure}{0.24\textwidth}
        \includegraphics[width=\linewidth]{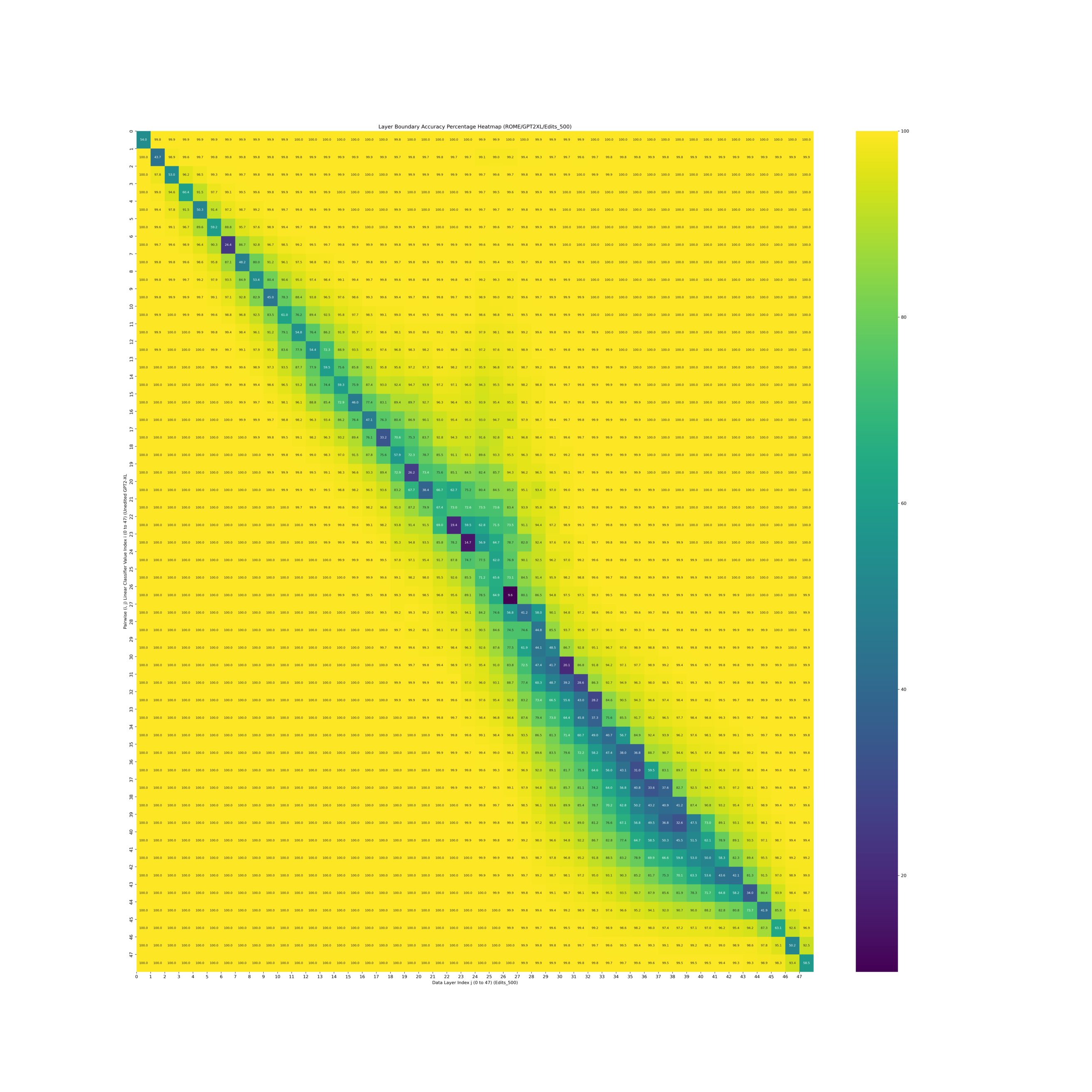}
        \caption{500 edits}
    \end{subfigure}
    \begin{subfigure}{0.24\textwidth}
        \includegraphics[width=\linewidth]{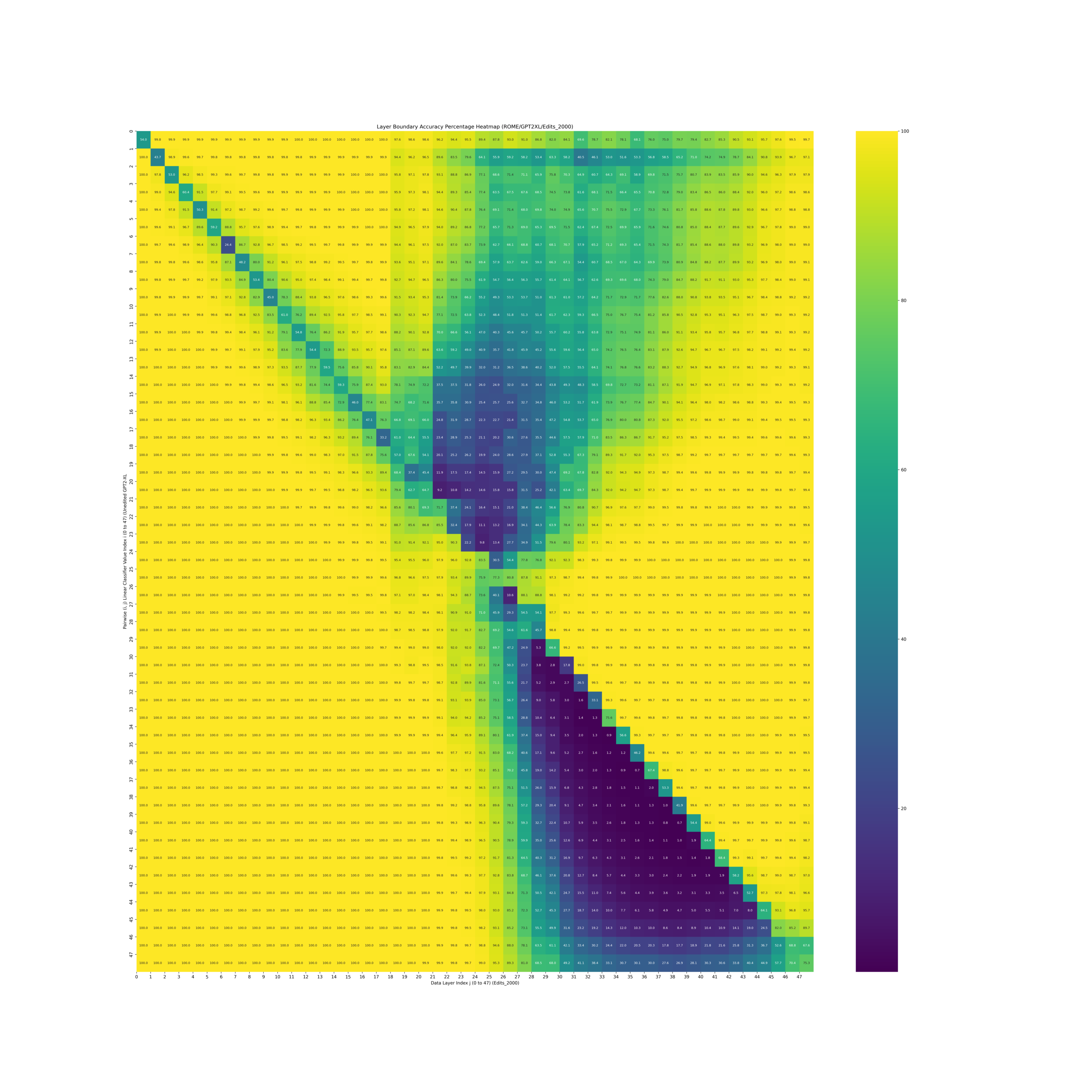}
        \caption{2000 edits}
    \end{subfigure}
       
    \vspace{0.5cm} 

    \caption{Classificatioin plots comparing GPT2-XL unedited model with model after 100, 500, 2000 edits using ROME.}
    \label{fig:classification}
\end{figure*}

\paragraph{Effect on Orientation of Internal Activations.}  We further analyze the orientations of the activation vectors. To do so, we create classifiers between the residual stream vectors of layers “$i$” versus “$j$”. Each classifier is a simple binary logistic regression classifier. A classifier $C_{i,j}$ is a binary classifier between two groups of vectors - residual stream vectors between layer $i$ and residual stream vectors between layer $j$. The classifier is trained with 800k training vectors and tested on 200k test vectors, equally divided into the two classes. This classifiers are trained only on the activations of the un-edited model. The accuracy of the classifier can be seen in Figure \ref{fig:classification} (a). We see that the simple binary classifiers reach almost a 100\% accuracy when classifying the activation vectors from different layers. \textbf{This shows that the activation vectors of different layers are linearly separable from each other}. When a classifier is trained with activation vectors from the same layers, which forms the diagonal line in Figure \ref{fig:classification} (a), we see that the classification accuracy is 50\% or at random. This shows that the activation vectors between the same layers are not linearly separable, stengthening the linear seprability claim. 

We save the classifiers trained on the un-edited model and use them to clasify the activation vectors of the edited model. Figure \ref{fig:classification} shows this for ROME, where we analyze the output for 200k activation vectors when passed through a model sequentially edited for 100, 500 and 2000 edits. The cell $i-j$ corresponds to the case where classifier $C_{i,j}$, trained on the activations coming from the $i^{th}$ and $j^{th}$ layer of the un-edited model, are used to classifiy the activations of the $j^{th}$ layer of the edited model. We want to re-iterate to the reader that the graphs in Figure \ref{fig:classification} are not supposed to be symmetrical by definition. We will give an example below. For cell 3-5, we use the classifier $C_{3,5}$ trained between activations of layer 3 and layer 5 of the un-edited model, but use it to classify the activations of layer 5 at test time. For cell 5-3, we use the same classifier $C_{3,5}$, but this time we use it to classify the activations of layer 3. 

For cell $i-j$, if the orientations in space of the activation vectors of the edited models remains the same as that of the unedited model, the classifier $C_{i,j}$ would assign it the class $j$. But we see in Figure \ref{fig:classification} that this is not the case. The test time accuracy of the classifier begins to get worse for activation vectors right after the edited layer. This shows that the activation vectors are not only changing in norms but also the regions they occupy in space. This is so much so that the activations coming out of layer $j$ of the edited model after 2000 edits now lie in a region for layer $i$, leading to an incorrect classification accuracy for classifier $C_{i,j}$.  

With this study, we show that the increasing norm not only changes the norm of the internal activations of the model, but also the orientations in space in which these vectors lie. We observe that post-editing, the internal activations now lie in a very different region in space. The region is so different that a linear classifier trained to identify the activation vectors of a specific layer is now unable to identify the output activations from the specific layer.

\section{Conclusion}
This study highlights a critical challenge in the domain of localized updates for large language models: the persistent increase in the norm of updated matrices during sequential knowledge editing. While this phenomenon appears universal across post-training intervention methods, its implications are particularly pronounced in localized knowledge editing, where only specific parts of the model are modified. The resulting imbalance leads to downstream performance degradation, as evidenced by changes in both the norm and orientation of internal activations. Our findings emphasize the need for innovative strategies to address these challenges, paving the way for more robust and sustainable approaches to localized knowledge editing in LLMs. This work serves as a foundational step towards understanding and mitigating the inherent limitations of current techniques, with the ultimate goal of enabling dynamic and scalable updates to pre-trained models. A follow-up work by \citet{encore} overcomes these highlighted limitations by using appropriate regularizations with knowledge editing, allowing long-term sequential knowledge editing.


\bibliography{aaai25}

\appendix

\subsection{A.1 Downstream Performance Measurement}\label{sec:app:downstream}

In this paper, we assess model degradation by measuring downstream performance at regular intervals of edits. Our evaluation suite is wide-ranging and consists of the following 8 tasks – sentiment analysis (SST2) \cite{sst2}, paraphrase detection (MRPC) \cite{mrpc}, natural language inference (NLI, RTE) \cite{nli1, nli2, nli3, nli4}, commonsense natural language inference (HellaSwag) \cite{hellaswag}, linguistic acceptability classification (CoLA) \cite{cola}, multi-turn dialogue reasoning (MuTual) \cite{mutual} and massive multitask language understanding (MMLU) \cite{MMLU}.

For each task, we created a subset of 100 examples balanced across all multiple-choice options. The models were evaluated on the tasks above, and the accuracy score was measured at intervals of 5 edits for MEND, and 20 edits for PMET, ROME, and MEMIT. MEND leads to a rapid degradation within 100 edits, hence why a smaller granularity of 5 edit interval was used; a 20 edit interval was used for other methods to cut down on computation time. 

In order to improve models' initial performance and achieve meaningful signals, we provided few-shot examples. The few-shot prompt templates used for each task are shown in Figures \ref{fig:sst-prompt}-\ref{fig:hellaswag-prompt}.

\begin{figure*}
    \centering
    \fbox{
        \parbox{0.8\textwidth}{
            Review : an exhilarating futuristic thriller-noir , minority report twists the best of technology around a gripping story , delivering a riveting , pulse intensifying escapist adventure of the first order \newline
            Sentiment : positive \newline \newline
            Review : try as i may , i ca n't think of a single good reason to see this movie , even though everyone in my group extemporaneously shouted , ` thank you ! ' \newline
            Sentiment : negative \newline \newline
            Review : the film 's performances are thrilling .  \newline
            Sentiment : positive \newline \newline
            Review : vera 's technical prowess ends up selling his film short ; he smoothes over hard truths even as he uncovers them . \newline
            Sentiment : negative \newline \newline
            Review : [input] \newline
            Sentiment :
        }
    }
    \caption{Few shot prompt template used for SST-2 }
    \label{fig:sst-prompt}
\end{figure*}

\begin{figure*}
    \centering
    \fbox{
        \parbox{0.8\textwidth}{
            Question: Which expression is equivalent to 4 x 9? \newline
            (A) (4x 4) + (4x5) \newline
            (B) (4+4) x (4+5) \newline
            (C) (4+4)+(4+5) \newline
            (D) (4x 4) x (4x5) \newline
            Answer: A\newline\newline
            Question: A marketing researcher is conducting a survey in a large selling area by contacting a small group of people that is representative of all people in that area. The small, representative group is known as the \newline
            (A) population\newline
            (B) sample\newline
            (C) stratification\newline
            (D) universe\newline
            Answer: B\newline\newline
            Question: A research participant eats half a bowl of M\&M candies, and then stops eating. How would a motivation researcher using drive reduction theory explain this participant's behavior?\newline
            (A) Humans are instinctively driven to eat sugar and fat when presented to them.\newline
            (B) The Yerkes-Dodson law explains that people will eat food when presented to them, but usually in moderate amounts in order to avoid being perceived as selfish.\newline
            (C) The primary drive of hunger motivated the person to eat, and then stop when she/he regained homeostasis.\newline
            (D) The research participant was satisfying the second step on the hierarchy of needs: Food needs.\newline
            Answer: C\newline\newline
            Question: In a deductively valid argument\newline
            (A) If all the premises are true, the conclusion must be true\newline
            (B) The conclusion has already been stated in its premises\newline
            (C) If all the premises are true, the conclusion may still be false\newline
            (D) Both A and B\newline
            Answer: D\newline\newline
            Question: [input] \newline
            Answer:
        }
    }
    \caption{Few shot prompt template used for MMLU}
    \label{fig:mmlu-prompt}
\end{figure*}

\begin{figure*}
    \centering
    \fbox{
        \parbox{0.8\textwidth}{
            Are the sentences paraphrases of each other. \newline
            Sentence 1: Federal regulators have turned from sour to sweet on a proposed \$2.8 billion merger of ice cream giants Nestle Holdings Inc. and Dreyer 's Grand Ice Cream Inc .\newline
            Sentence 2: Federal regulators have changed their minds on a proposed \$2.8 billion merger of ice cream giants Nestle Holdings and Dreyer 's Grand Ice Cream .\newline
            Answer: Yes\newline\newline
            Are the sentences paraphrases of each other.\newline
            Sentence 1: In the year-ago quarter , the steelmaker recorded a profit of \$16.2 million , or 15 cents per share , on sales of \$1.14 billion .\newline
            Sentence 2: In the second quarter last year , AK Steel reported a profit of \$16.2 million , or 15 cents a share .\newline
            Answer: No\newline\newline
            Are the sentences paraphrases of each other.\newline
            Sentence 1: He added : ``I 've never heard of more reprehensiblebehaviour by a doctor .\newline
            Sentence 2: The Harrisons ’ lawyer Paul LiCalsi said : “ I ’ ve never heard of more reprehensible behaviour by a doctor .\newline
            Answer: Yes\newline\newline
            Are the sentences paraphrases of each other.\newline
            Sentence 1: While dioxin levels in the environment were up last year , they have dropped by 75 percent since the 1970s , said Caswell .\newline
            Sentence 2: The Institute said dioxin levels in the environment have fallen by as much as 76 percent since the 1970s .\newline
            Answer: No\newline\newline
            Are the sentences paraphrases of each other.\newline
            Sentence 1: [input 1]\newline
            Sentence 2: [input 2]\newline
            Answer:
        }
    }
    \caption{Few shot prompt template used for MRPC}
    \label{fig:mrpc-prompt}
\end{figure*}

\begin{figure*}
    \centering
    \fbox{
        \parbox{0.8\textwidth}{
            Is this sentence linguistically acceptable?\newline
            Sentence: The evidence assembled by the prosecution convinced the jury.\newline
            Answer: Yes\newline\newline
            Is this sentence linguistically acceptable?\newline
            Sentence: I live at the place where Route 150 crosses the Hudson River and my dad lives at it too.\newline
            Answer: No\newline\newline
            Is this sentence linguistically acceptable?\newline
            Sentence: The government's imposition of a fine.\newline
            Answer: Yes\newline\newline
            Is this sentence linguistically acceptable?\newline
            Sentence: Sam gave the ball out of the basket.\newline
            Answer: No\newline\newline
            Is this sentence linguistically acceptable?\newline
            Sentence: [input] \newline
            Answer: 
        }
    }
    \caption{Few shot prompt template used for RTE}
    \label{fig:rte-prompt}
\end{figure*}

\begin{figure*}
    \centering
    \fbox{
        \parbox{0.8\textwidth}{
            The town is also home to the Dalai Lama and to more than 10,000 Tibetans living in exile. \newline
            Question: The Dalai Lama has been living in exile since 10,000. True or False? \newline
            Answer: True \newline\newline
            P. Prayong, who like Kevala belongs to the Theravada sect of Buddhism, chose India over other Buddhist majority nations as it is the birthplace of Gautama Buddha. \newline
            Question: P. Prayong is a member of Theravada. True or False? \newline
            Answer: False \newline\newline
            The medical student accused of murdering an erotic masseuse he met on Craigslist is drowning in more than \$100,000 in student loan debt and is so broke he can't afford to pay an attorney, according to court papers. Philip Markoff, a 23-year-old suspended Boston University medical school student, owes \$130,000 in student loans and does not get money from his parents, leaving him to lean on a taxpayer-funded attorney for his defense, according to a court document in Boston Municipal Court that labels him indigent. Markoff graduated from the State University of New York-Albany and was a second-year medical student at BU.\newline
            Question: The medical student Philip Markoff was engaged. True or False?\newline
            Answer: True\newline\newline
            Traditionally, the Brahui of the Raisani tribe are in charge of the law and order situation through the Pass area. This tribe is still living in present day Balochistan in Pakistan. \newline
            Question: The Raisani tribe resides in Pakistan. True or False? \newline
            Answer: False \newline\newline
            The latest attacks targeted the U-S embassy and a top prosecutor's office in the Uzbek capital.\newline
            Question: [input]. True or False?\newline
            Answer: 
        }
    }
    \caption{Few shot prompt template used for CoLA}
    \label{fig:cola-prompt}
\end{figure*}

\begin{figure*}
    \centering
    \fbox{
        \parbox{0.8\textwidth}{
            Turkey is unlikely to become involved in, or allow U.S. forces to use Turkish territory in a Middle East war that does not threaten her territory directly. entails the U.S. to use Turkish military bases. \newline True or False? \newline Answer: False \newline \newline
            Brooklyn Borough Hall featured a Who's Who in New York's literary community during the second annual Brooklyn Book Festival. According to Brooklyn Borough President Marty Markowitz, the borough's zip code 11215 boasts more authors than anywhere else in the country. It appeared to be the case on Sunday. More than 100 authors were featured at the day-long event, including The Basketball Diaries writer Jim Carroll, former M*A*S*H star Mike Farrell, author and illustrator Mo Willems, Jack Kerouac's sometime lover and National Book Critics Circle Award recipient Joyce Johnson and PEN American Center President Francine Prose. entails the The Brooklyn Book Festival is held in Brooklyn Borough every year. \newline True or False? \newline Answer: True \newline\newline
            NASA's Saturn exploration spacecraft, Cassini , has discovered an atmosphere about the moon Enceladus . This is the first such discovery by Cassini, other than Titan , of the presence of an atmosphere around a Saturn moon. entails the Titan is the fifteenth of Saturn's known satellites.\newline True or False? \newline Answer: False \newline\newline
            Dr. Eric Goosby, a pioneer in the fight against AIDS, is President Obama's choice to run the American effort to combat the disease globally, the White House announced Monday. The President's Emergency Plan For AIDS Relief, known as Pepfar, was championed by President George W. Bush. It is expected to spend \$48 billion over the next five years and is credited with markedly reducing the disease's death rate. Its prevention policy has been controversial because of its emphasis on socially conservative methods. With a new administration and a Democratic majority in the House, organizations seeking prevention choices beyond abstinence and fidelity — including a renewed commitment to distributing condoms — are eager to try to rewrite the guidelines. entails the Pepfar is committed to fighting AIDS. \newline True or False? \newline Answer: True\newline\newline
            [input] \newline True or False? \newline Answer:
        }
    }
    \caption{Few shot prompt template used for NLI}
    \label{fig:nli-prompt}
\end{figure*}

\begin{figure*}
    \centering
    \fbox{
        \parbox{0.8\textwidth}{
            Given the following: \newline f : hi how are you doing ? \newline m : i 've been good . i 'm in school right now . \newline f : what school do you go to ? \newline m : i go to a cooking school . i will spend one year there . \newline f : really ? i know you love drawing and designing most . how do you like cooking so far ? \newline m : i like it so far , my classes are pretty good , and i plan to have my own restaurant in the future . \newline 
            Which choice is correct?\newline 
            (A)f : really ? you mean you do n't like cooking but you plan to start a restaurant in the future ? \newline 
            (B)f : really ? you mean your classes are pretty good and you plan to start a restaurant in the future ? \newline 
            (C)f : so , although your classes are not pretty good , you plan to become a teacher in the future ? \newline 
            (D)f : so , although you do n't love drawing or designing , you want to design a building in the future ? \newline 
            Answer: B \newline\newline
            
            Given the following: \newline f : dad , can i go out tonight ? \newline m : no , i 'm sorry . you ca n't . \newline f : can i ask nancy for dinner ?\newline  m : ok , but you ca n't let your brother alone .\newline 
            Which choice is correct?\newline 
            (A)f : ok. then i will ask nancy for dinner tonight .\newline 
            (B)f : i will stay at home alone because i do n't want ask nancy for dinner .\newline 
            (C)f : ok. so i can ask nancy for dinner tonight if i do n't have to have my brother companied .\newline 
            (D)f : i have to stay home with me brother because i will not ask nancy to have dinner .\newline 
            Answer: A\newline\newline
            
            Given the following:\newline m : that was such an interesting english program , i wish you enjoyed it as much as i did .\newline f : i must tell you the truth that i fell asleep after the first few minutes , as i could n't understand many of the words in the program .\newline
            Which choice is correct?\newline
            (A)m : seems that you found the program boring .\newline
            (B)m : the english program is indeed difficult . i feel it , too !\newline
            (C)m : so you think the english program is difficult . do n't worry , let me help you .\newline
            (D)m : good to know that you also found it interesting .\newline
            Answer: C \newline\newline
            
            Given the following: \newline m : should n't we invite kathy to the party tonight ? \newline f : invite kathy ? she is the one who 's planning the whole thing .\newline
            Which choice is correct?\newline
            (A)m : i have invited cathy .\newline
            (B)m : cathy planed the party . but she wo n't attend it cause she has no time .\newline
            (C)m : what a pity ! cathy is too busy to come .\newline
            (D)m : cathy planed the party . of course she will attend it .\newline
            Answer: D\newline\newline
            
            Q: Given the following: [input] \newline
            Which choice is correct? \newline
            (A) \newline
            (B) \newline
            (C) \newline
            (D) \newline
            Answer:
        }
    }
    \caption{Few shot prompt template used for MuTual}
    \label{fig:mutual-prompt}
\end{figure*}

\begin{figure*}
    \centering
    \fbox{
        \parbox{0.8\textwidth}{
            Activity: Hitting a pinata \newline
            Finish this sentence: A small girl runs away from the pinata while holding the stick. The child returns and hits the pinata one time. the people \newline
            Options: \newline
            (A) continue to remove key ingredients from the bags. \newline
            (B) cheer for the child getting hit. \newline
            (C) watch while the girl throws the stick twice. \newline
            (D) all clap and the girl smiles and turns toward the camera. \newline
            Answer: D \newline\newline
            Activity: Personal Care and Style \newline
            Finish this sentence: [header] How to apply base makeup [title] Use concealer. [step] You will wreck your look if you've got dark circles under your eyes or have blemishes throughout your face. Don't feel badly about such circles or blemishes, though. \newline
            Options: \newline
            (A) This is a natural part of makeup, and you can use that to your advantage. If you must draw your brows in, that will be where most of the concealer and powder will come in.\newline
            (B) You can always cover them up if need be. You can also apply some false lashes-it will look better, even if you only have the eyelashes.\newline
            (C) You can cover them up with concealer. Apply the concealer in an upside-down pyramid shape.\newline
            (D) [substeps] Use some concealer up off of your eyelids so you can wear concealer successfully without using too much. Apply concealer a little higher up on your eyelid than the same way.\newline
            Answer: C \newline\newline
            Activity: Vacuuming floor \newline
            Finish this sentence: The video shows a demonstration of dyson vacuum cleaner and how well it can clean particles from the floor. There's rice grain, flour and other food particles scattered on the floor. the demonstrator\newline
            Options:\newline
            (A) then kneels down next to the tested vacuum cleaner and then goes on the floor to clean the floor using a towel.\newline
            (B) turns on the vacuum cleaner and begins vacuuming the floor with the yellow dyson vacuum.\newline
            (C) wipes down the floor with towels.\newline
            (D) talks about a very hard floor that is soft and easy to clean.\newline
            Answer: B\newline\newline
            Activity: Personal Care and Style\newline
            Finish this sentence: [header] How to grow an afro with african american hair [title] Use the right comb. [step] Traditional combs and brushes will damage your curly hair, making it frizzier. Instead of these, use a wide-tooth comb, afro pick, or even your fingers to comb your hair.\newline
            Options:\newline
            (A) [substeps] Don't comb more than necessary! Just use your comb, pick, or fingers to get rid of any tangles. If you have a lot of tangles, try buying a detangling spray specifically designed for african american hair. \newline
            (B) Most african american hair is fairly straight, but with wavy hair that isn't straight, you have a naturally naturally curly afro. [substeps] If you have short hair, use a comb with some thinner bristles to help tame your hair and prevent frizz. \newline
            (C) Using a large comb will really work as well to take care of your hair without damaging its natural oils. [substeps] Small combs are best because more tangles can be too rough for curly hair. \newline
            (D) If you have very curly hair, you may prefer using a cap or a hair net instead, as these tend to do little damage instead of much. [title] Trim your hair every six to eight months. \newline
            Answer: A \newline\newline
            Activity: [input]\newline
            Finish this sentence: [input] \newline
            Options: \newline
            (A)  \newline
            (B)  \newline
            (C) \newline
            (D) \newline
            Answer:
        }
    }
    \caption{Few shot prompt template used for HellaSwag}
    \label{fig:hellaswag-prompt}
\end{figure*}

\begin{figure*}[h]
    \centering
    \begin{subfigure}{0.22\textwidth}
        \includegraphics[width=\linewidth]{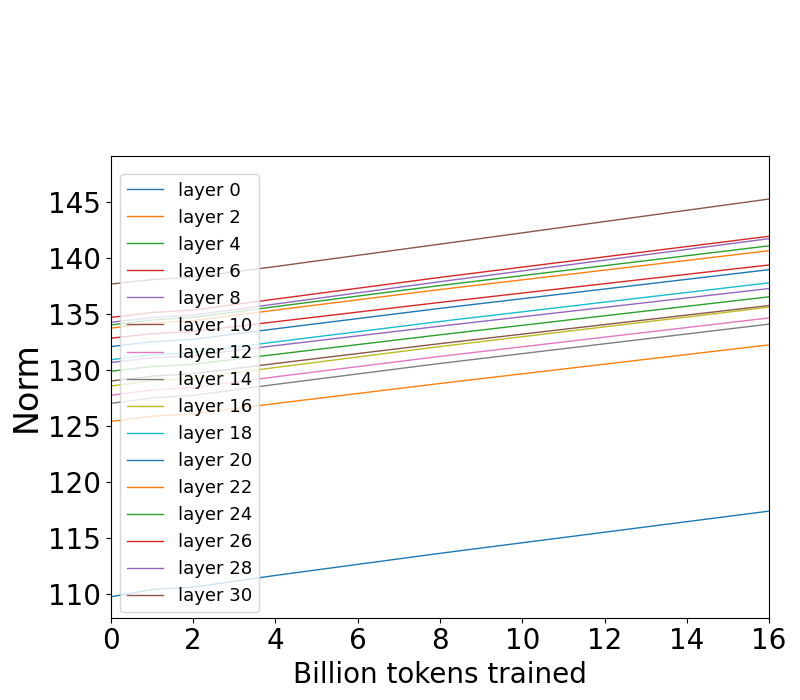}
        \caption{CPT MLP-Gate}
    \end{subfigure}
    \begin{subfigure}{0.22\textwidth}
        \includegraphics[width=\linewidth]{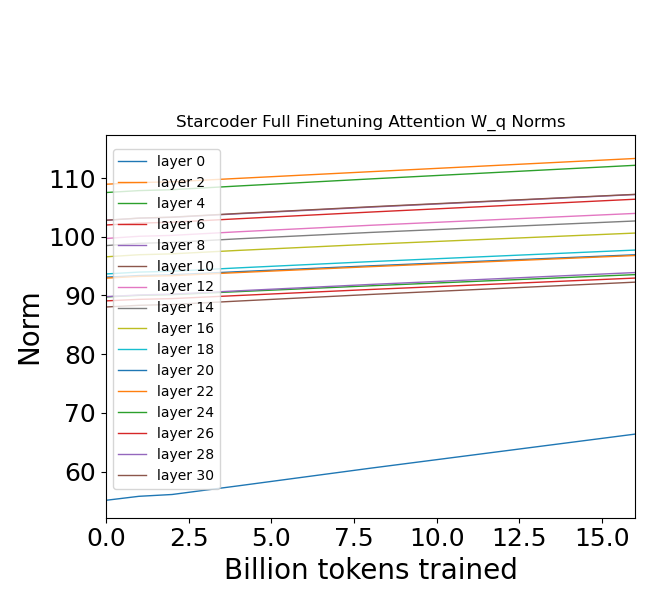}
        \caption{CPT Attention-Query}
    \end{subfigure}
    \begin{subfigure}{0.22\textwidth}
        \includegraphics[width=\linewidth]{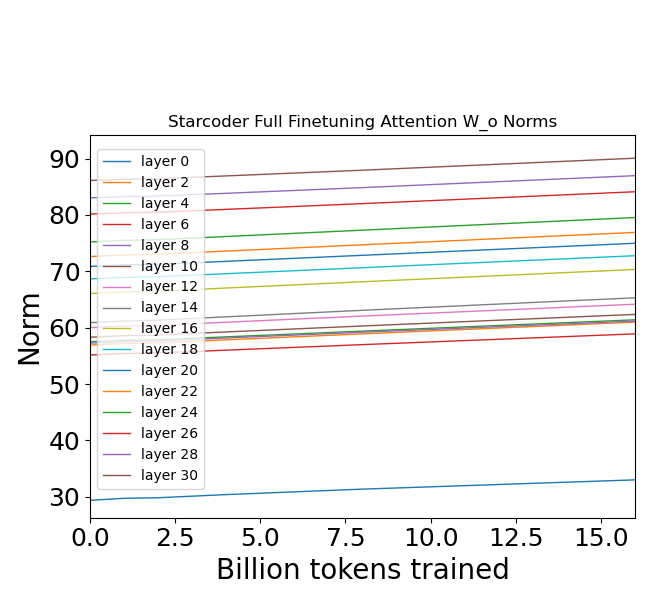}
        \caption{CPT Attention-Combo}
    \end{subfigure}

    \vspace{0.5cm} 

    \begin{subfigure}{0.22\textwidth}
        \includegraphics[width=\linewidth]{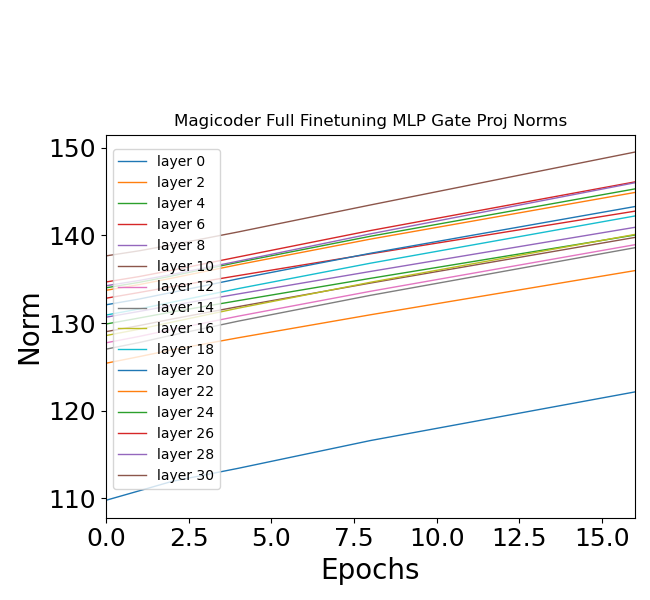}
        \caption{FFT MLP-Gate}
    \end{subfigure}
    \begin{subfigure}{0.22\textwidth}
        \includegraphics[width=\linewidth]{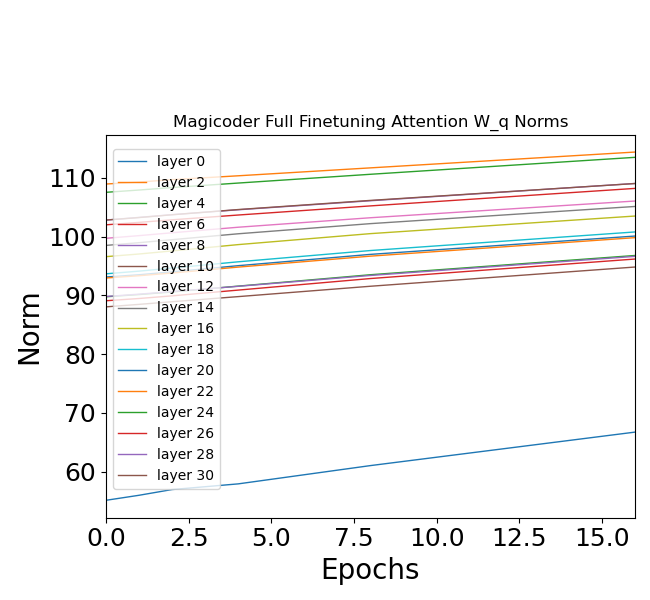}
        \caption{FFT Attention-Query}
    \end{subfigure}
    \begin{subfigure}{0.22\textwidth}
        \includegraphics[width=\linewidth]{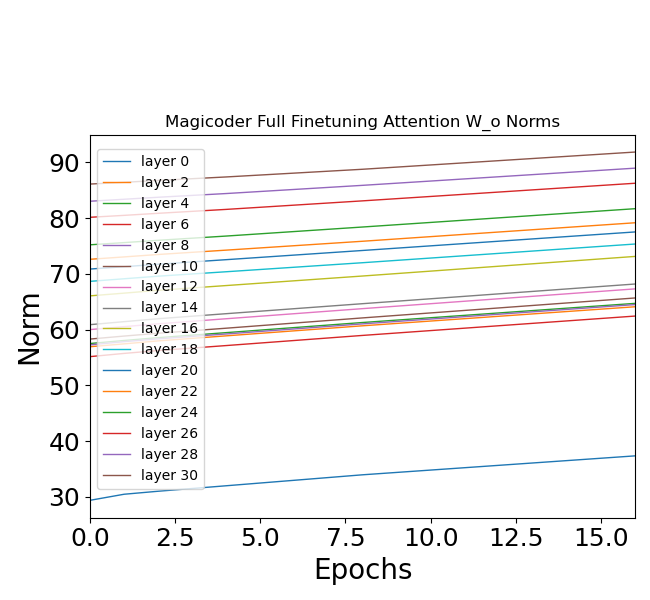}
        \caption{FFT Attention-Combo}
    \end{subfigure}

    \vspace{0.5cm} 

    \begin{subfigure}{0.22\textwidth}
        \includegraphics[width=\linewidth]{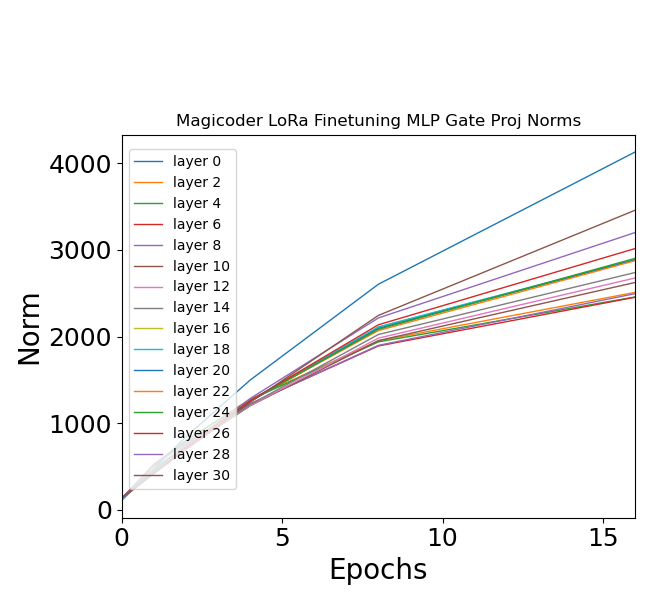}
        \caption{LFFT MLP-Gate}
    \end{subfigure}
    \begin{subfigure}{0.22\textwidth}
        \includegraphics[width=\linewidth]{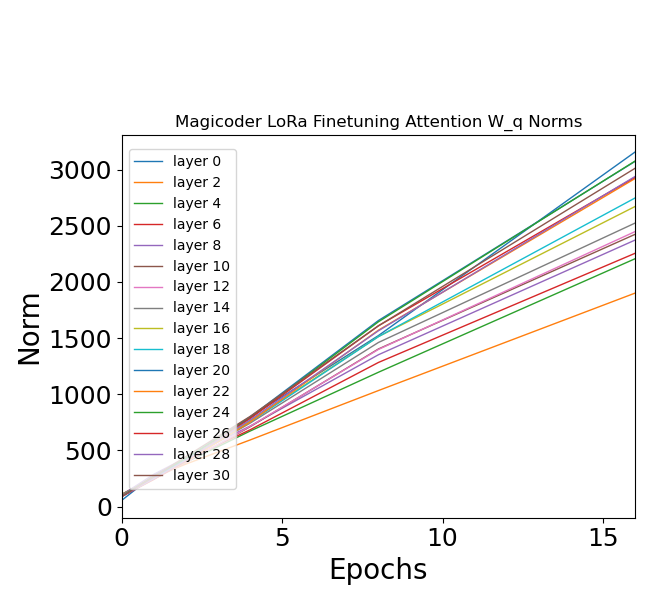}
        \caption{LFFT Attention-Query}
    \end{subfigure}
    \begin{subfigure}{0.22\textwidth}
        \includegraphics[width=\linewidth]{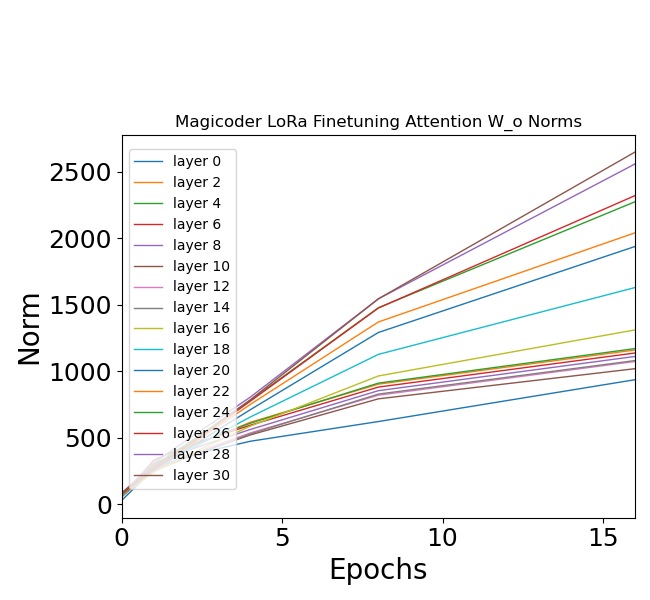}
        \caption{LFFT Attention-Combo}
    \end{subfigure}

    \caption{Additional figures for post-training interventions.}
    \label{fig:post-training-appendix}
\end{figure*}

\begin{figure*}[h]
    \centering
    \begin{subfigure}{0.22\textwidth}
        \includegraphics[width=\linewidth]{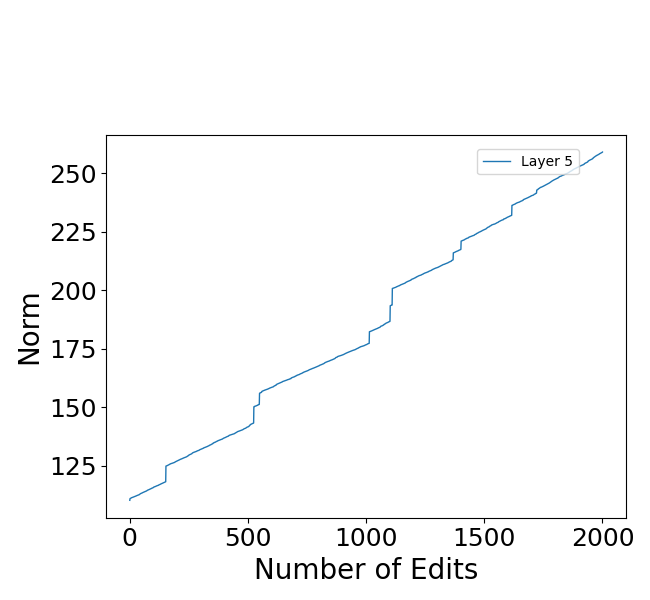}
        \caption{ROME-norm}
    \end{subfigure}
    \begin{subfigure}{0.22\textwidth}
        \includegraphics[width=\linewidth]{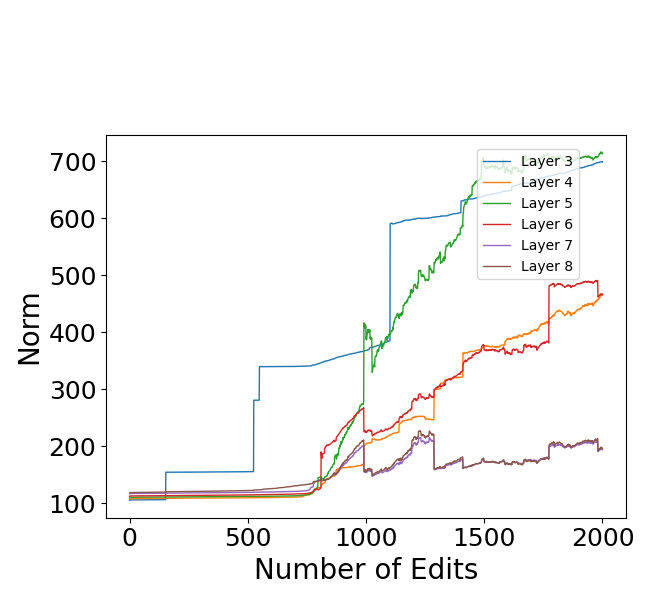}
        \caption{MEMIT-norm}
    \end{subfigure}
    \begin{subfigure}{0.22\textwidth}
        \includegraphics[width=\linewidth]{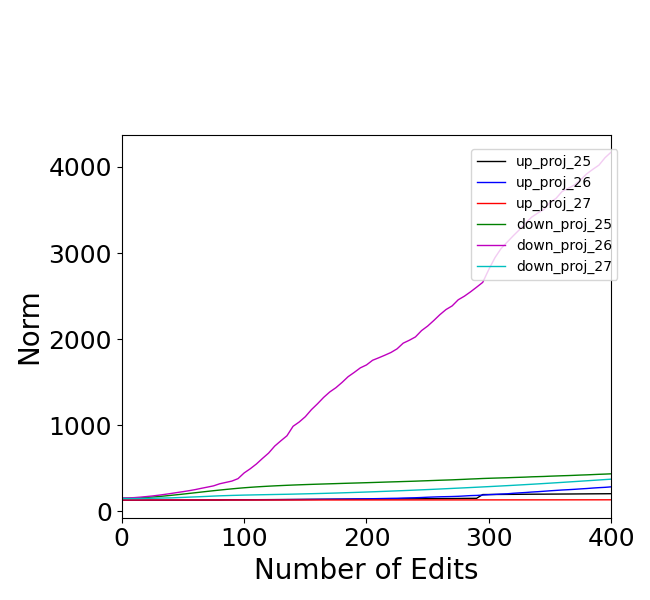}
        \caption{MEND-norm}
    \end{subfigure}
        \begin{subfigure}{0.22\textwidth}
        \includegraphics[width=\linewidth]{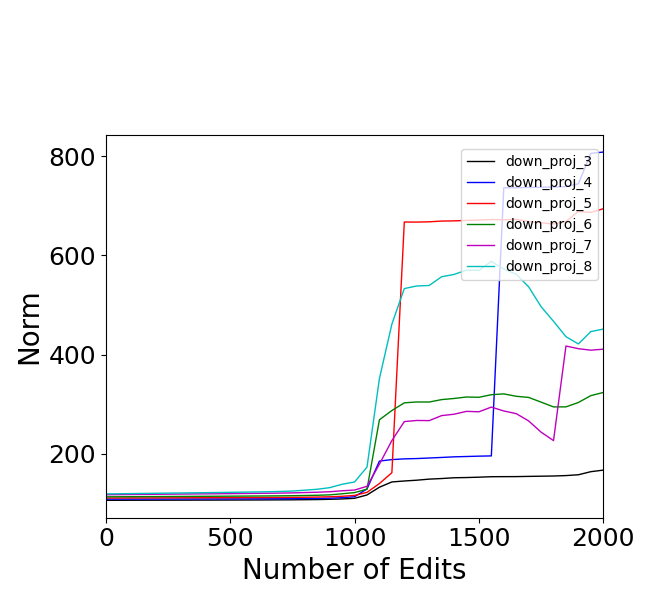}
        \caption{PMET-norm}
    \end{subfigure}

    \vspace{0.5cm} 

    \begin{subfigure}{0.22\textwidth}
        \includegraphics[width=\linewidth]{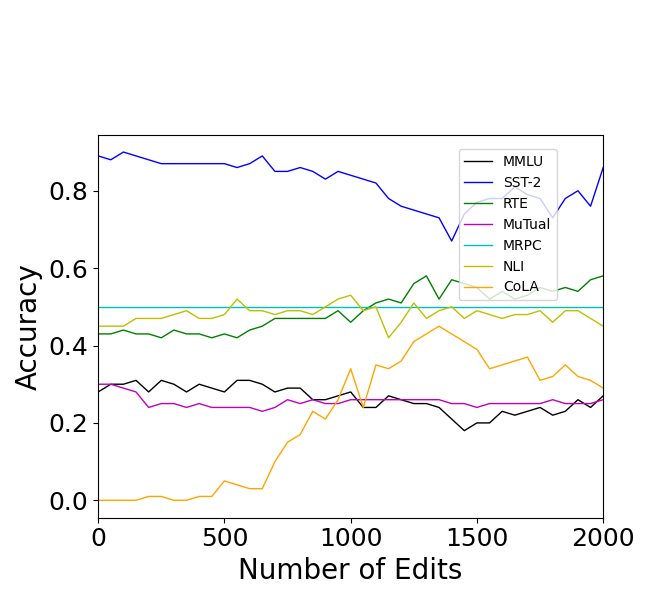}
        \caption{ROME-downstream}
    \end{subfigure}
    \begin{subfigure}{0.22\textwidth}
        \includegraphics[width=\linewidth]{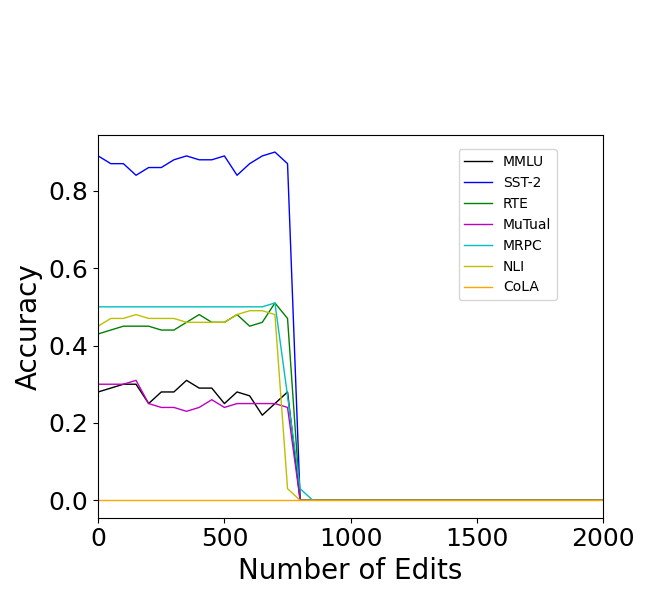}
        \caption{MEMIT-downstream}
    \end{subfigure}
    \begin{subfigure}{0.22\textwidth}
        \includegraphics[width=\linewidth]{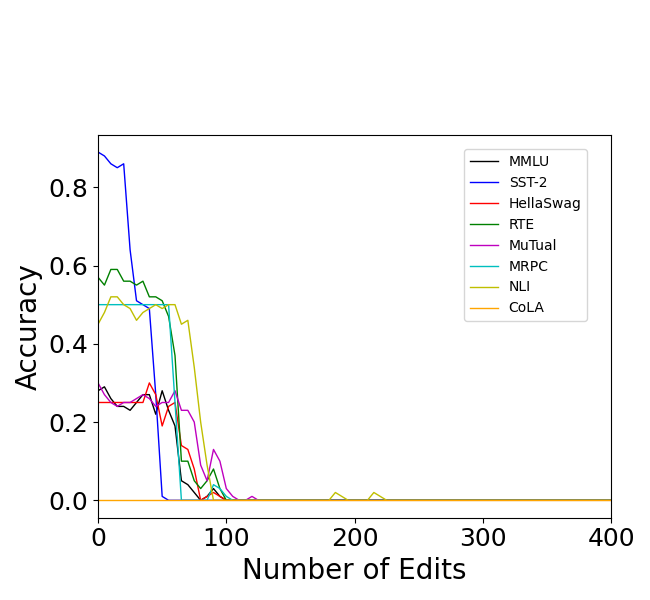}
        \caption{MEND-downstream}
    \end{subfigure}
        \begin{subfigure}{0.22\textwidth}
        \includegraphics[width=\linewidth]{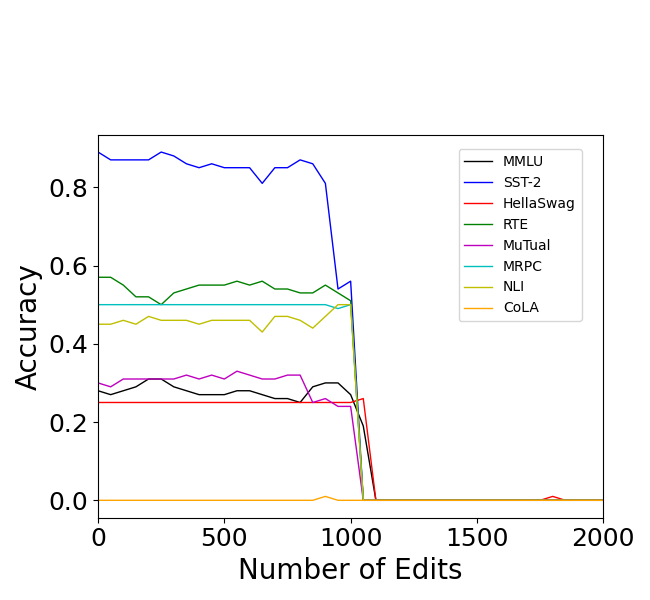}
        \caption{PMET-downstream}
    \end{subfigure}

    \vspace{0.5cm} 

    \begin{subfigure}{0.22\textwidth}
        \includegraphics[width=\linewidth]{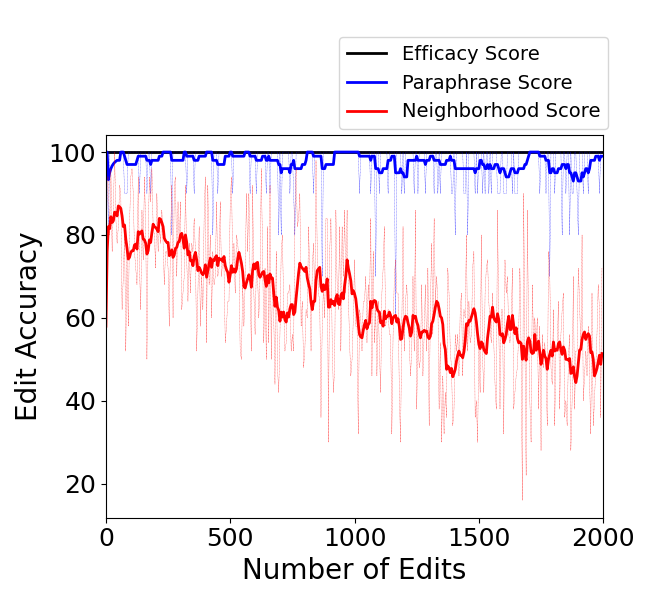}
        \caption{ROME-editing}
    \end{subfigure}
    \begin{subfigure}{0.22\textwidth}
        \includegraphics[width=\linewidth]{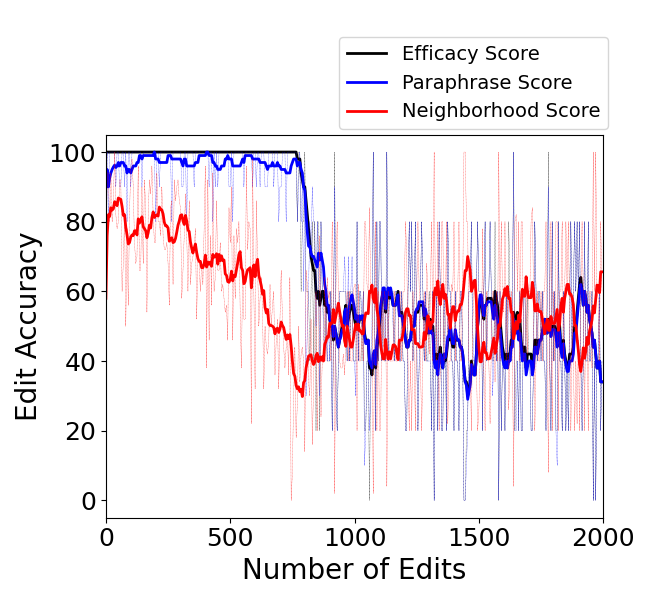}
        \caption{MEMIT-editing}
    \end{subfigure}
    \begin{subfigure}{0.22\textwidth}
        \includegraphics[width=\linewidth]{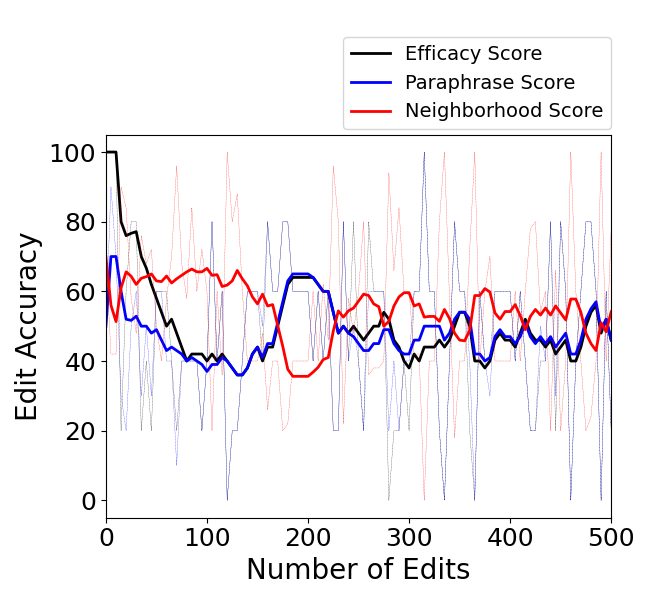}
        \caption{MEND-editing}
    \end{subfigure}
        \begin{subfigure}{0.22\textwidth}
        \includegraphics[width=\linewidth]{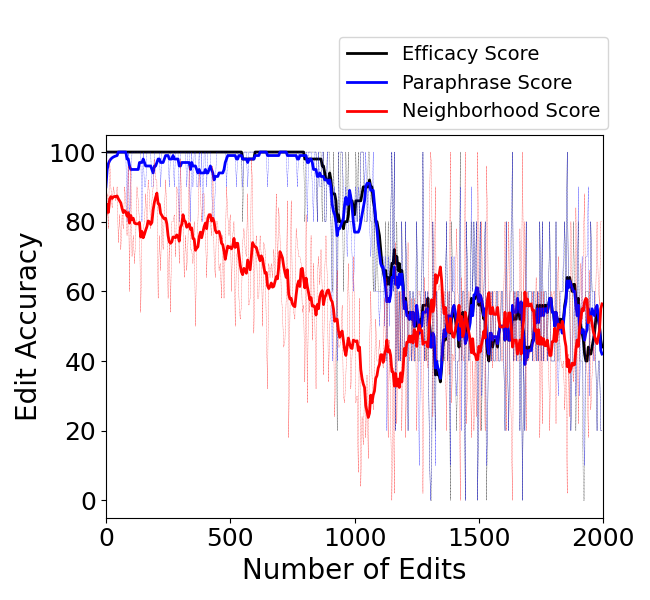}
        \caption{PMET-editing}
    \end{subfigure}

    \caption{Norm growth during knowledge editing on GPT-J.}
    \label{fig:editing-appendix}
\end{figure*}

\begin{figure*}[h]
    \centering
    \begin{subfigure}{0.22\textwidth}
        \includegraphics[width=\linewidth]{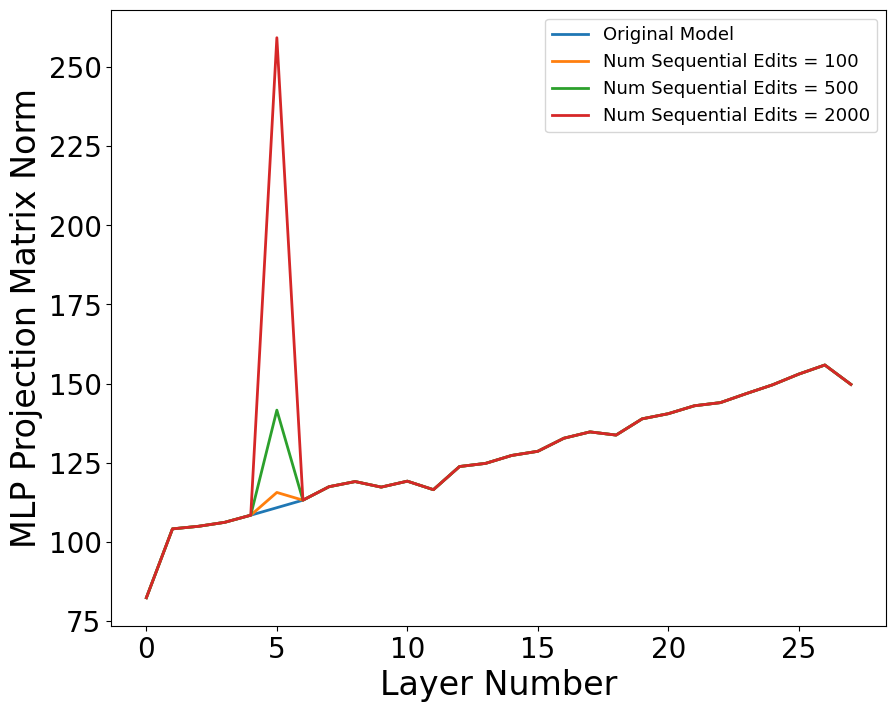}
        \caption{ROME}
    \end{subfigure}
    \begin{subfigure}{0.22\textwidth}
        \includegraphics[width=\linewidth]{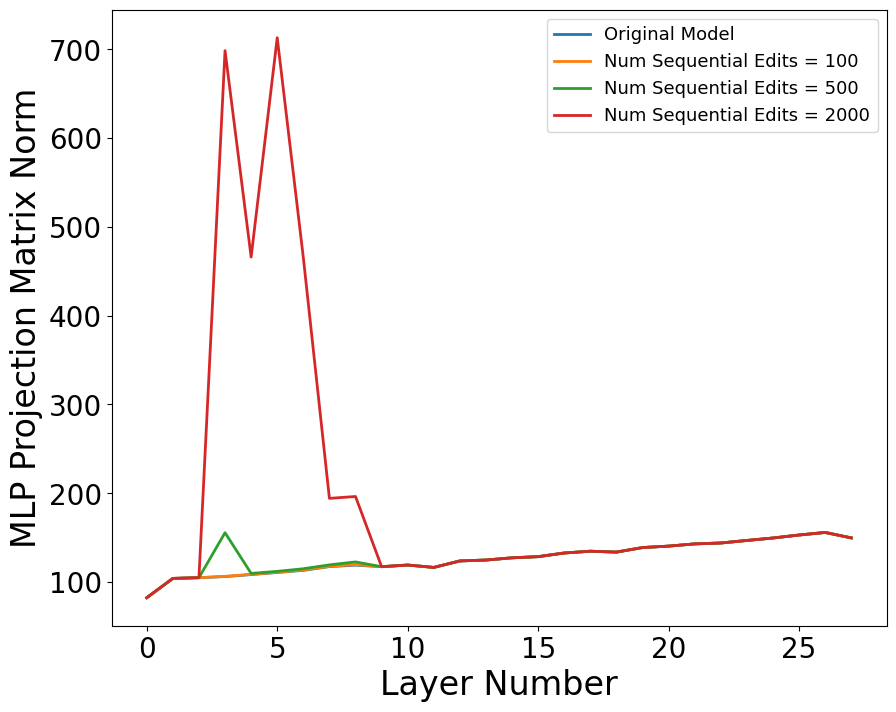}
        \caption{MEMIT}
    \end{subfigure}
    \begin{subfigure}{0.22\textwidth}
        \includegraphics[width=\linewidth]{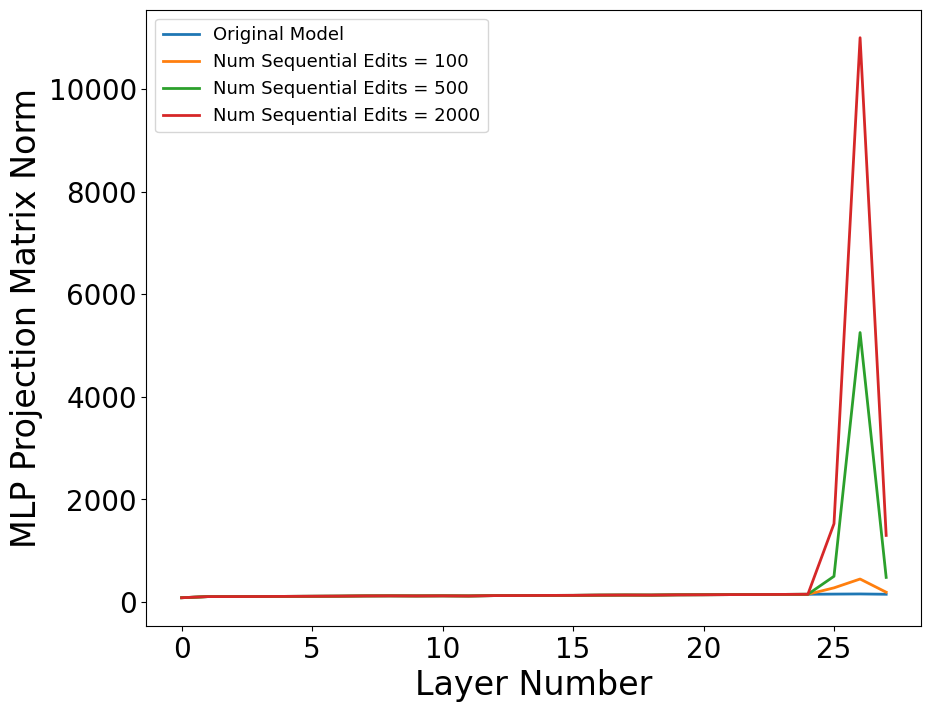}
        \caption{MEND}
    \end{subfigure}
        \begin{subfigure}{0.22\textwidth}
        \includegraphics[width=\linewidth]{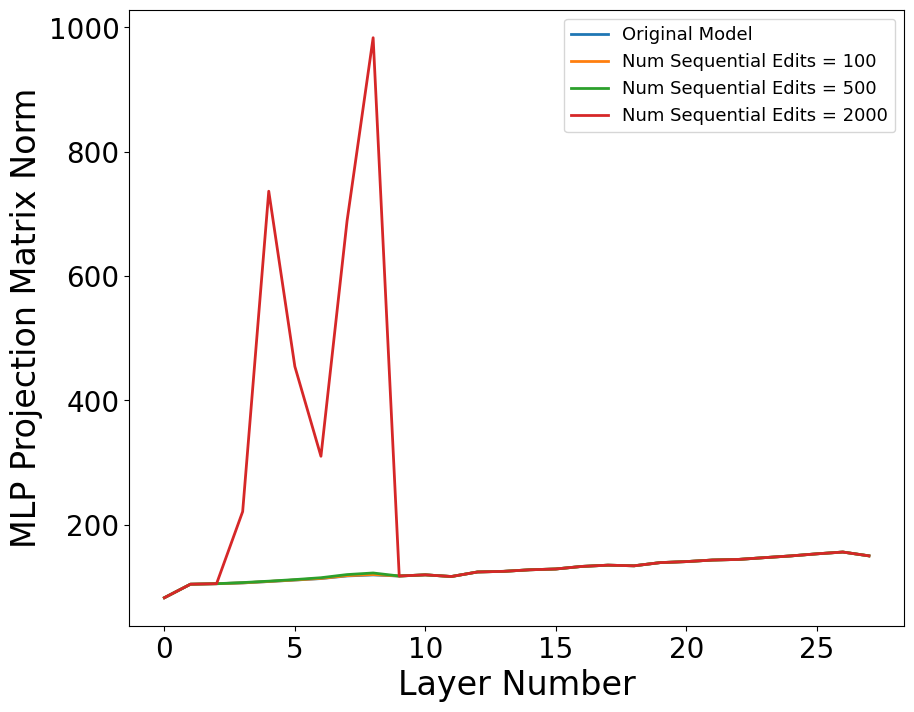}
        \caption{PMET}
    \end{subfigure}

    \vspace{0.5cm} 

    \caption{Norm growth for edits 100, 500, 2000 for GPT-J.}
    \label{fig:norm-growth-gptj}
\end{figure*}

\begin{figure*}[h]
    \centering
    \begin{subfigure}{0.22\textwidth}
        \includegraphics[width=\linewidth]{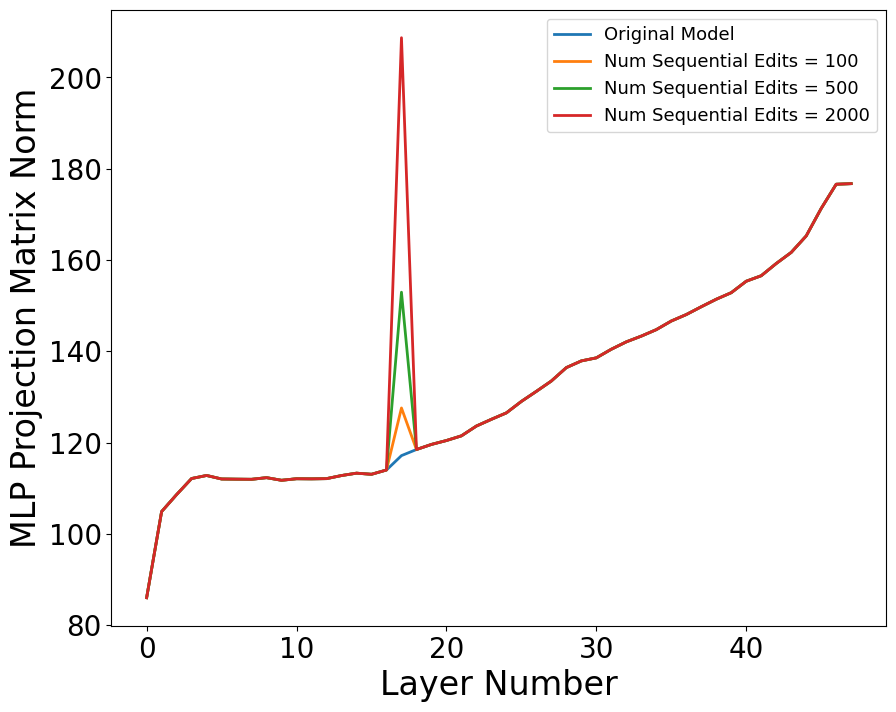}
        \caption{GPT2-XL/FT}
    \end{subfigure}
    \begin{subfigure}{0.22\textwidth}
        \includegraphics[width=\linewidth]{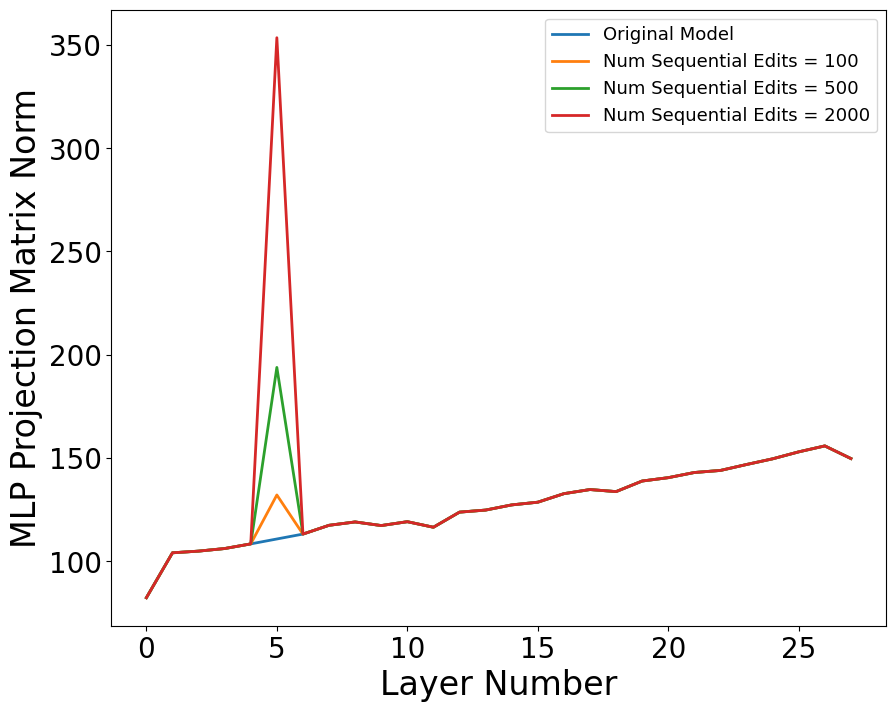}
        \caption{GPT-J/FT}
    \end{subfigure}
    \vspace{0.5cm} 

    \caption{Norm growth for edits 100, 500, 2000 for FT editing.}
    \label{fig:norm-growth-ft}
\end{figure*}

\begin{figure*}[h]
    \centering
    \begin{subfigure}{0.32\textwidth}
        \includegraphics[width=\linewidth]{figures/boundary_plots/boundary_gpt2xl_unedited.png}
        \caption{GPT2-XL}
    \end{subfigure}\
    \begin{subfigure}{0.32\textwidth}
        \includegraphics[width=\linewidth]{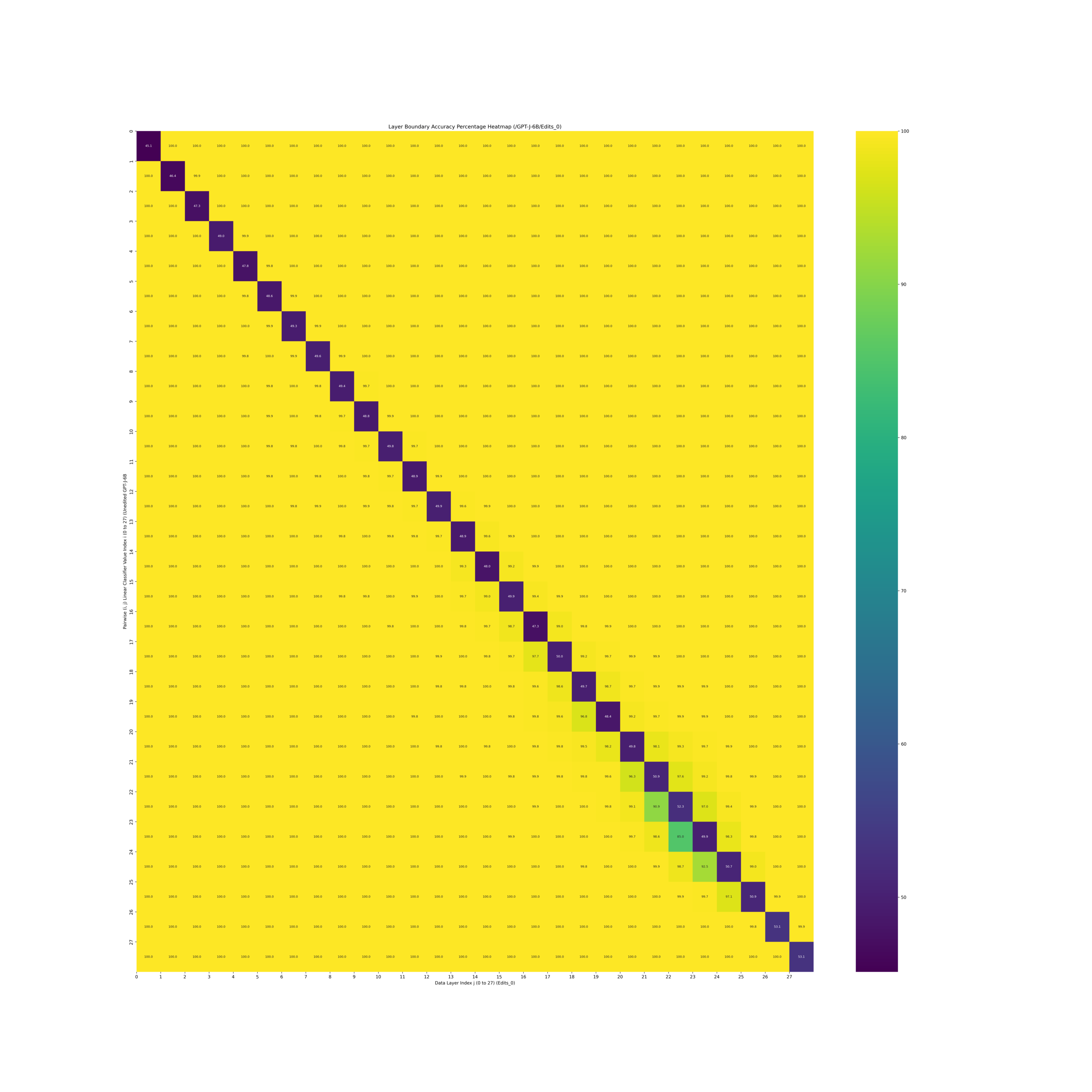}
        \caption{GPT-J}
    \end{subfigure}

    \vspace{0.5cm} 

    \caption{Boundary plots for unedited GPT2-XL and GPT-J.}
    \label{fig:boundary-unedited}
\end{figure*}

\begin{figure*}[h]
    \centering
    \begin{subfigure}{0.32\textwidth}
        \includegraphics[width=\linewidth]{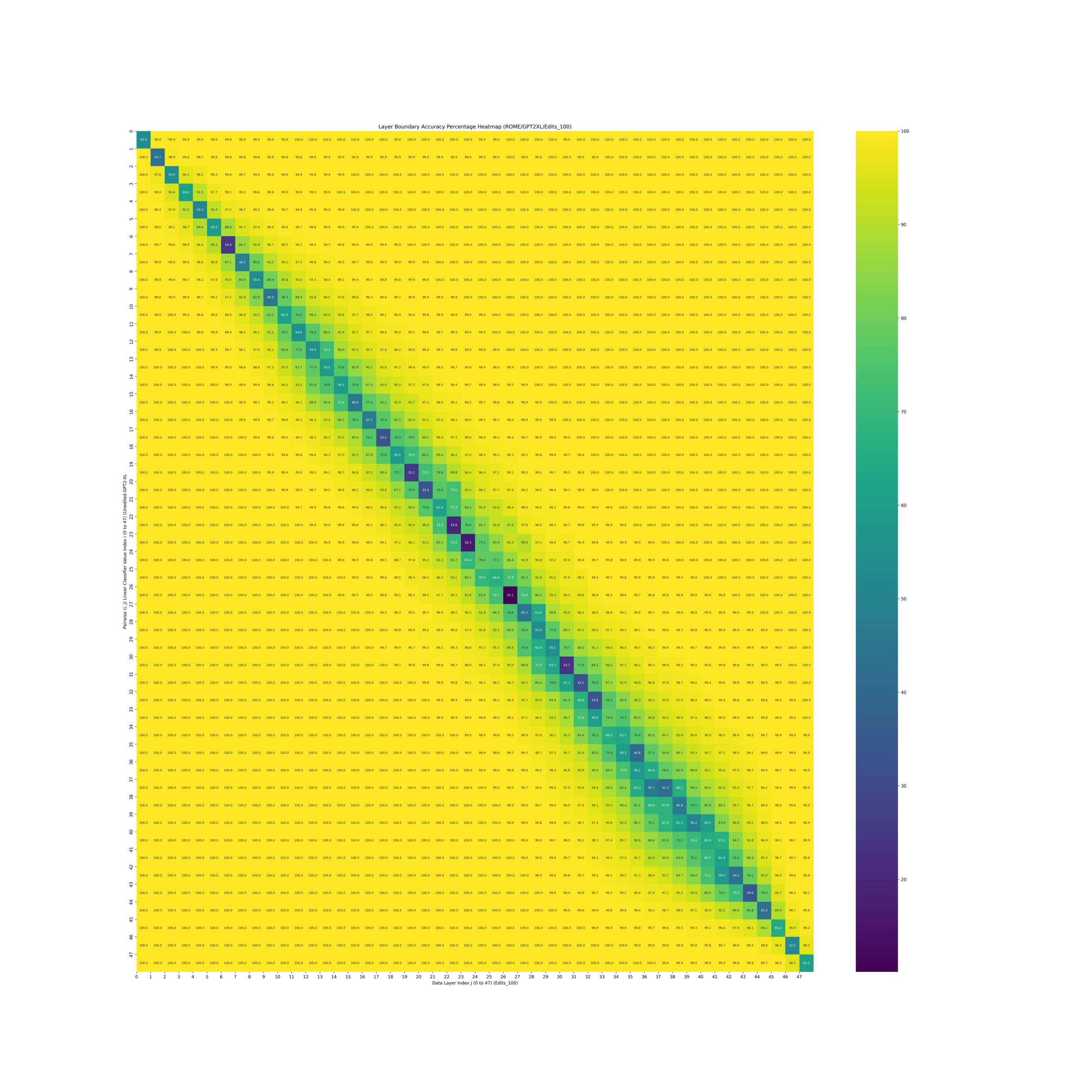}
        \caption{100}
    \end{subfigure}\
    \begin{subfigure}{0.32\textwidth}
        \includegraphics[width=\linewidth]{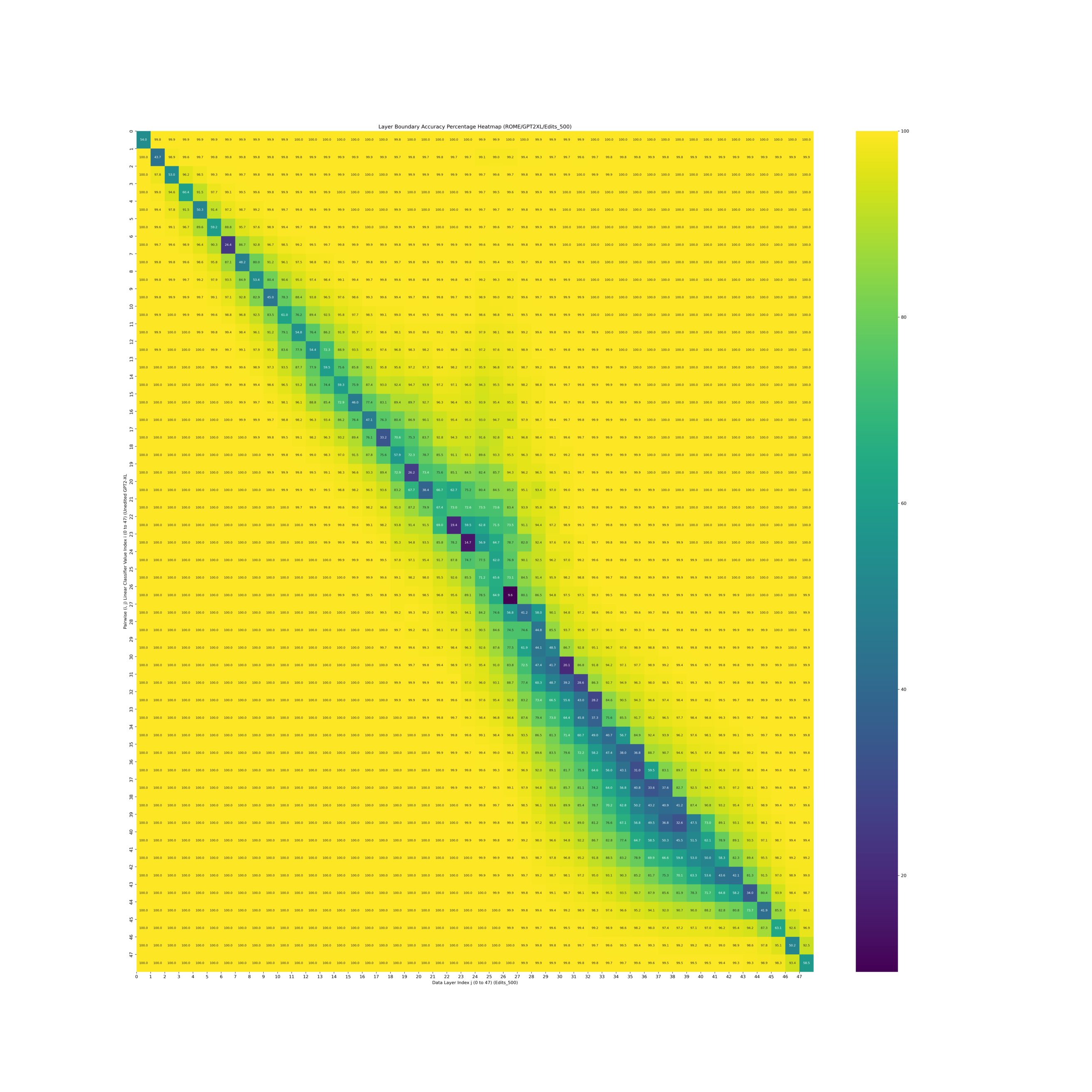}
        \caption{500}
    \end{subfigure}
    \begin{subfigure}{0.32\textwidth}
        \includegraphics[width=\linewidth]{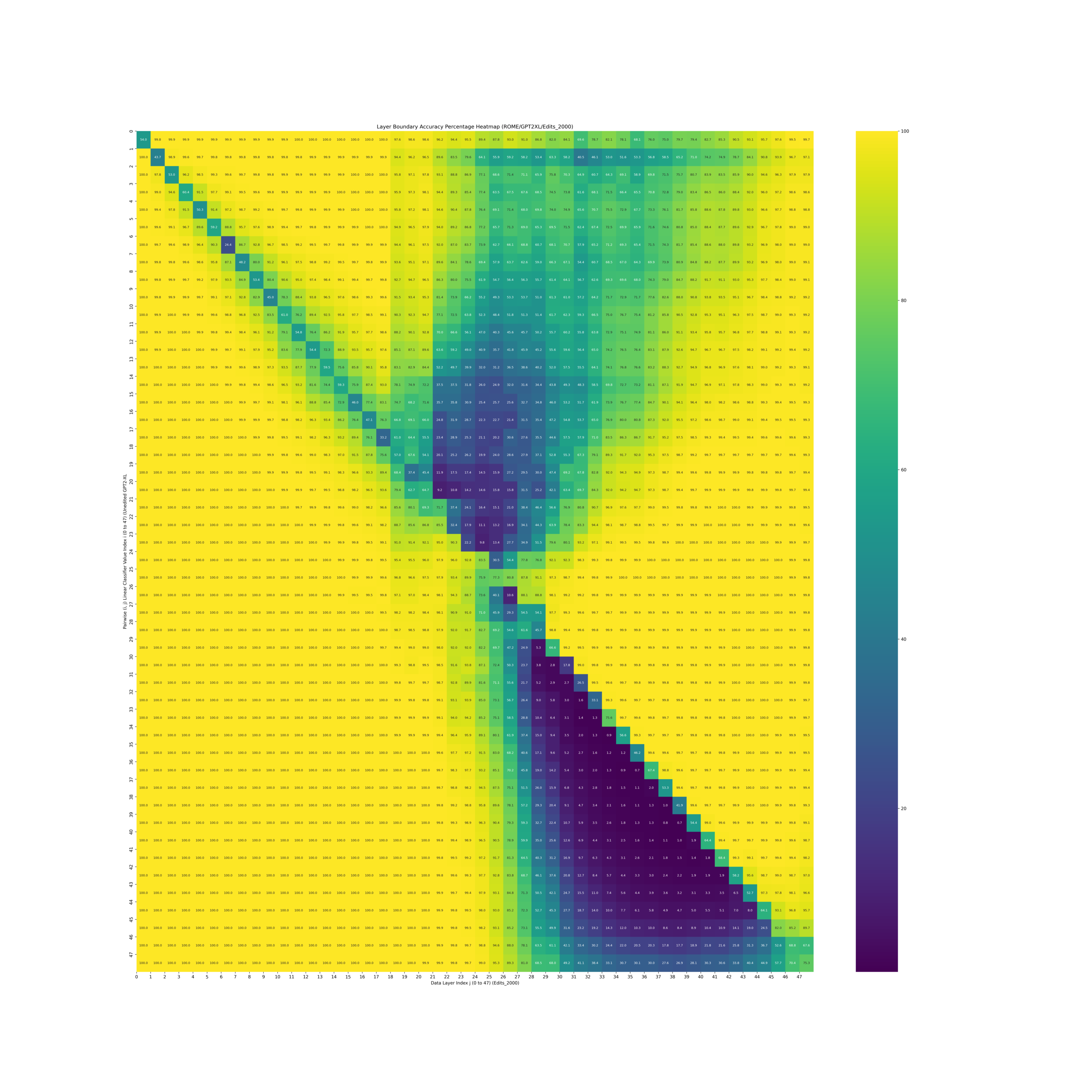}
        \caption{2000}
    \end{subfigure}
       
    \vspace{0.5cm} 

    \caption{Boundary plots for edits 100, 500, 2000 for GPT2-XL/ROME.}
    \label{fig:boundary-GPT2XL-ROME}
\end{figure*}

\begin{figure*}[h]
    \centering
    \begin{subfigure}{0.32\textwidth}
        \includegraphics[width=\linewidth]{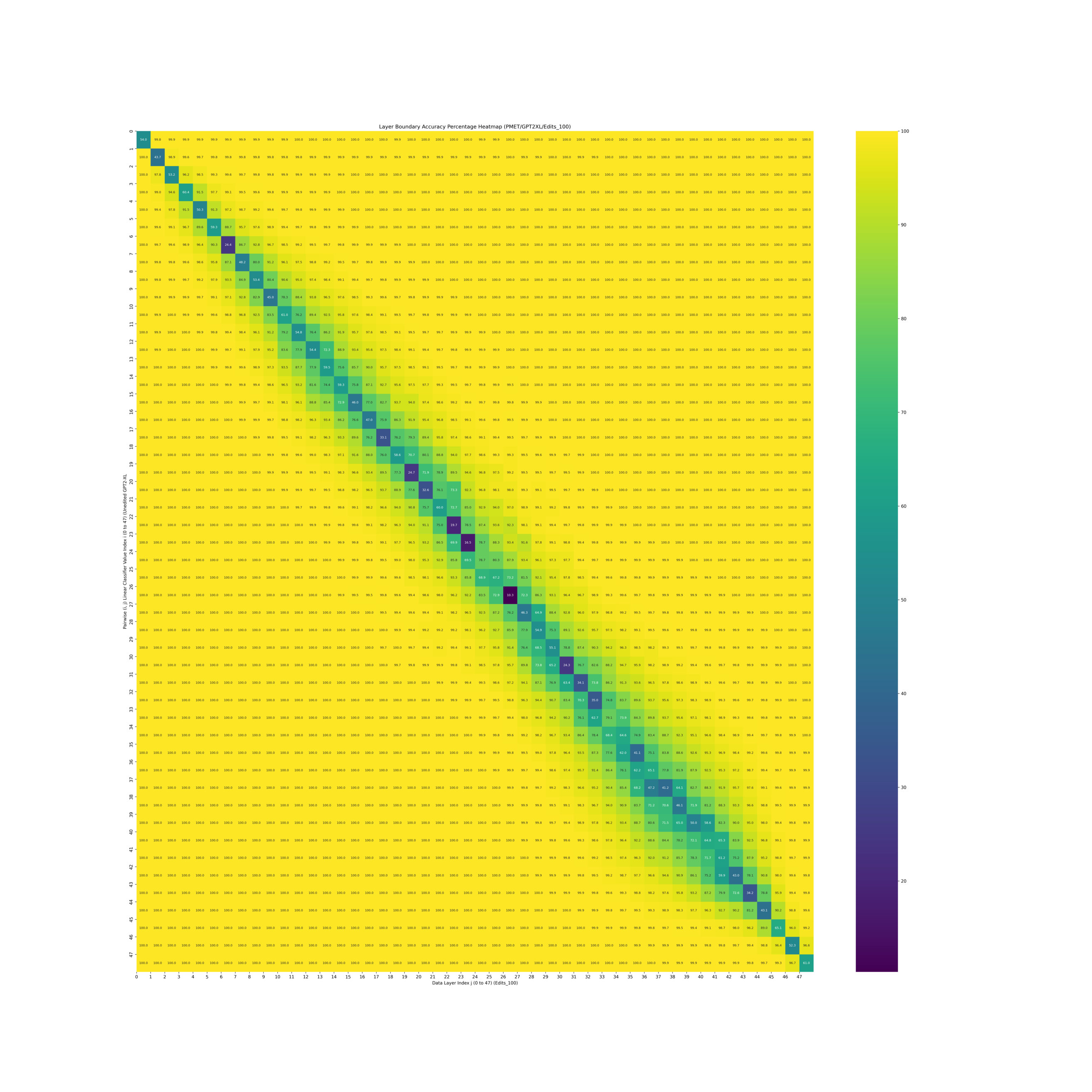}
        \caption{100}
    \end{subfigure}
    \begin{subfigure}{0.32\textwidth}
        \includegraphics[width=\linewidth]{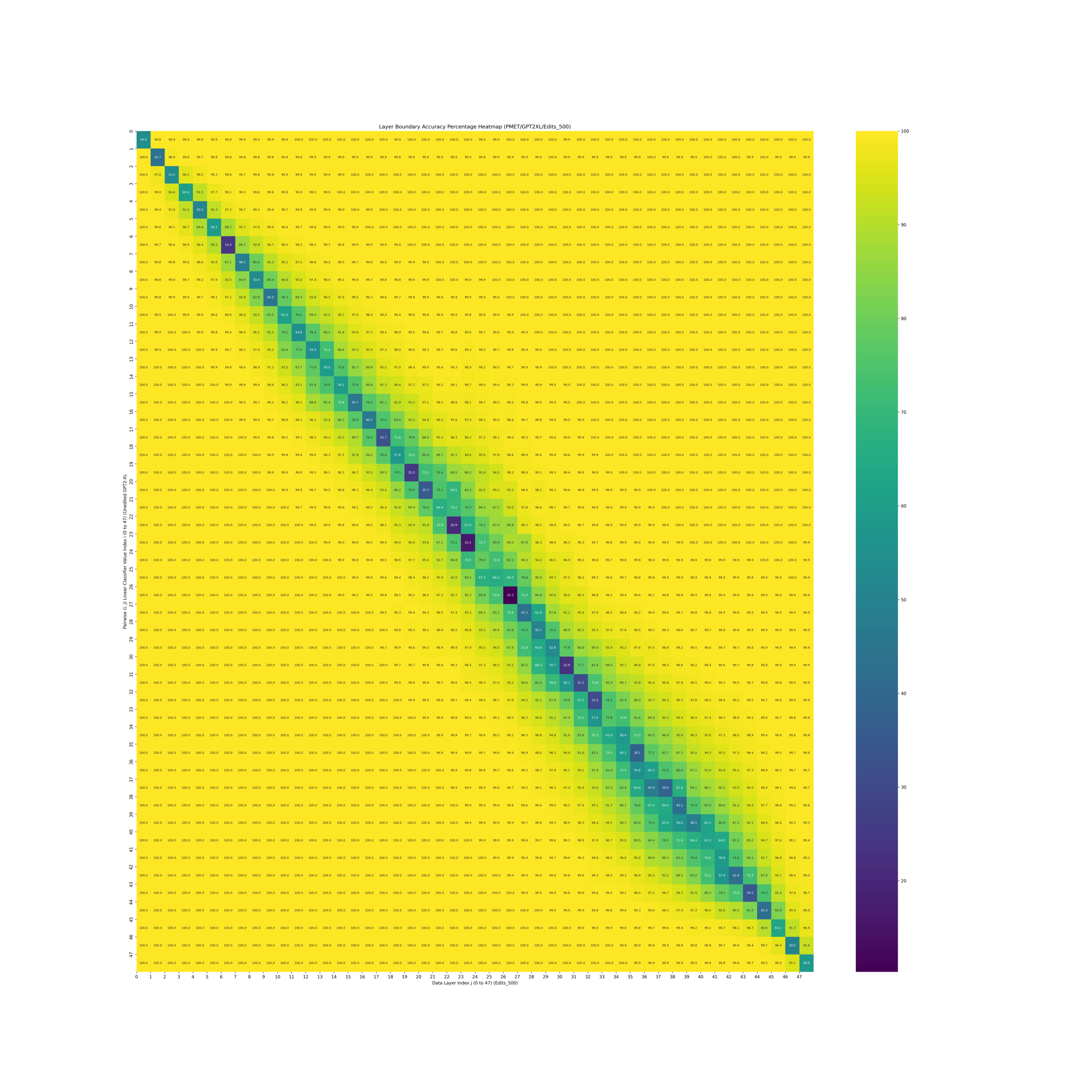}
        \caption{500}
    \end{subfigure}
    \begin{subfigure}{0.32\textwidth}
        \includegraphics[width=\linewidth]{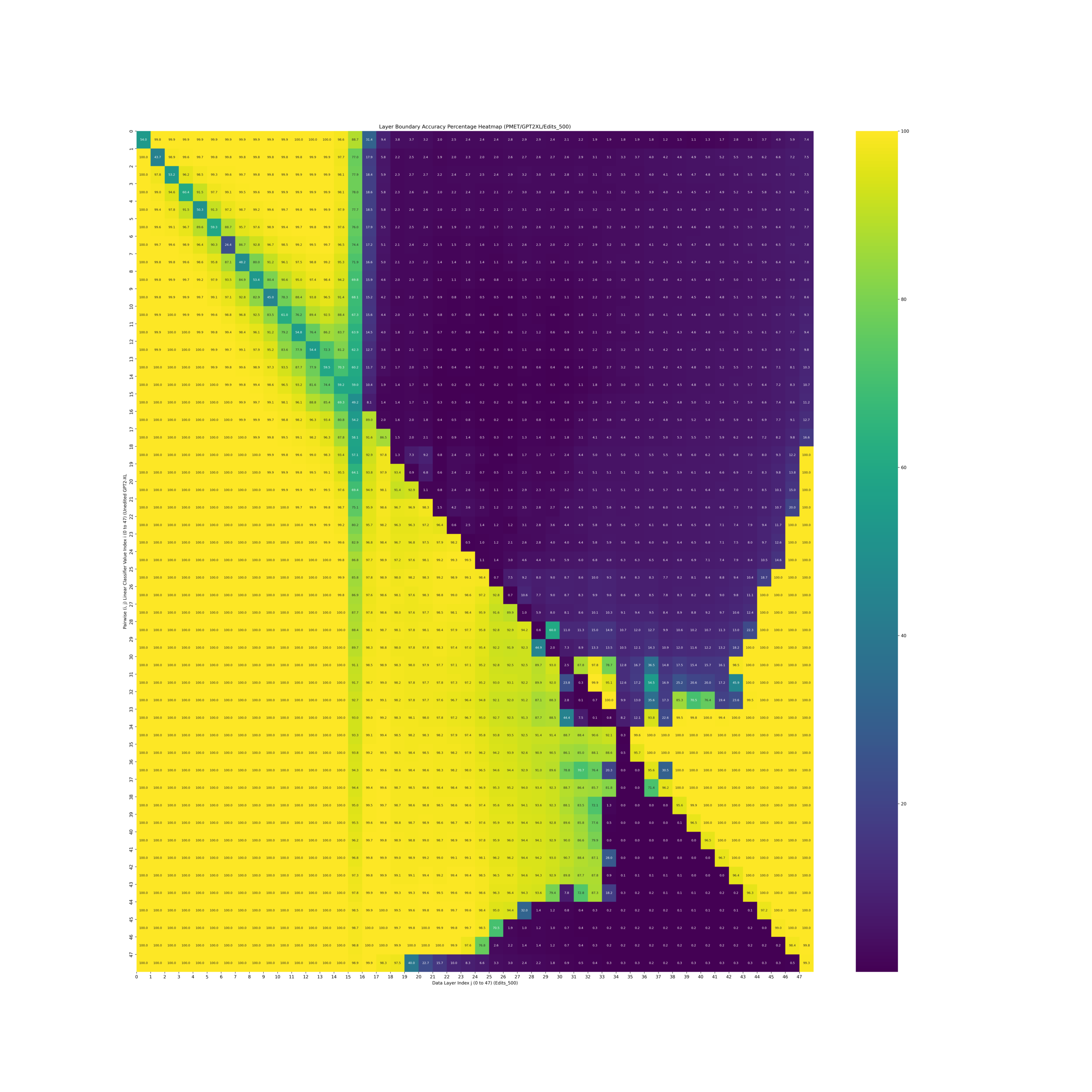}
        \caption{2000}
    \end{subfigure}
       
    \vspace{0.5cm} 

    \caption{Boundary plots for edits 100, 500, 2000 for GPT2-XL/PMET.}
    \label{fig:boundary-GPT2XL-PMET}
\end{figure*}

\begin{figure*}[h]
    \centering
    \begin{subfigure}{0.32\textwidth}
        \includegraphics[width=\linewidth]{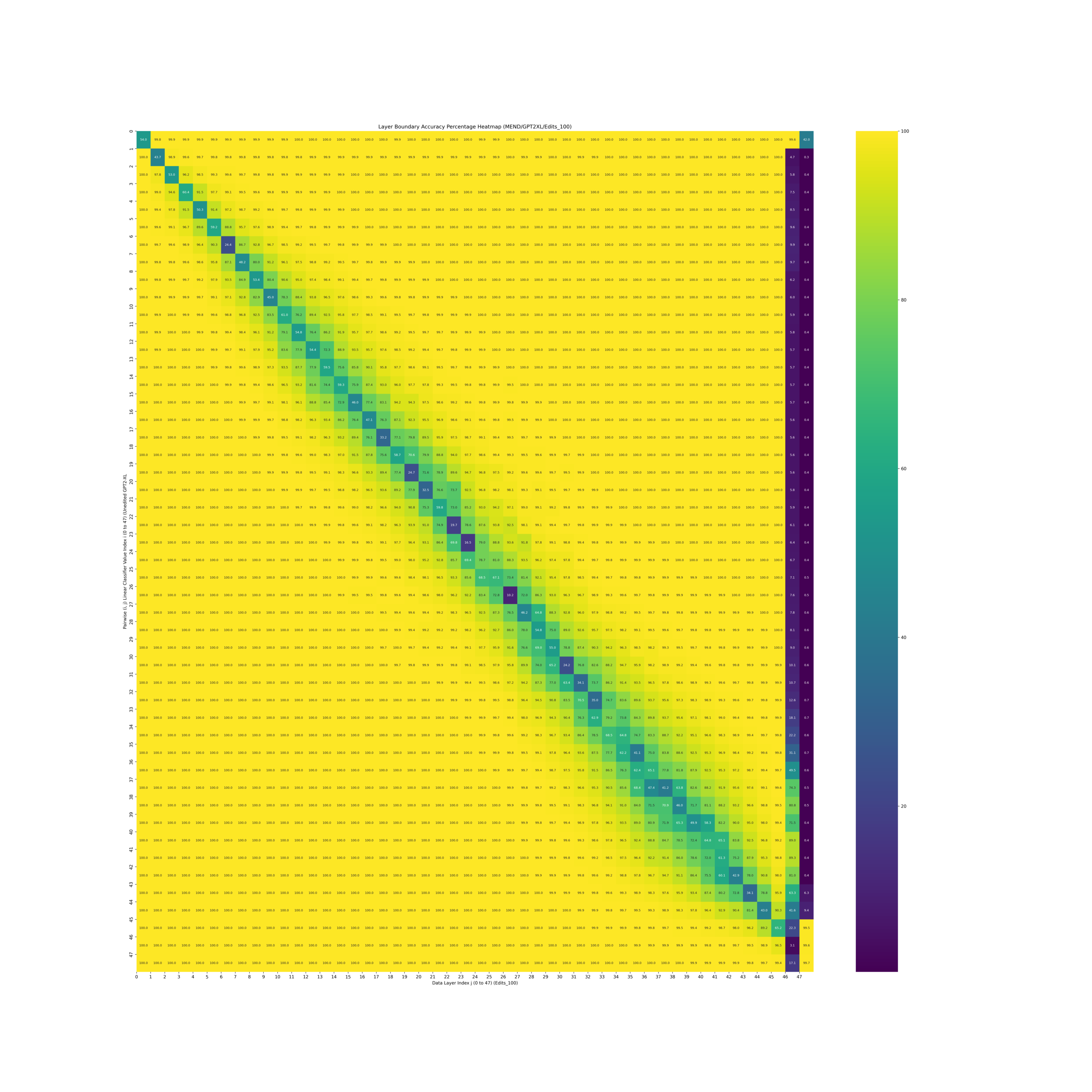}
        \caption{100}
    \end{subfigure}
    \begin{subfigure}{0.32\textwidth}
        \includegraphics[width=\linewidth]{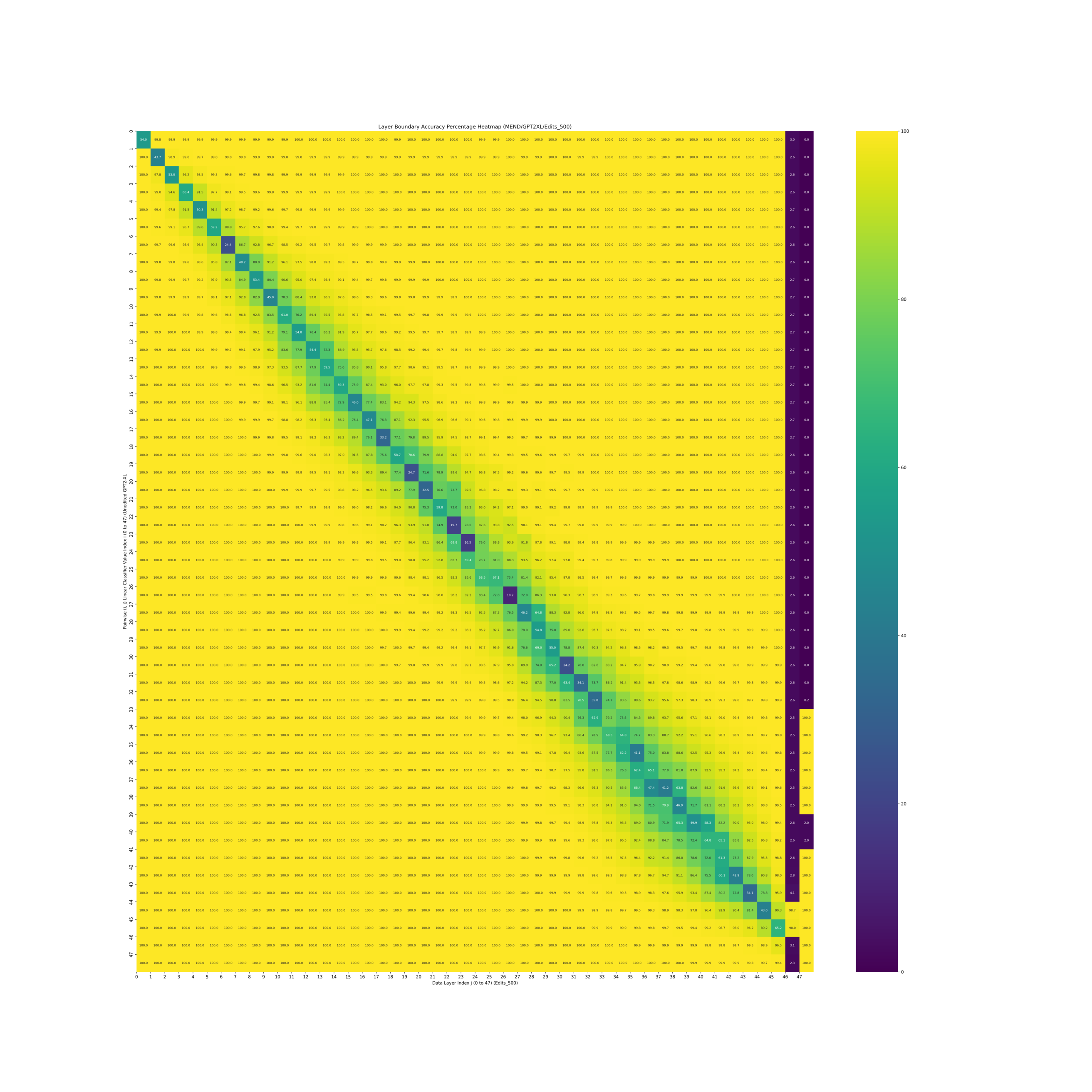}
        \caption{500}
    \end{subfigure}
    \begin{subfigure}{0.32\textwidth}
        \includegraphics[width=\linewidth]{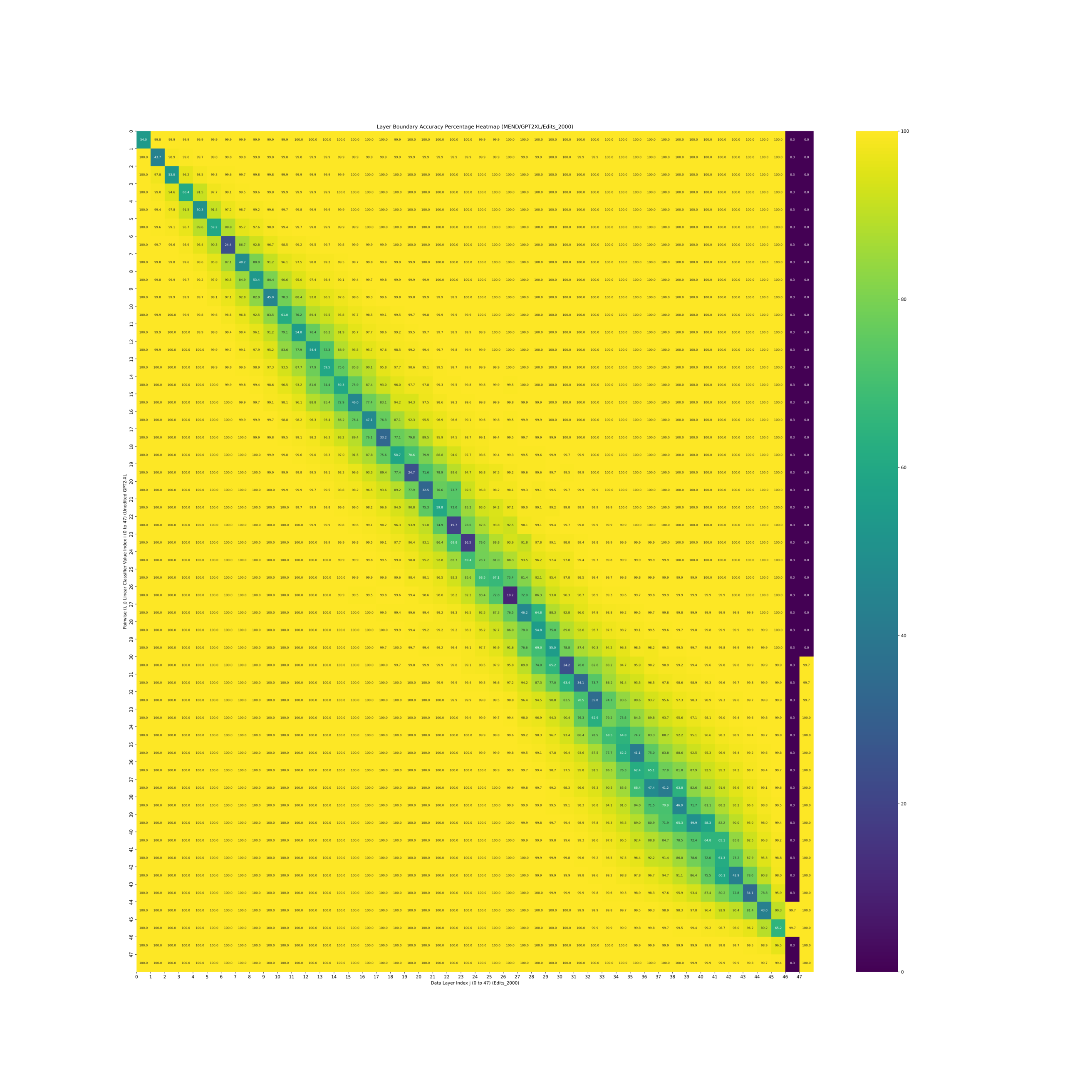}
        \caption{2000}
    \end{subfigure}
       
    \vspace{0.5cm} 

    \caption{Boundary plots for edits 100, 500, 2000 for GPT2-XL/MEND.}
    \label{fig:boundary-GPT2XL-MEND}
\end{figure*}

\begin{figure*}[h]
    \centering
    \begin{subfigure}{0.32\textwidth}
        \includegraphics[width=\linewidth]{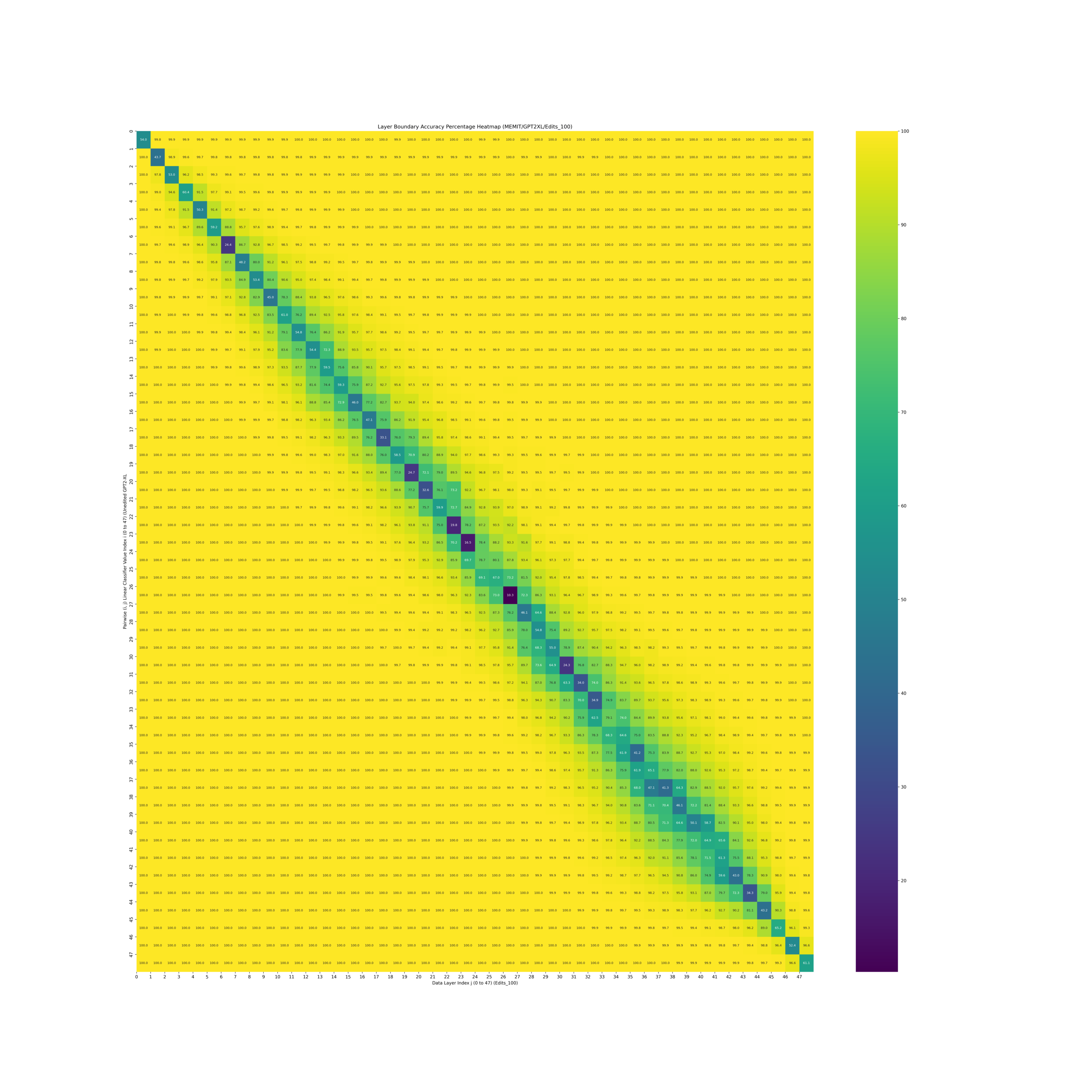}
        \caption{100}
    \end{subfigure}
    \begin{subfigure}{0.32\textwidth}
        \includegraphics[width=\linewidth]{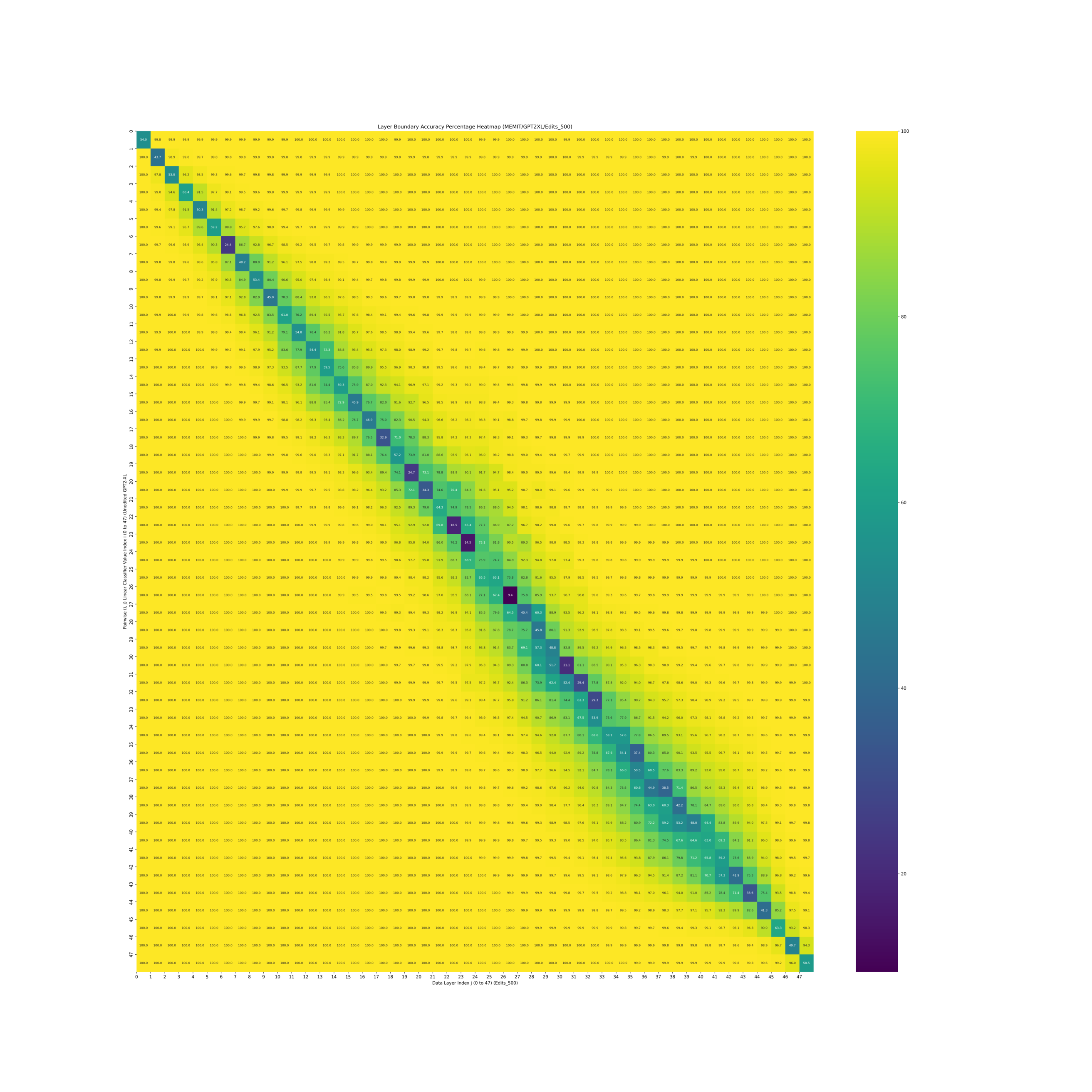}
        \caption{500}
    \end{subfigure}
    \begin{subfigure}{0.32\textwidth}
        \includegraphics[width=\linewidth]{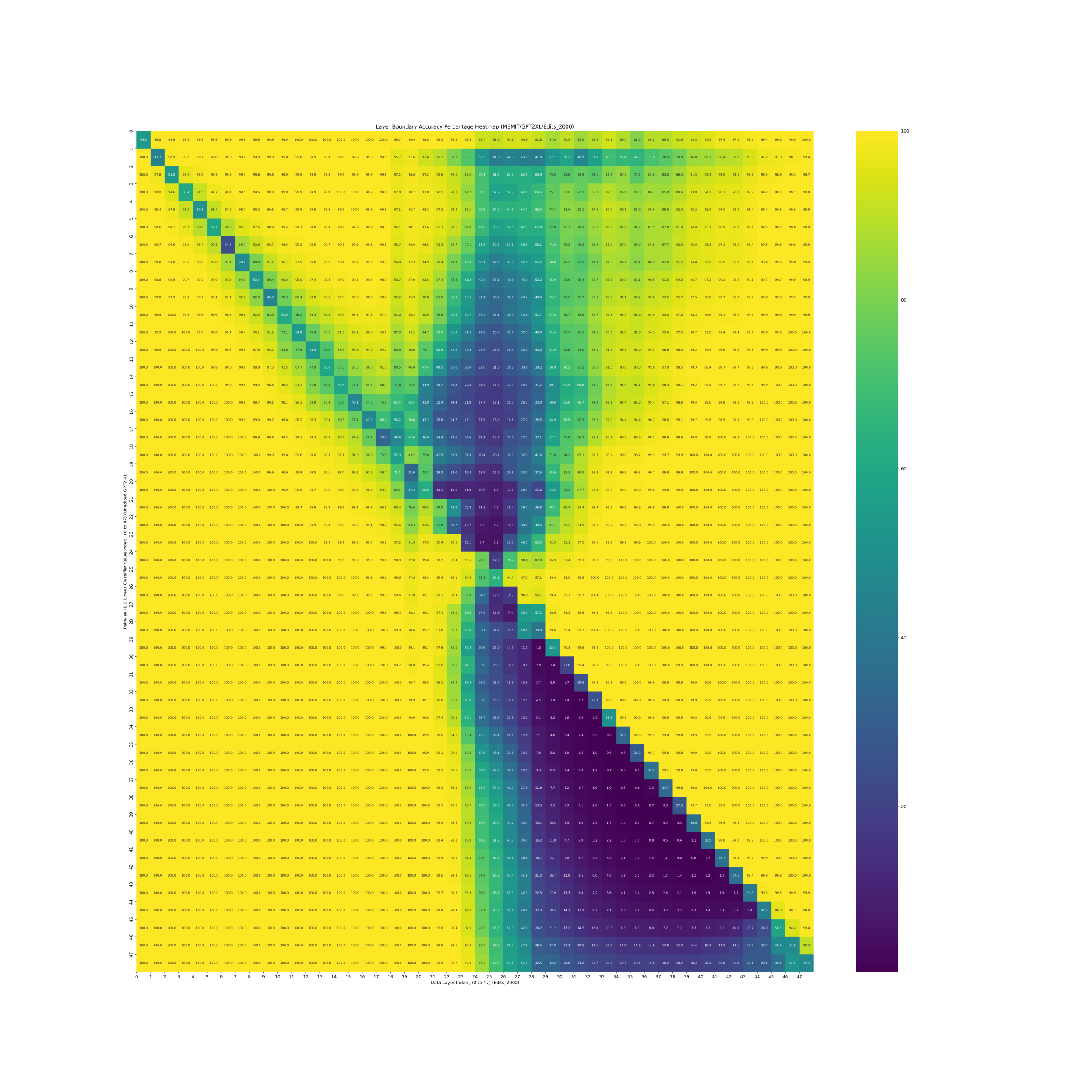}
        \caption{2000}
    \end{subfigure}
       
    \vspace{0.5cm} 

    \caption{Boundary plots for edits 100, 500, 2000 for GPT2-XL/MEMIT.}
    \label{fig:boundary-GPT2XL-MEMIT}
\end{figure*}

\begin{figure*}[h]
    \centering
    \begin{subfigure}{0.32\textwidth}
        \includegraphics[width=\linewidth]{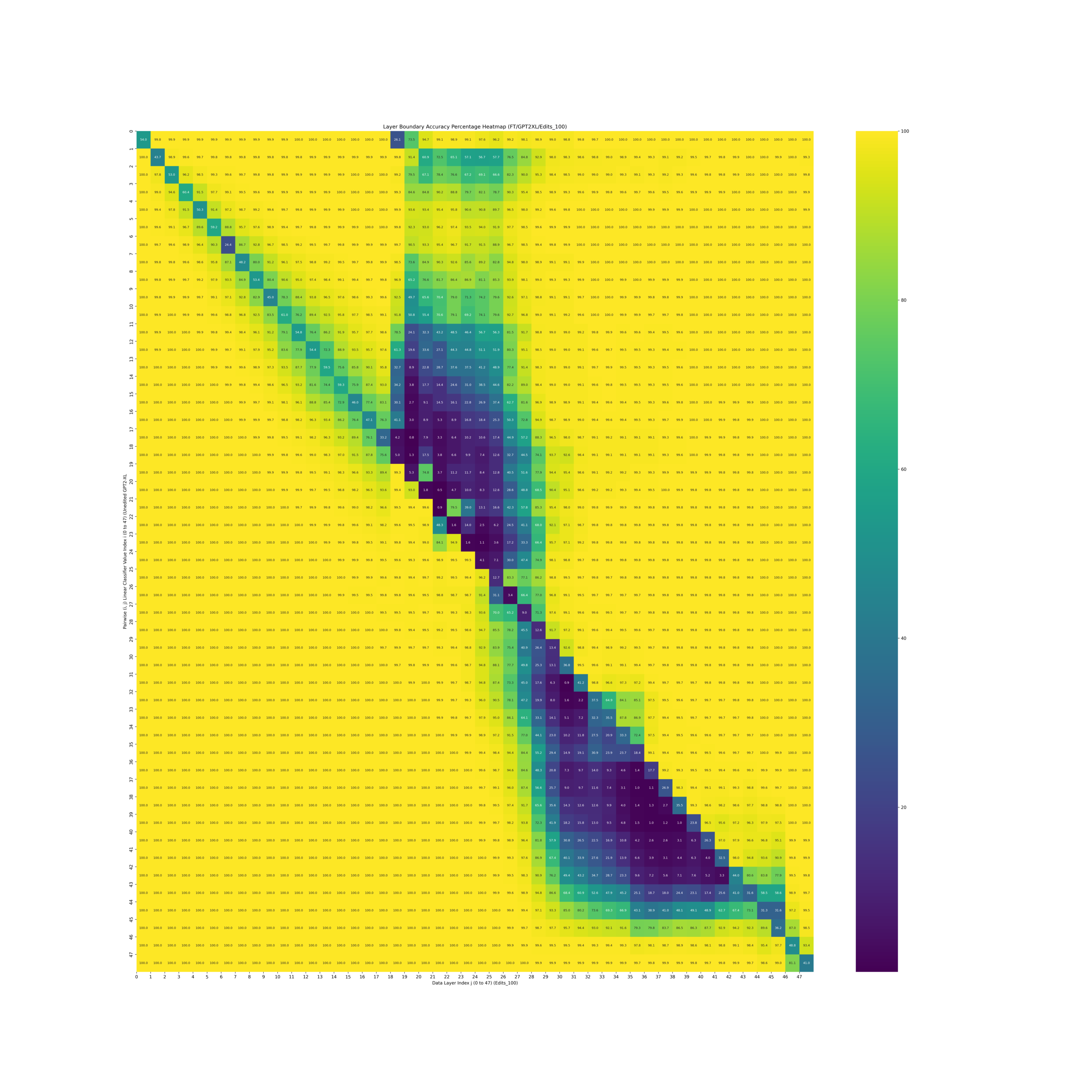}
        \caption{100}
    \end{subfigure}
    \begin{subfigure}{0.32\textwidth}
        \includegraphics[width=\linewidth]{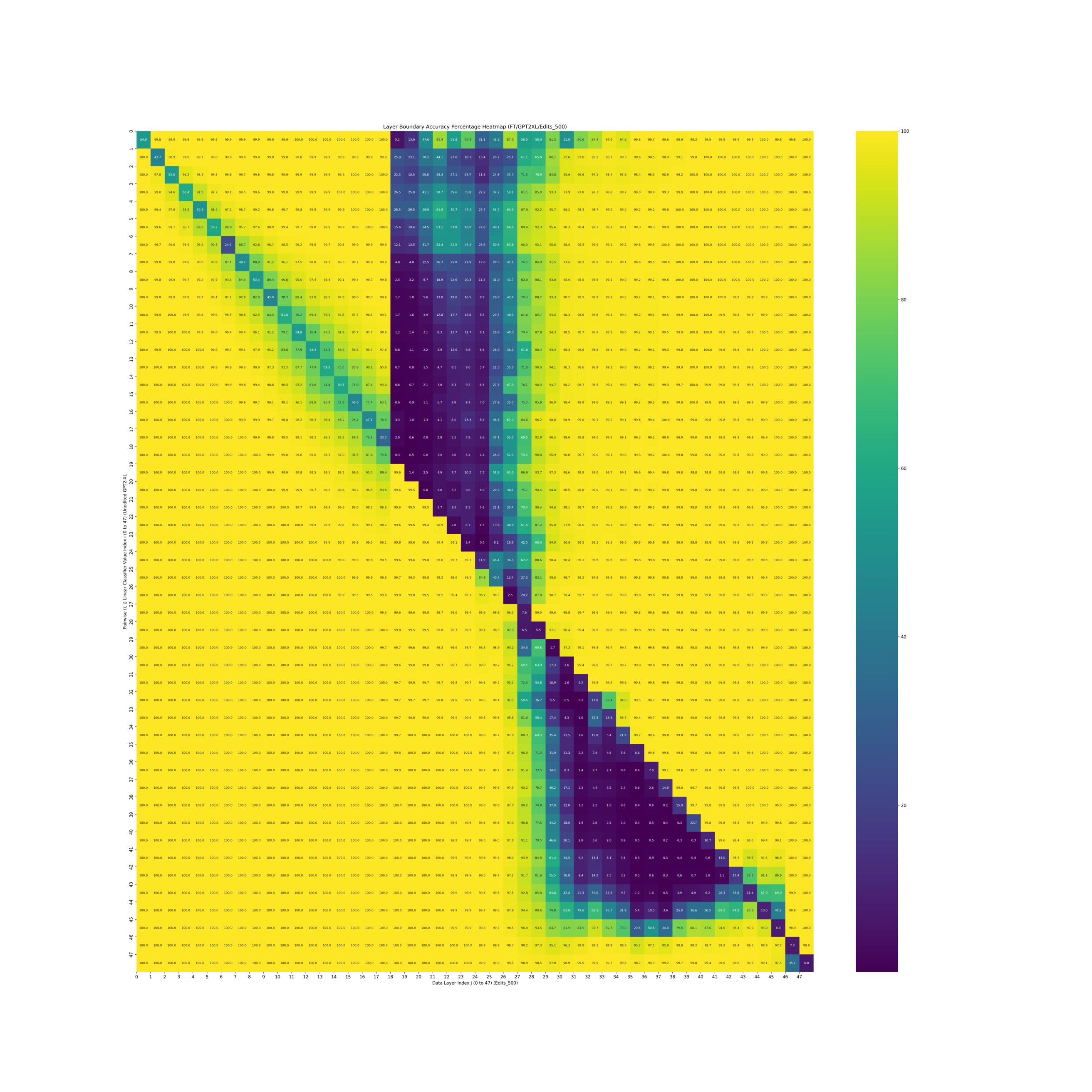}
        \caption{500}
    \end{subfigure}
    \begin{subfigure}{0.32\textwidth}
        \includegraphics[width=\linewidth]{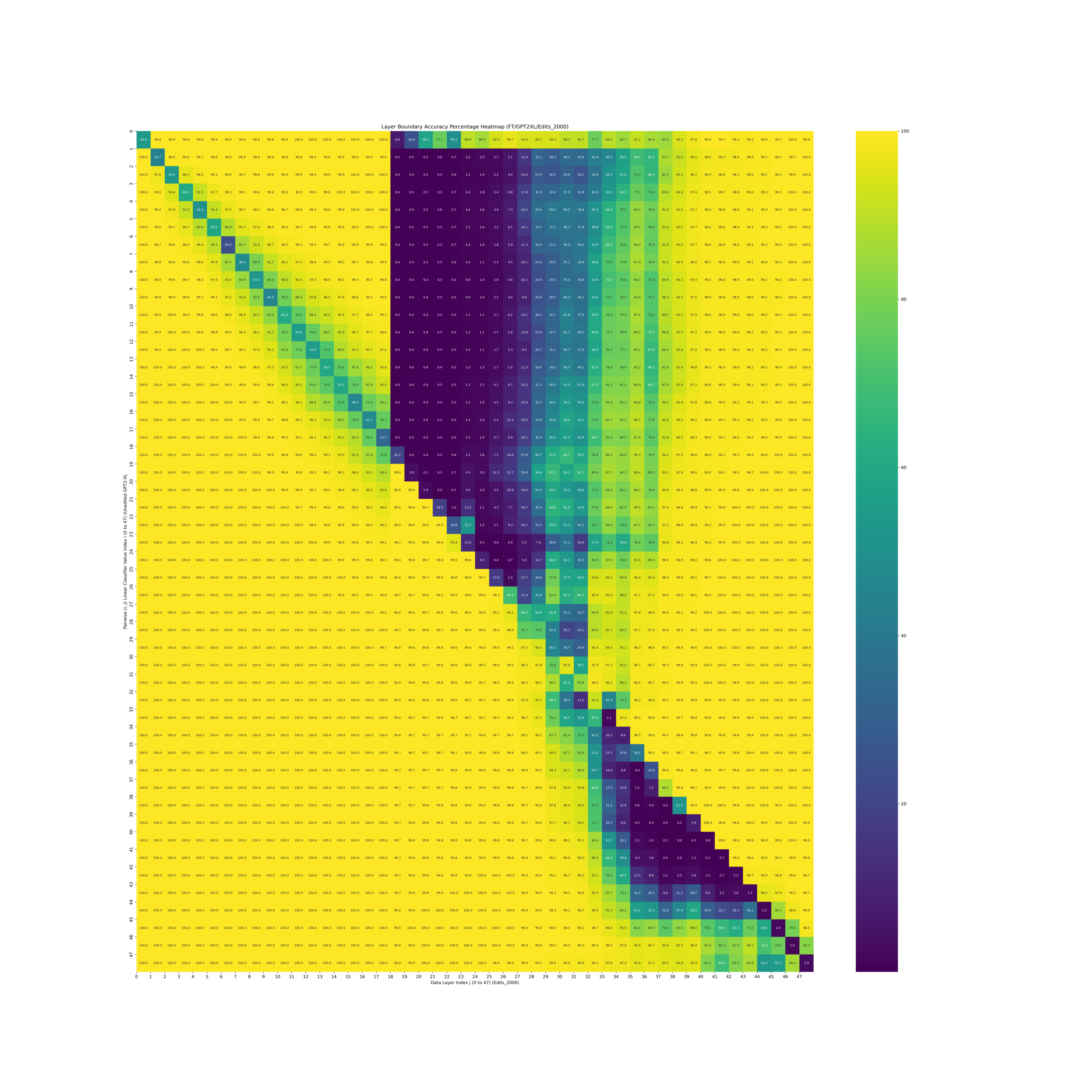}
        \caption{2000}
    \end{subfigure}
       
    \vspace{0.5cm} 

    \caption{Boundary plots for edits 100, 500, 2000 for GPT2-XL/FT.}
    \label{fig:boundary-GPT2XL-FT}
\end{figure*}

\begin{figure*}[h]
    \centering
    \begin{subfigure}{0.32\textwidth}
        \includegraphics[width=\linewidth]{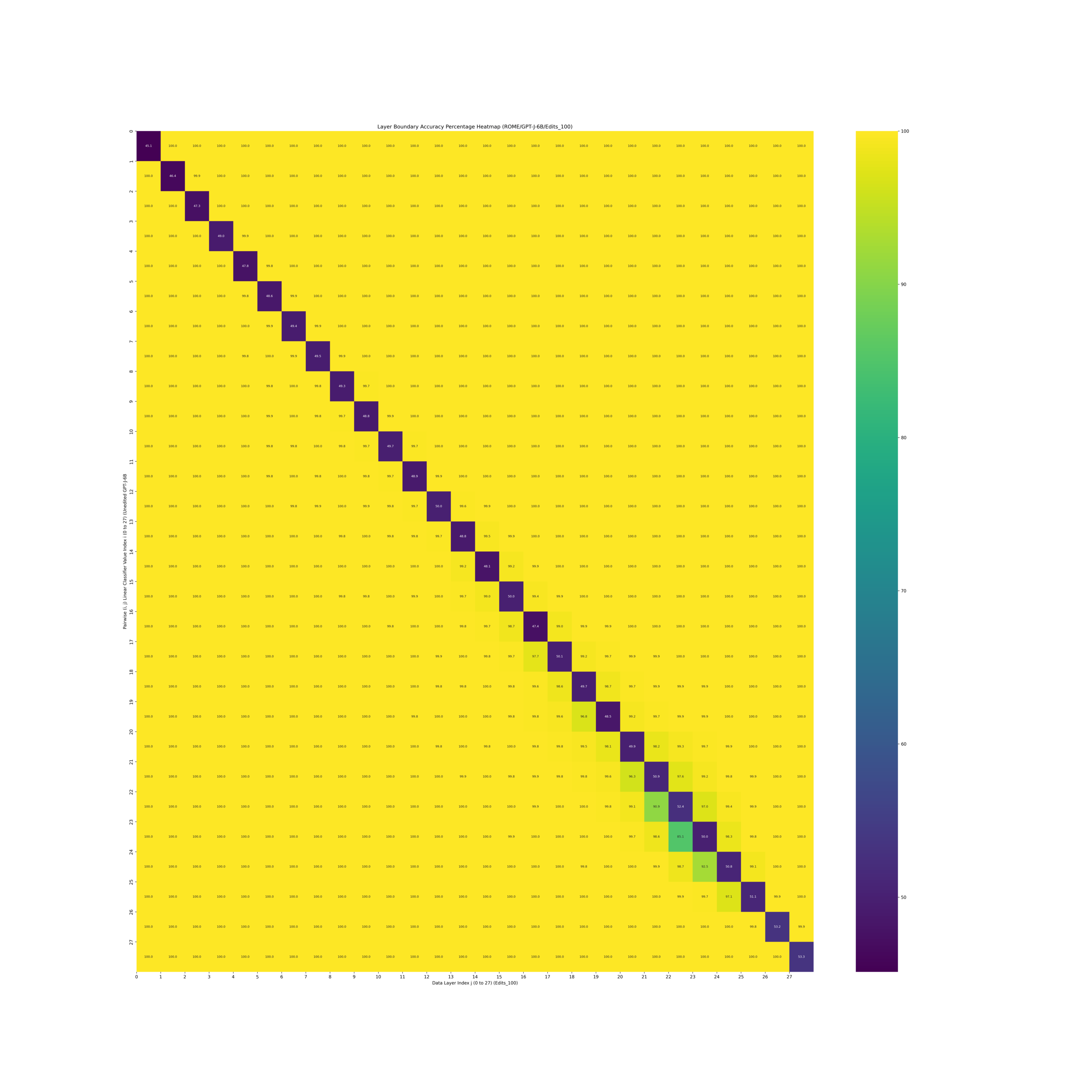}
        \caption{100}
    \end{subfigure}
    \begin{subfigure}{0.32\textwidth}
        \includegraphics[width=\linewidth]{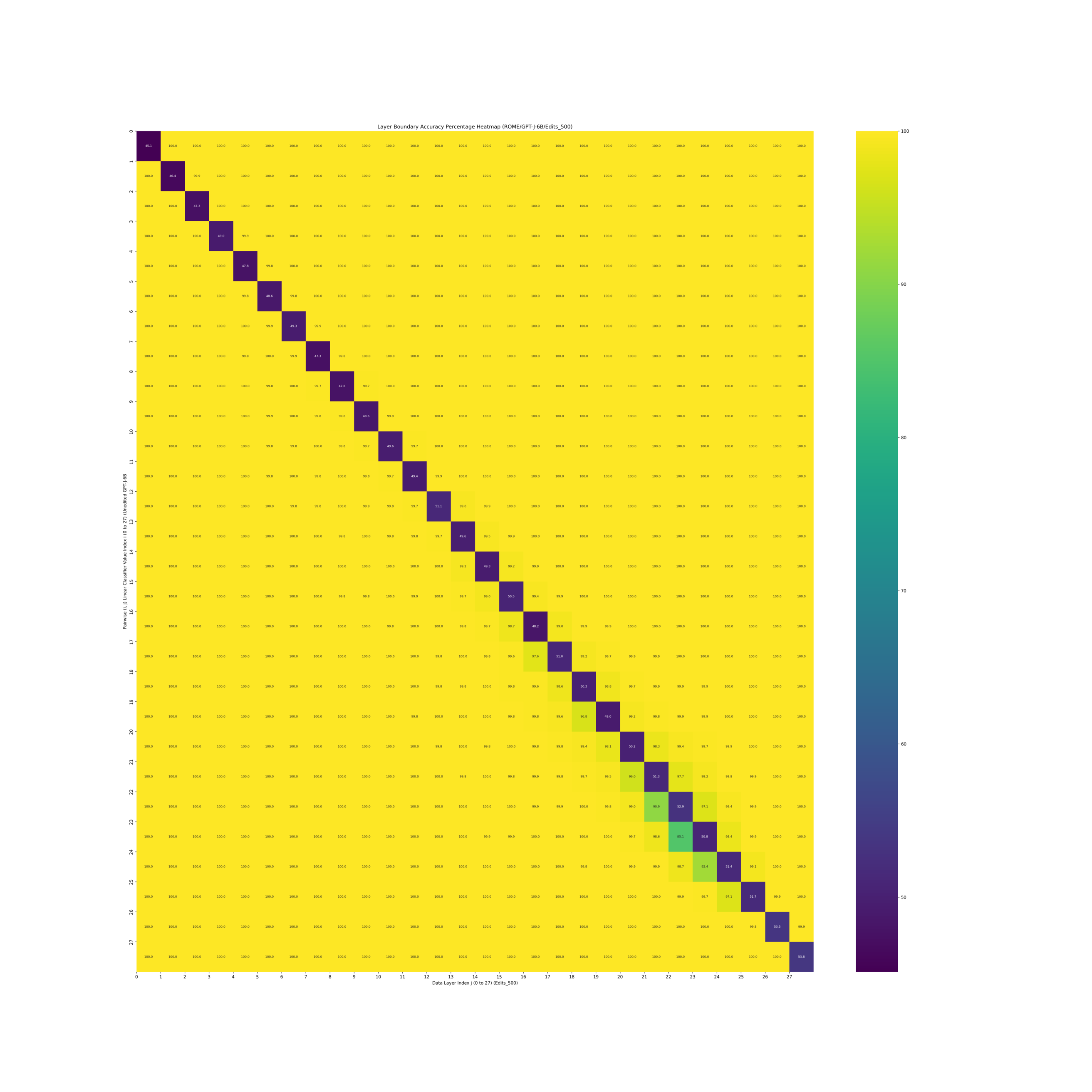}
        \caption{500}
    \end{subfigure}
    \begin{subfigure}{0.32\textwidth}
        \includegraphics[width=\linewidth]{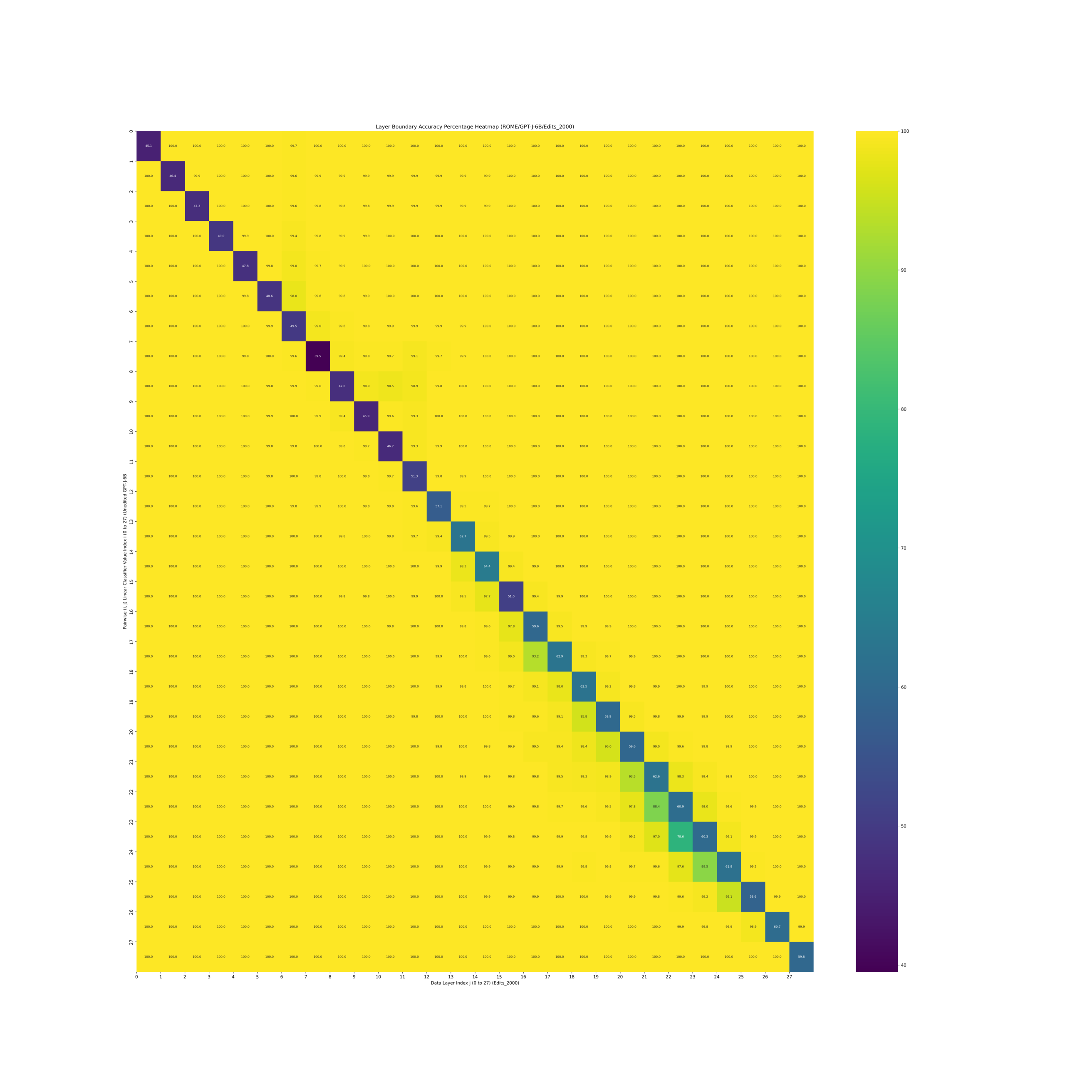}
        \caption{2000}
    \end{subfigure}
       
    \vspace{0.5cm} 

    \caption{Boundary plots for edits 100, 500, 2000 for GPT-J/ROME.}
    \label{fig:boundary-GPTJ-ROME}
\end{figure*}

\begin{figure*}[h]
    \centering
    \begin{subfigure}{0.32\textwidth}
        \includegraphics[width=\linewidth]{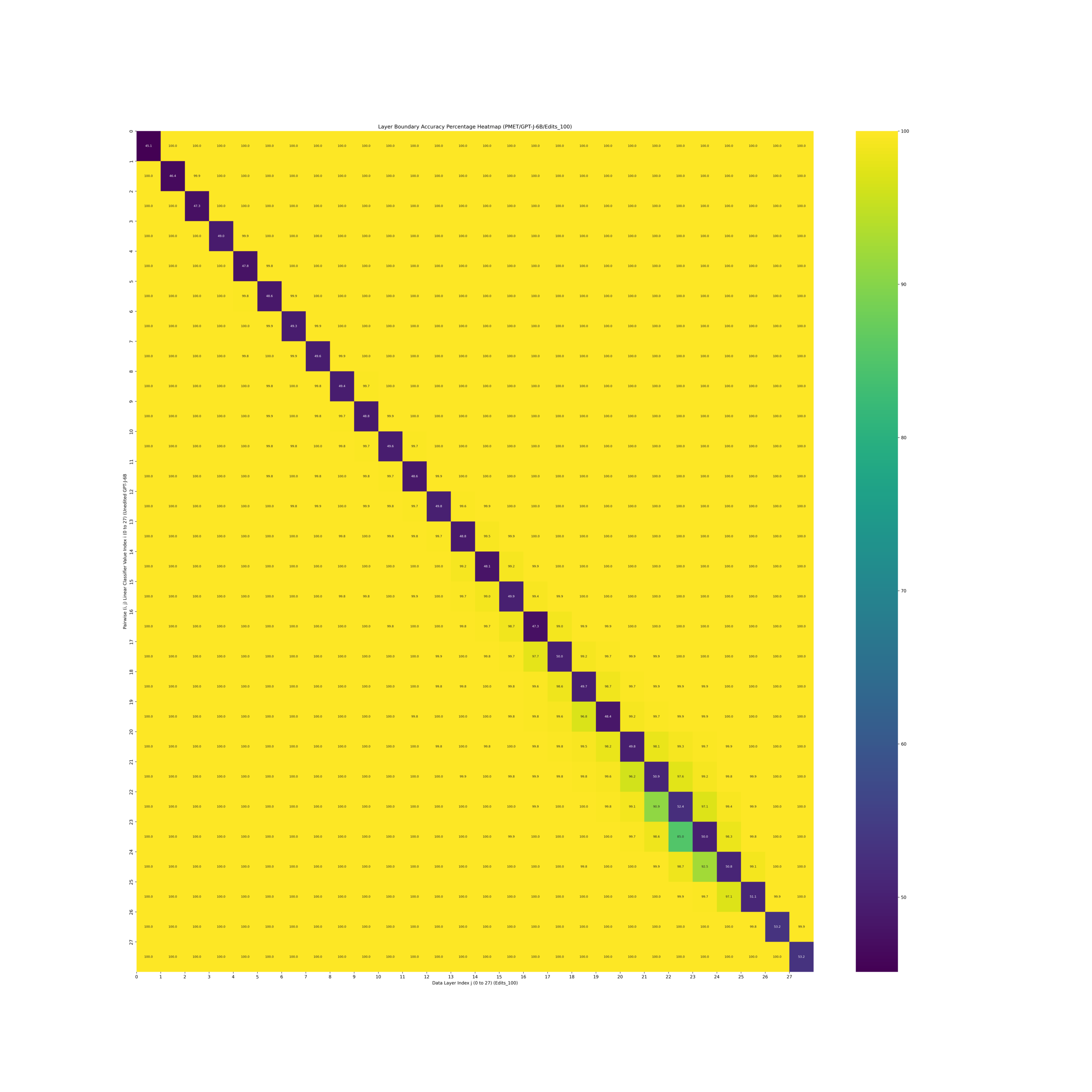}
        \caption{100}
    \end{subfigure}
    \begin{subfigure}{0.32\textwidth}
        \includegraphics[width=\linewidth]{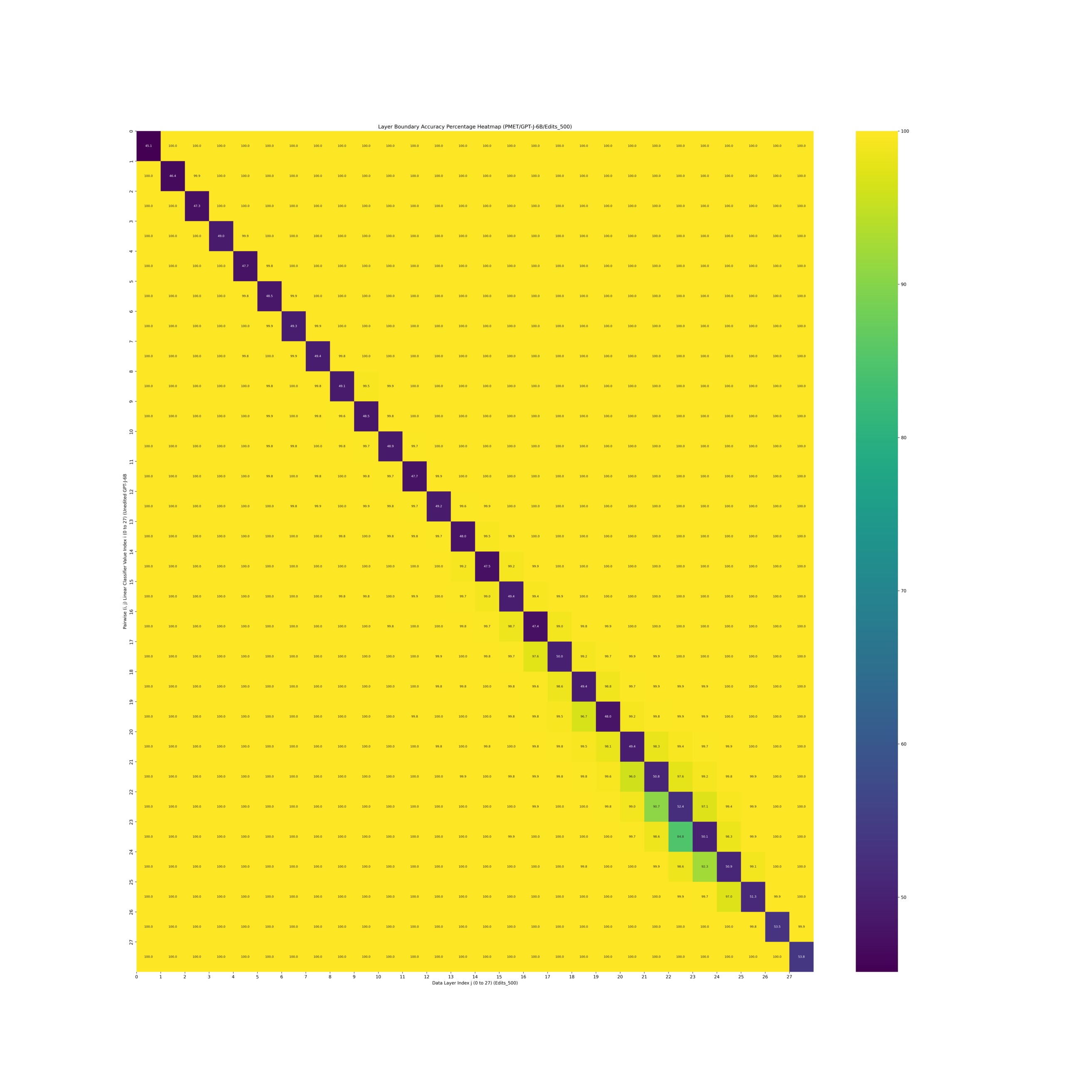}
        \caption{500}
    \end{subfigure}
    \begin{subfigure}{0.32\textwidth}
        \includegraphics[width=\linewidth]{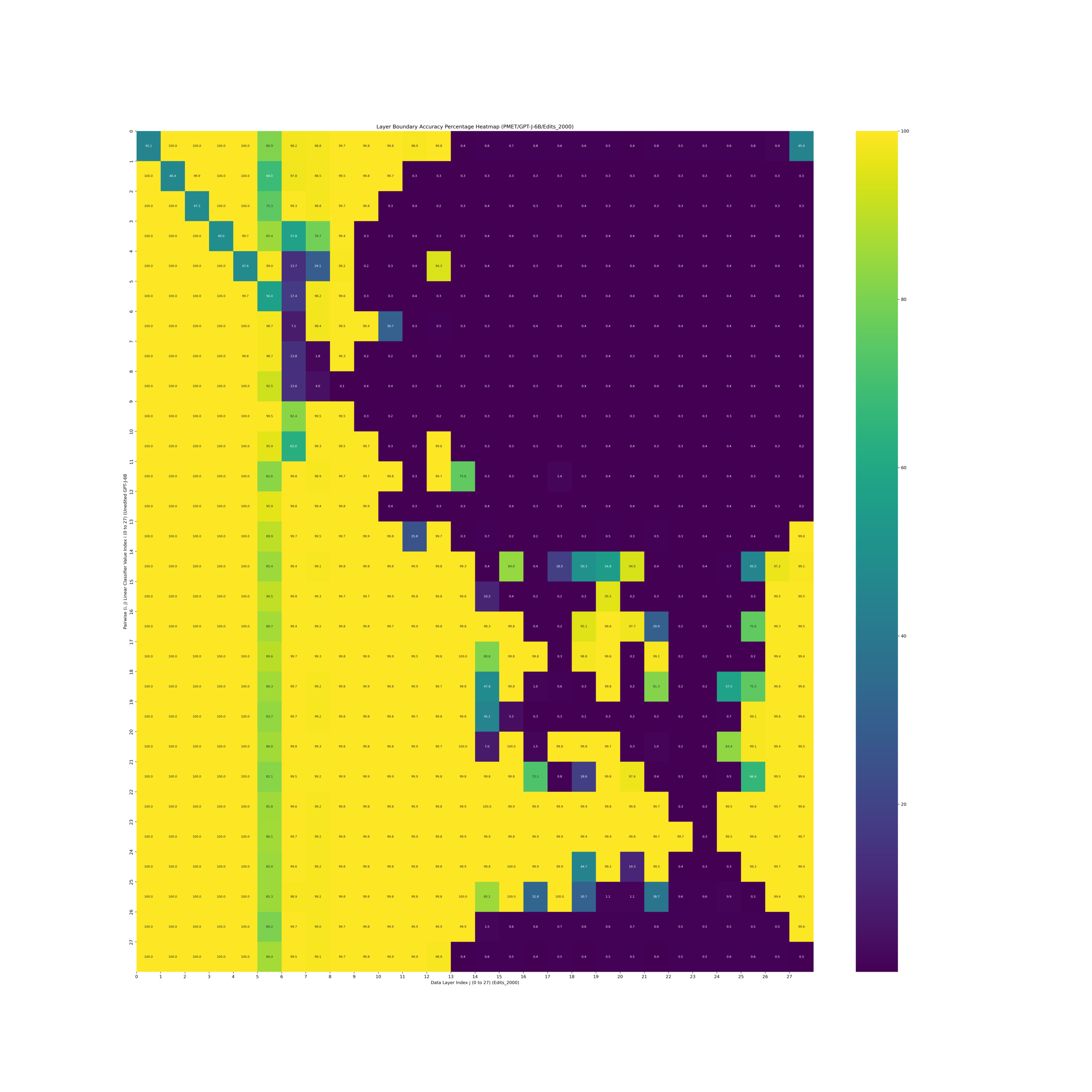}
        \caption{2000}
    \end{subfigure}
       
    \vspace{0.5cm} 

    \caption{Boundary plots for edits 100, 500, 2000 for GPT-J/PMET.}
    \label{fig:boundary-GPTJ-PMET}
\end{figure*}

\begin{figure*}[h]
    \centering
    \begin{subfigure}{0.32\textwidth}
        \includegraphics[width=\linewidth]{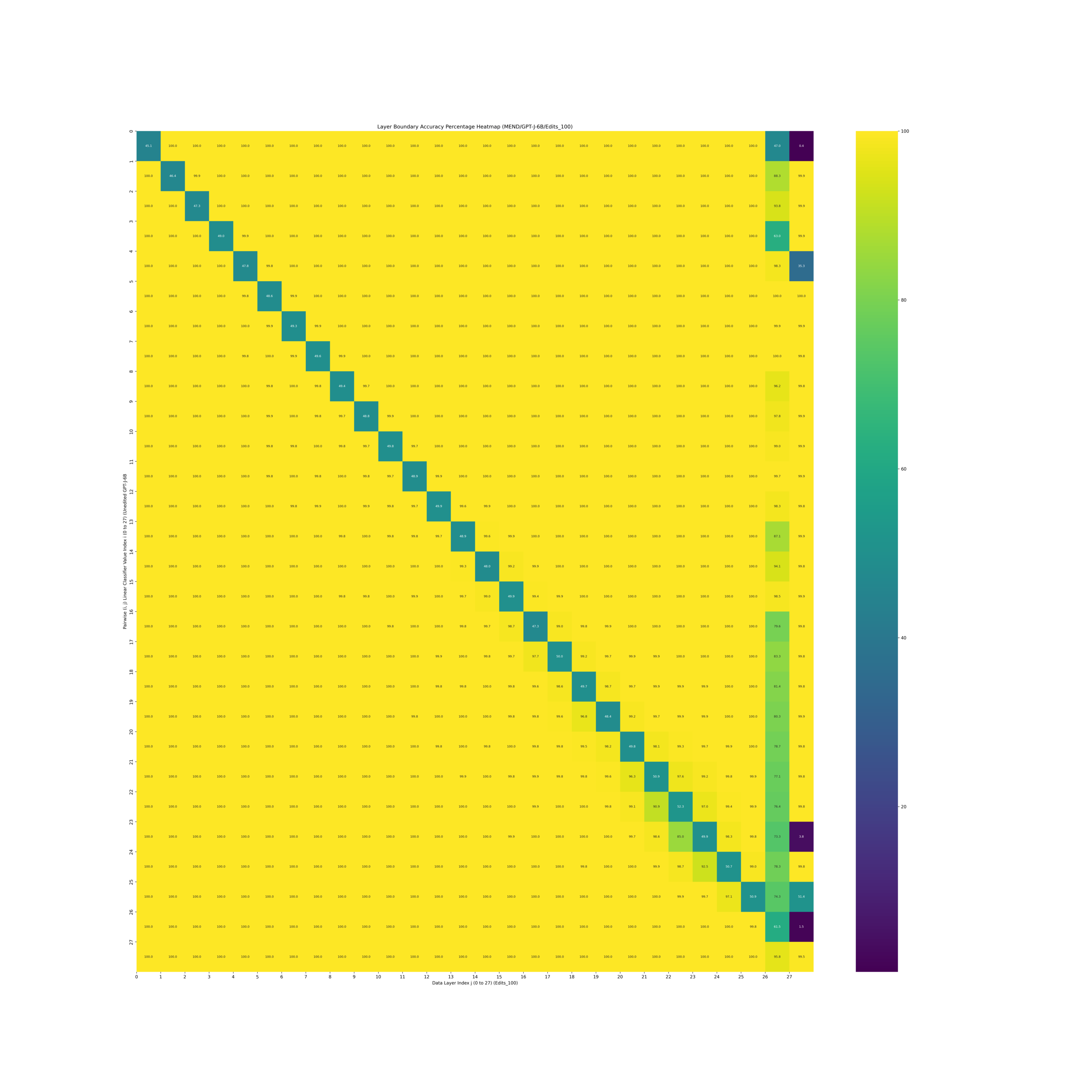}
        \caption{100}
    \end{subfigure}
    \begin{subfigure}{0.32\textwidth}
        \includegraphics[width=\linewidth]{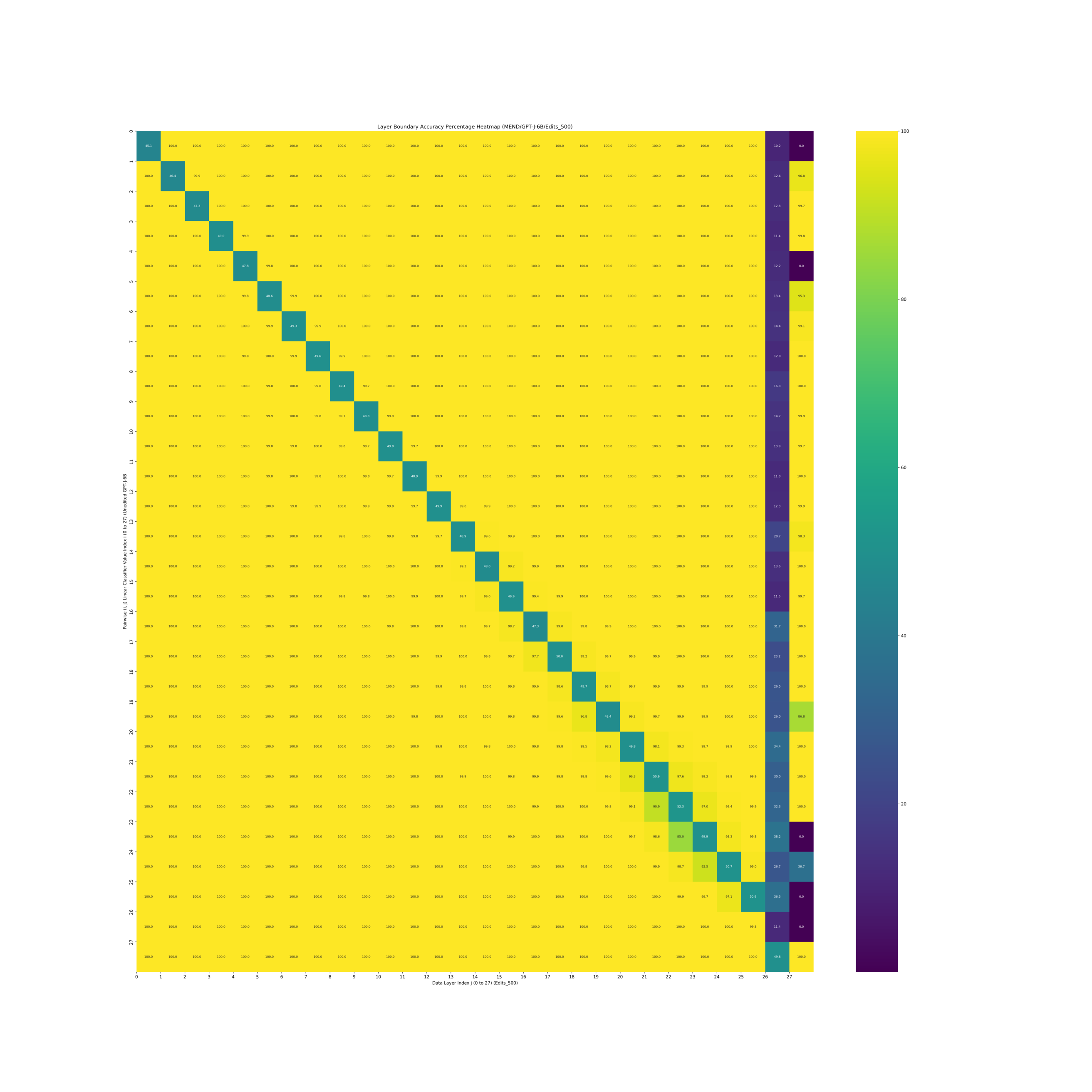}
        \caption{500}
    \end{subfigure}
    \begin{subfigure}{0.32\textwidth}
        \includegraphics[width=\linewidth]{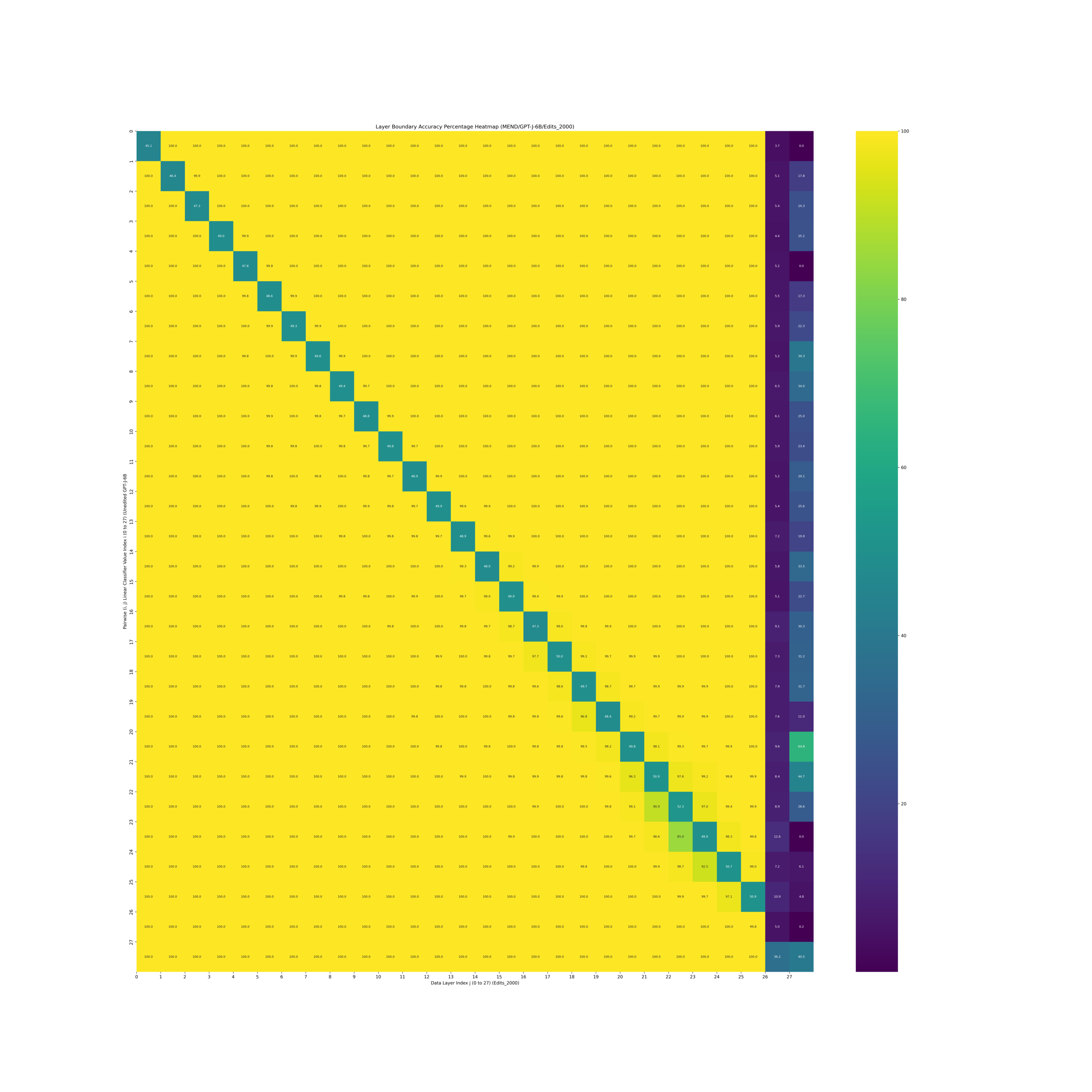}
        \caption{2000}
    \end{subfigure}
       
    \vspace{0.5cm} 

    \caption{Boundary plots for edits 100, 500, 2000 for GPT-J/MEND.}
    \label{fig:boundary-GPTJ-MEND}
\end{figure*}

\begin{figure*}[h]
    \centering
    \begin{subfigure}{0.32\textwidth}
        \includegraphics[width=\linewidth]{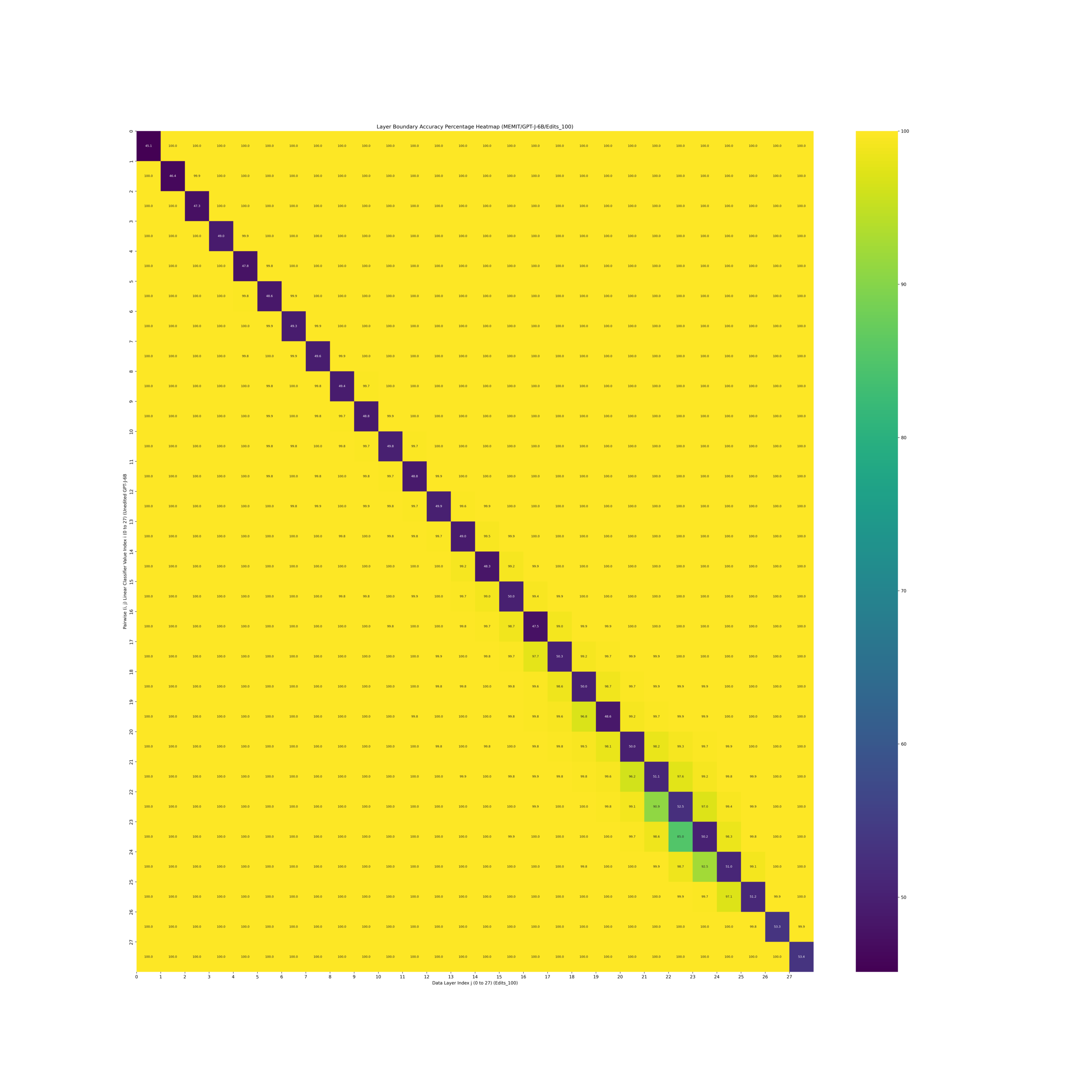}
        \caption{100}
    \end{subfigure}
    \begin{subfigure}{0.32\textwidth}
        \includegraphics[width=\linewidth]{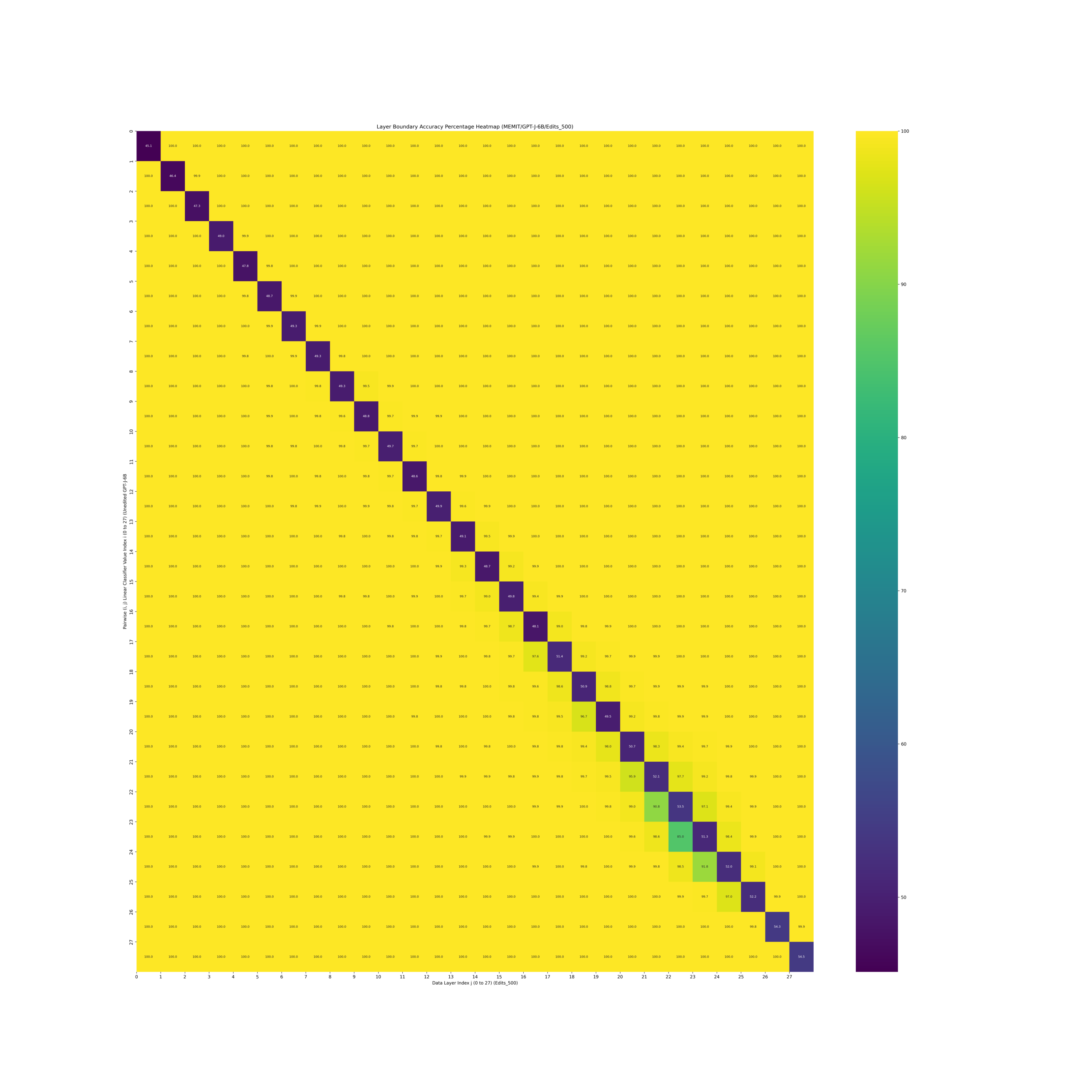}
        \caption{500}
    \end{subfigure}
    \begin{subfigure}{0.32\textwidth}
        \includegraphics[width=\linewidth]{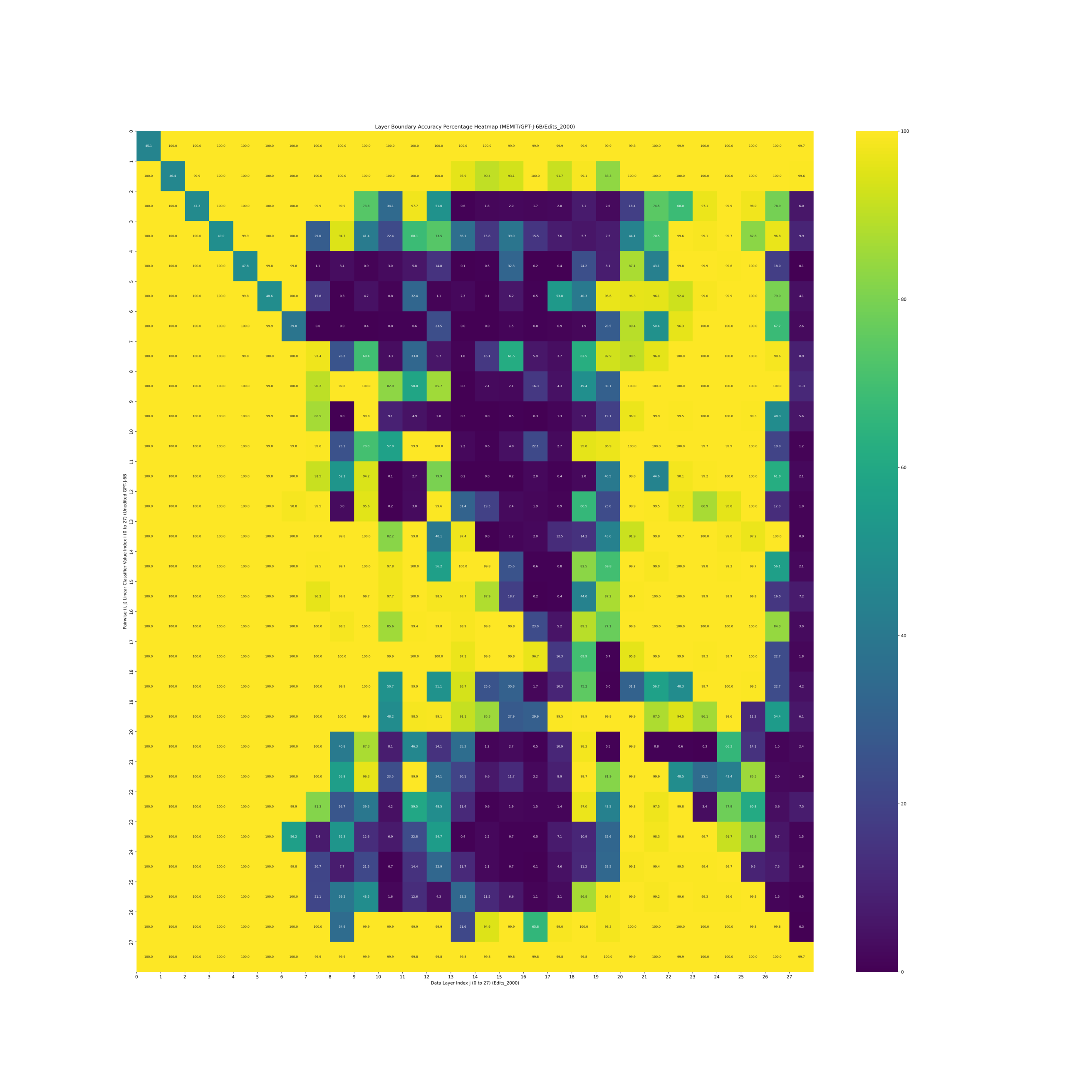}
        \caption{2000}
    \end{subfigure}
       
    \vspace{0.5cm} 

    \caption{Boundary plots for edits 100, 500, 2000 for GPT-J/MEMIT.}
    \label{fig:boundary-GPTJ-MEMIT}
\end{figure*}

\begin{figure*}[h]
    \centering
    \begin{subfigure}{0.32\textwidth}
        \includegraphics[width=\linewidth]{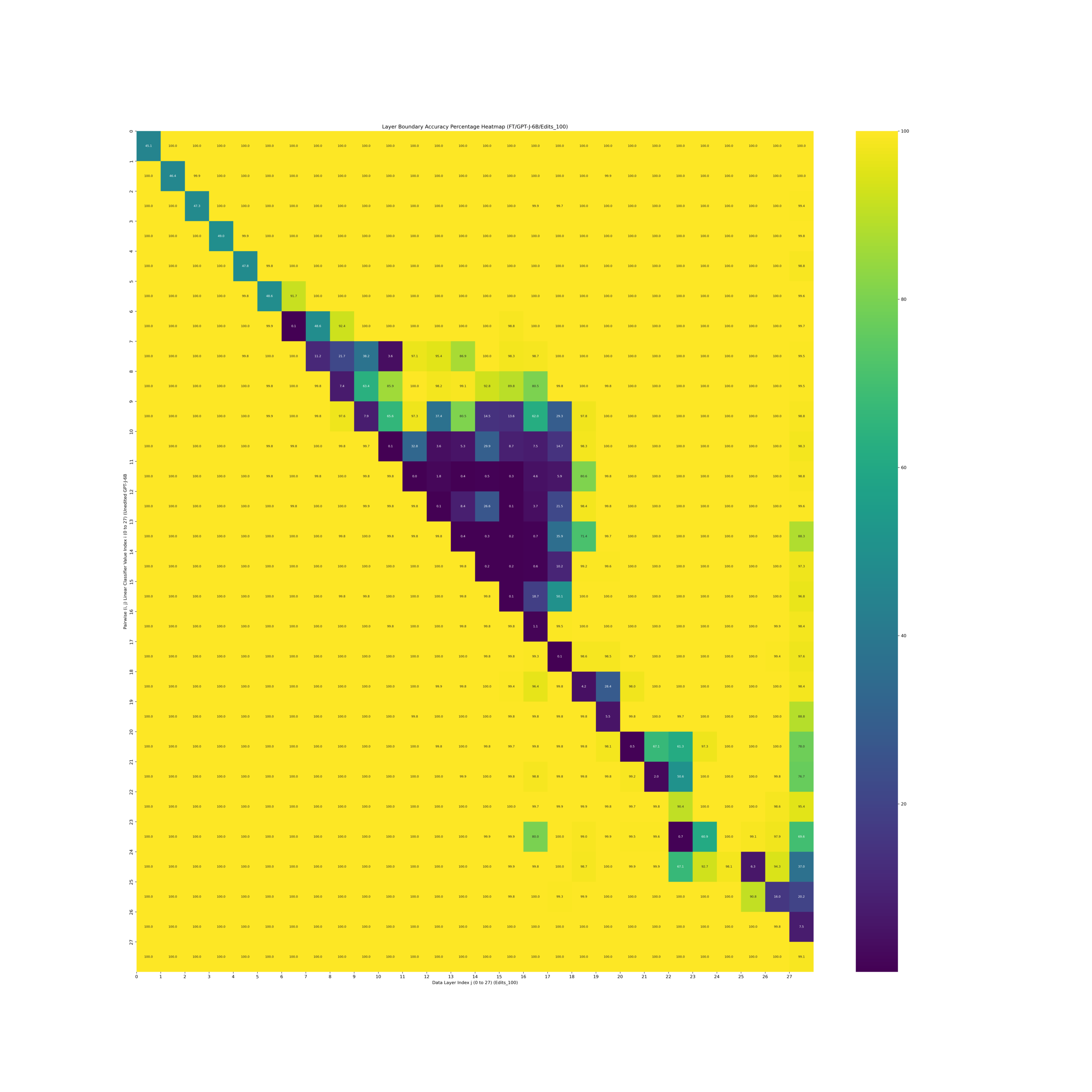}
        \caption{100}
    \end{subfigure}
    \begin{subfigure}{0.32\textwidth}
        \includegraphics[width=\linewidth]{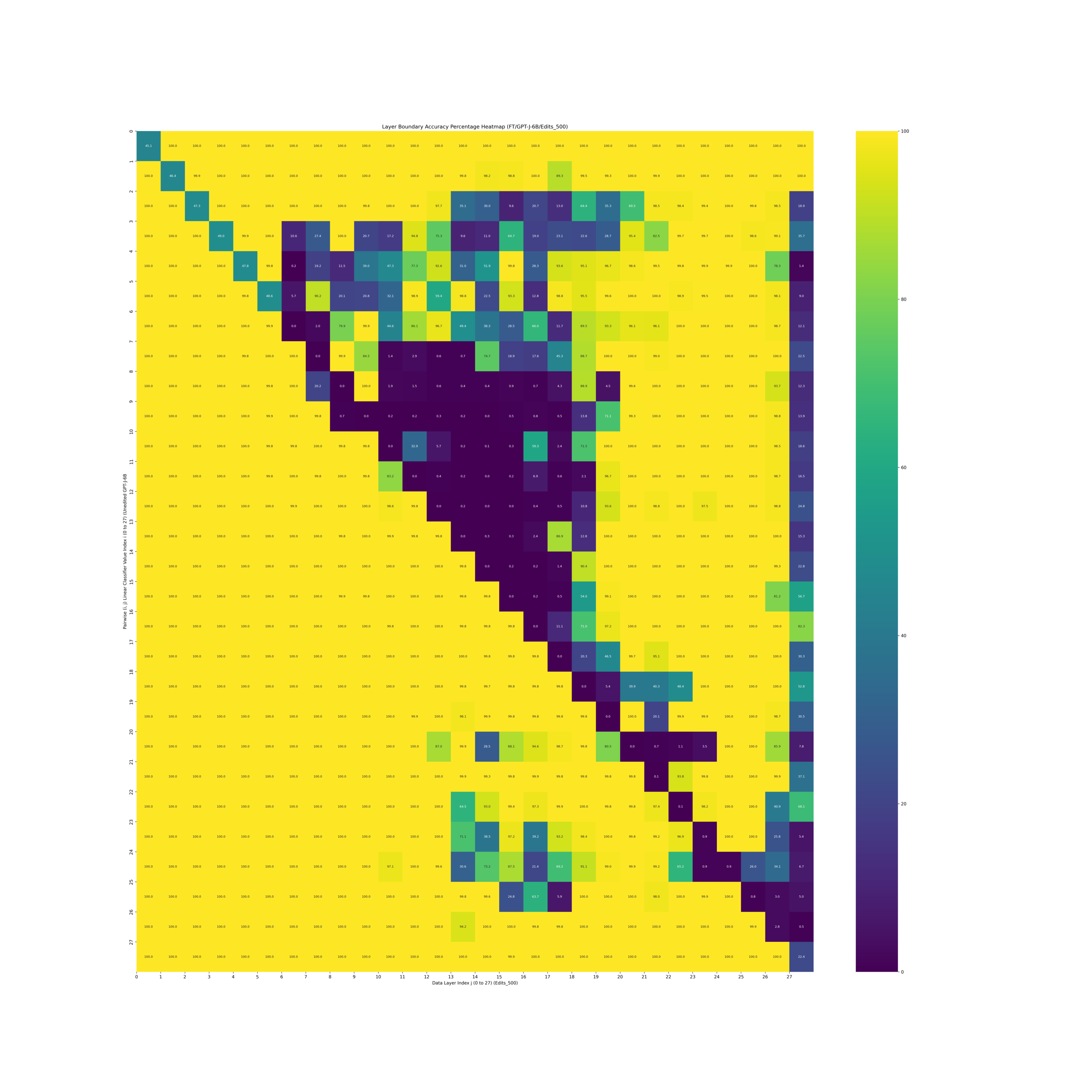}
        \caption{500}
    \end{subfigure}
    \begin{subfigure}{0.32\textwidth}
        \includegraphics[width=\linewidth]{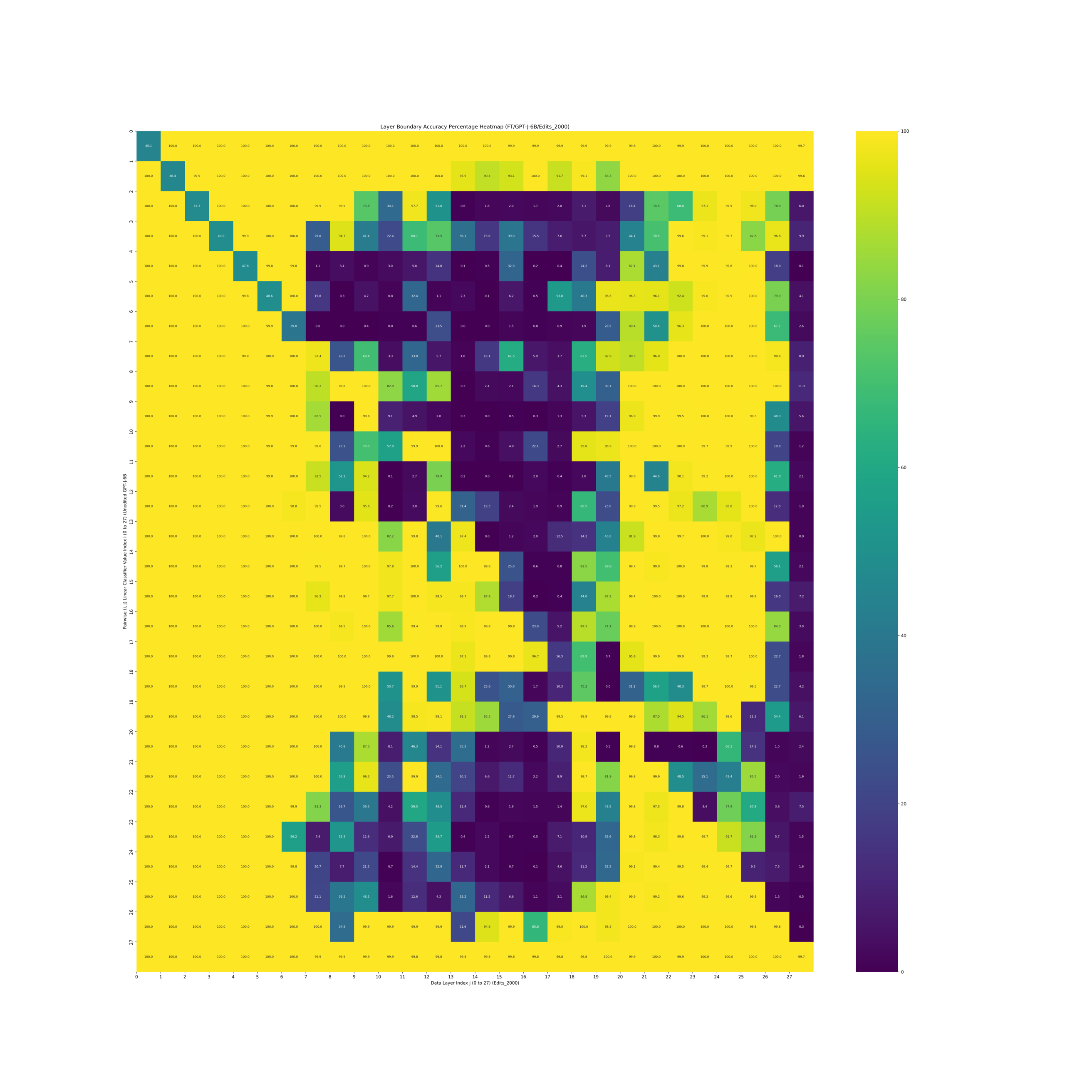}
        \caption{2000}
    \end{subfigure}
       
    \vspace{0.5cm} 

    \caption{Boundary plots for edits 100, 500, 2000 for GPT-J/FT.}
    \label{fig:boundary-GPTJ-FT}
\end{figure*}

\begin{figure*}[h]
    \centering
    \begin{subfigure}{0.22\textwidth}
        \includegraphics[width=\linewidth]{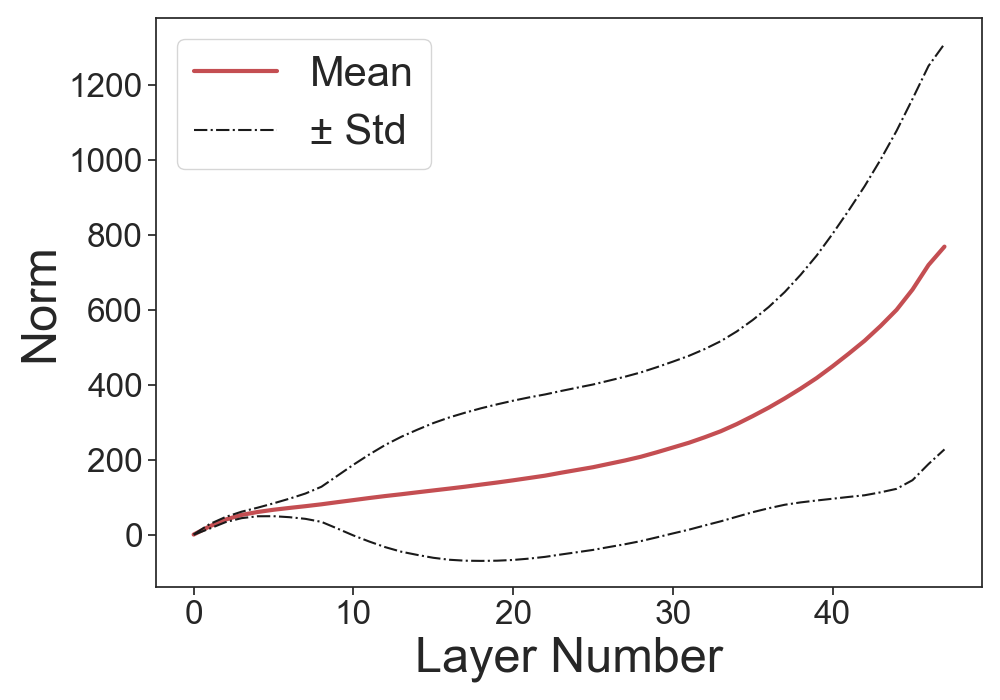}
        \caption{MEMIT/100}
    \end{subfigure}
    \begin{subfigure}{0.22\textwidth}
        \includegraphics[width=\linewidth]{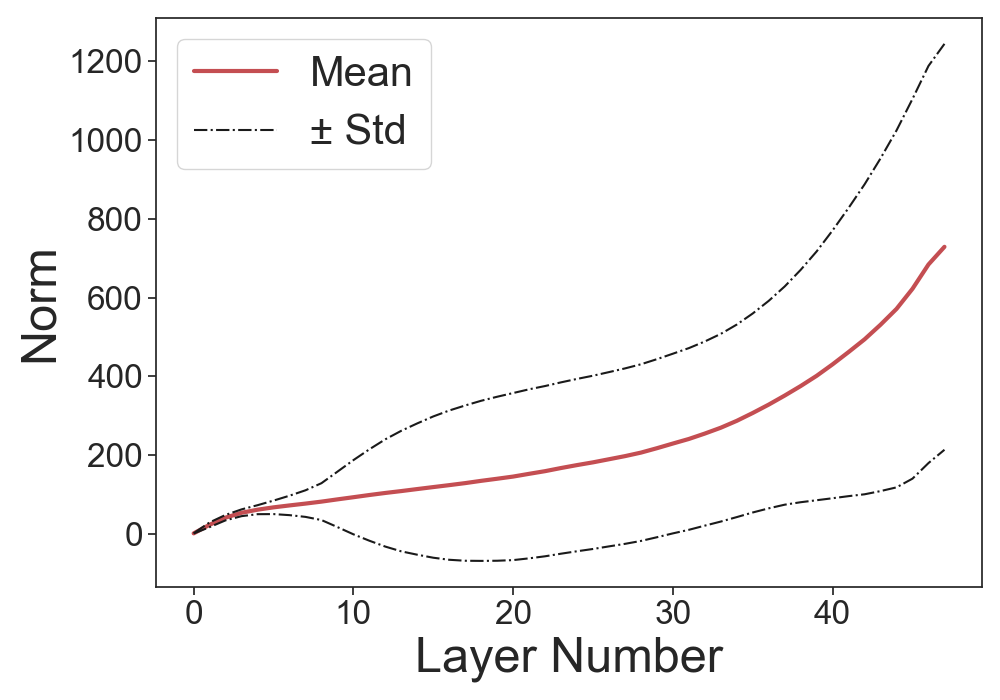}
        \caption{MEMIT/500}
    \end{subfigure}
    \begin{subfigure}{0.22\textwidth}
        \includegraphics[width=\linewidth]{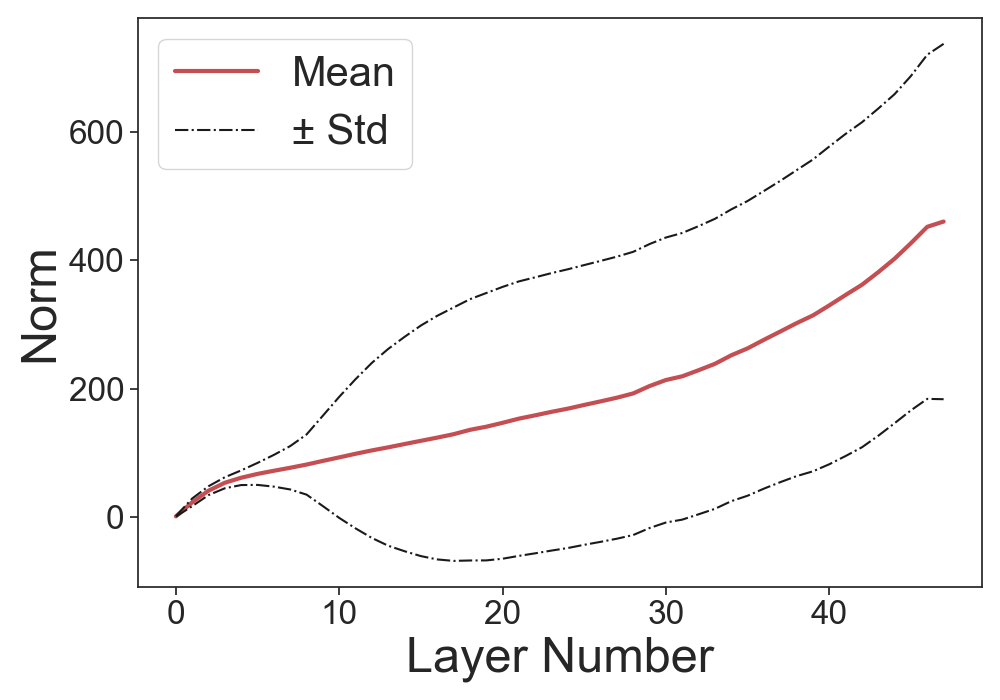}
        \caption{MEMIT/2000}
    \end{subfigure}

    \vspace{0.5cm}

    \begin{subfigure}{0.22\textwidth}
        \includegraphics[width=\linewidth]{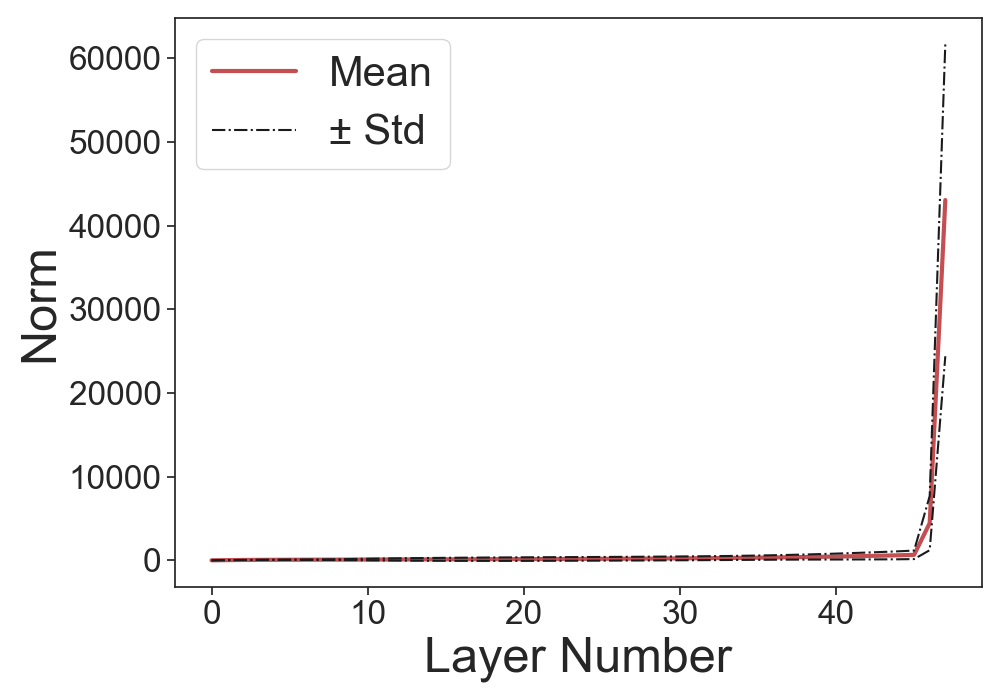}
        \caption{MEND/100}
    \end{subfigure}
    \begin{subfigure}{0.22\textwidth}
        \includegraphics[width=\linewidth]{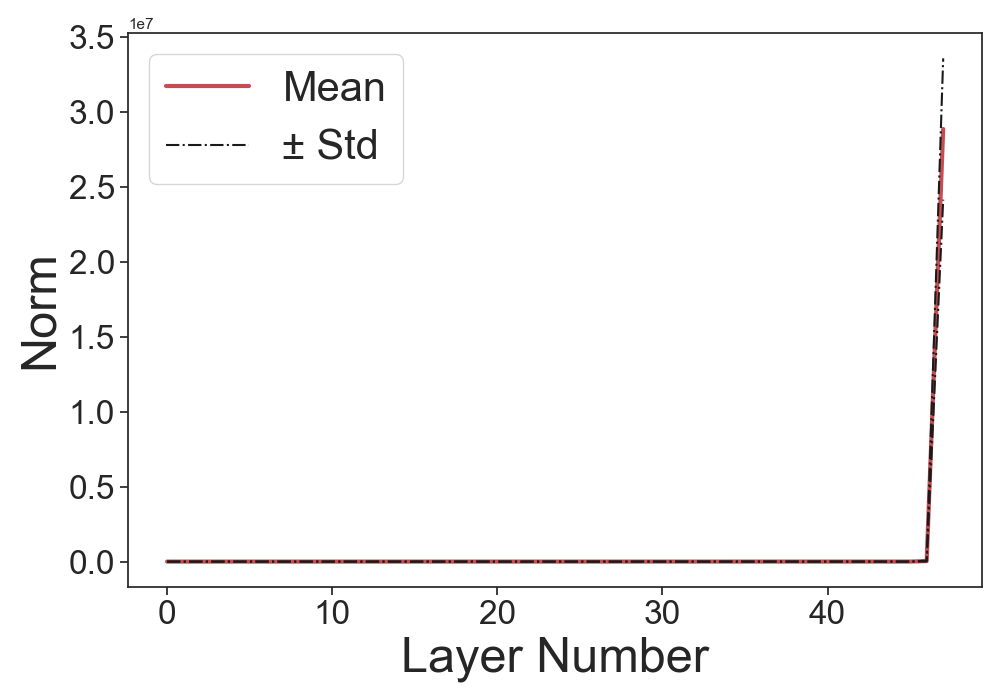}
        \caption{MEND/500}
    \end{subfigure}
    \begin{subfigure}{0.22\textwidth}
        \includegraphics[width=\linewidth]{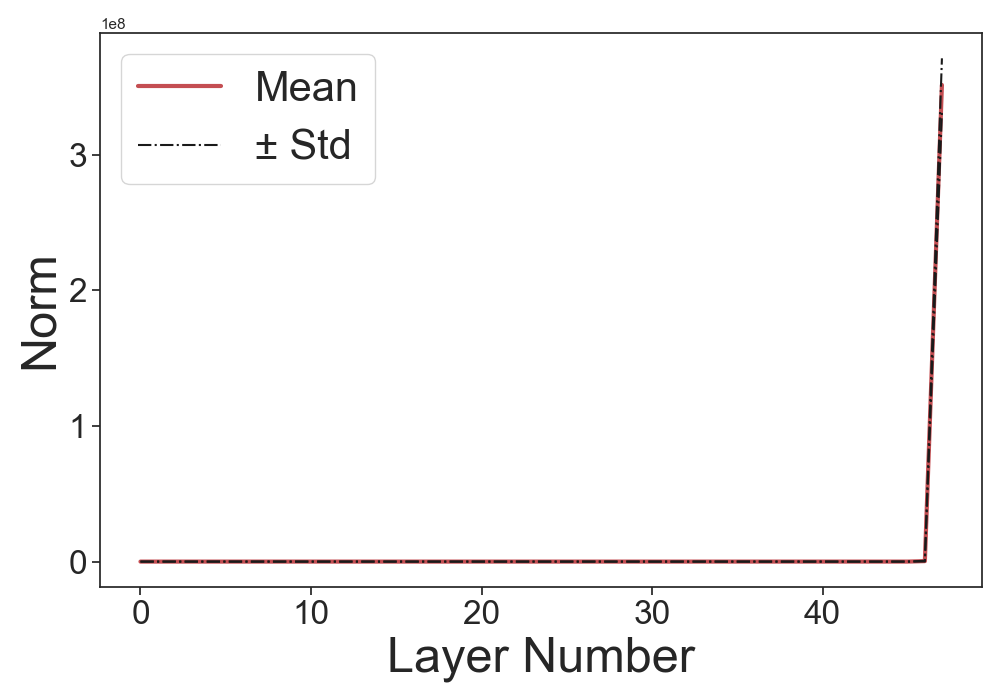}
        \caption{MEND/2000}
    \end{subfigure}

    \vspace{0.5cm}

    \begin{subfigure}{0.22\textwidth}
        \includegraphics[width=\linewidth]{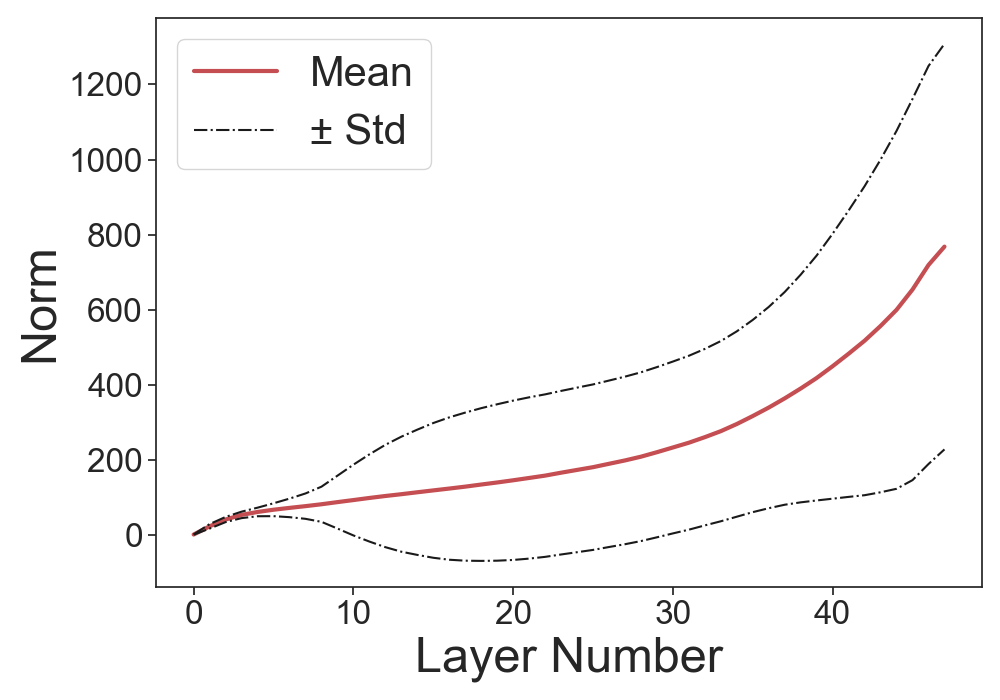}
        \caption{PMET/100}
    \end{subfigure}
    \begin{subfigure}{0.22\textwidth}
        \includegraphics[width=\linewidth]{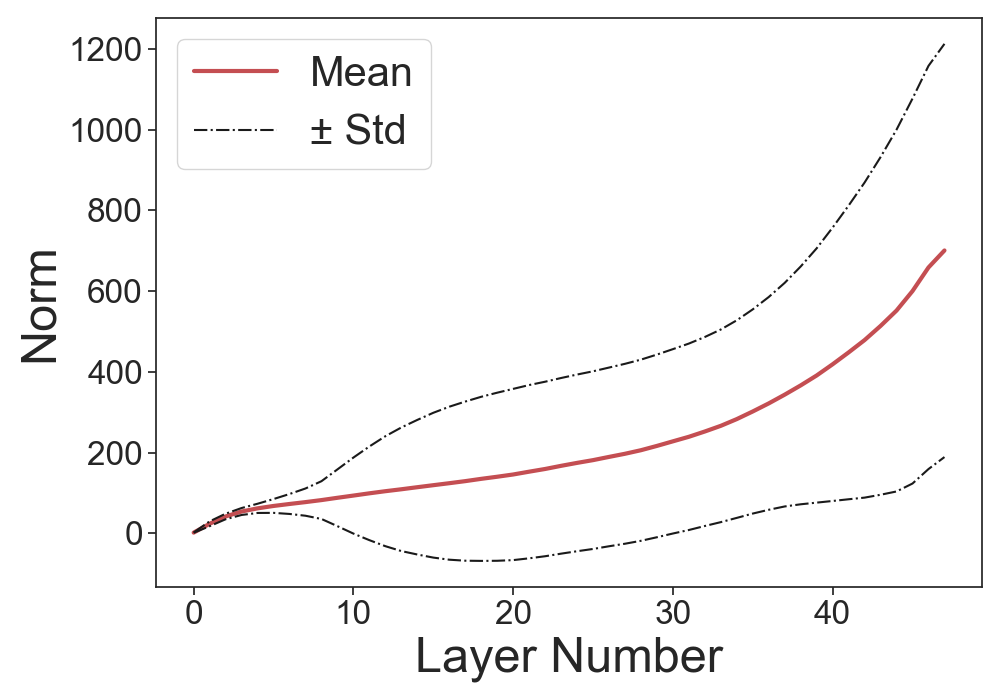}
        \caption{PMET/500}
    \end{subfigure}
    \begin{subfigure}{0.22\textwidth}
        \includegraphics[width=\linewidth]{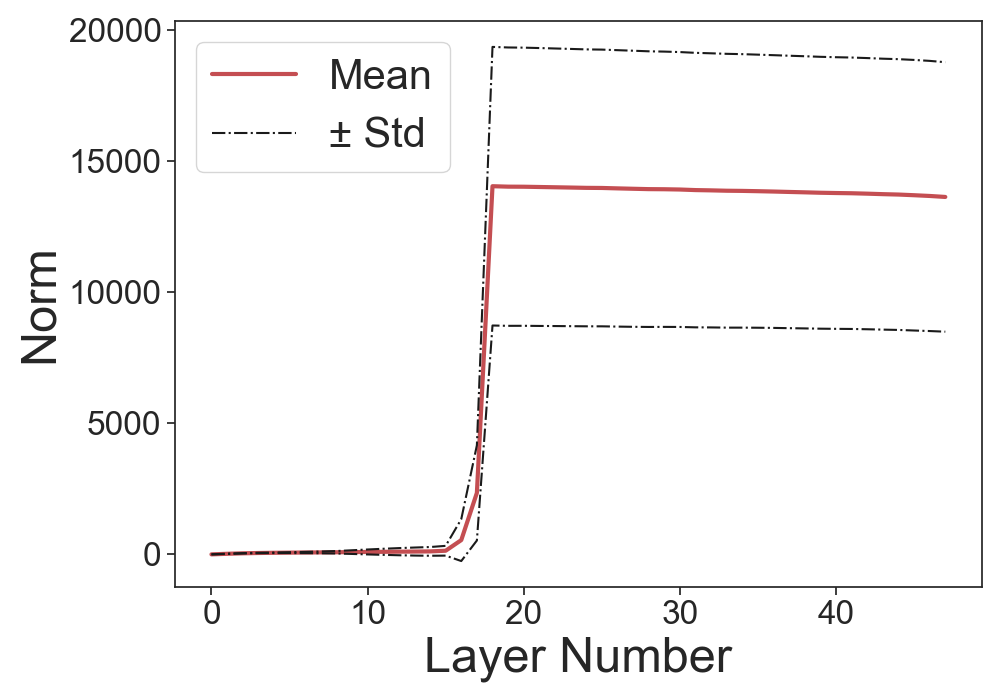}
        \caption{PMET/2000}
    \end{subfigure}

    \vspace{0.5cm}

    \begin{subfigure}{0.22\textwidth}
        \includegraphics[width=\linewidth]{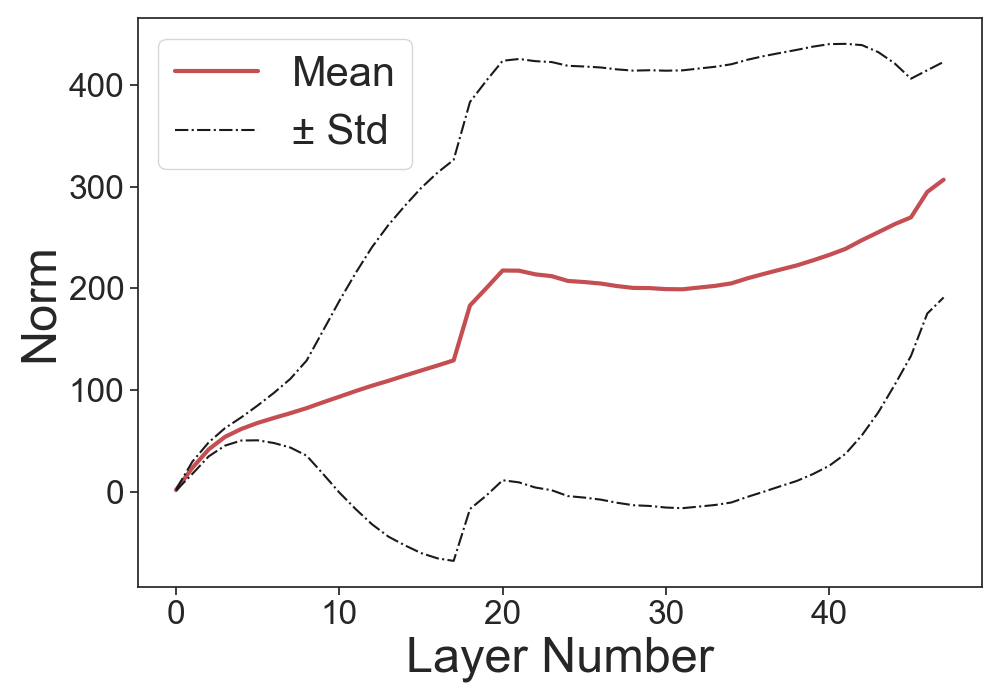}
        \caption{FT/100}
    \end{subfigure}
    \begin{subfigure}{0.22\textwidth}
        \includegraphics[width=\linewidth]{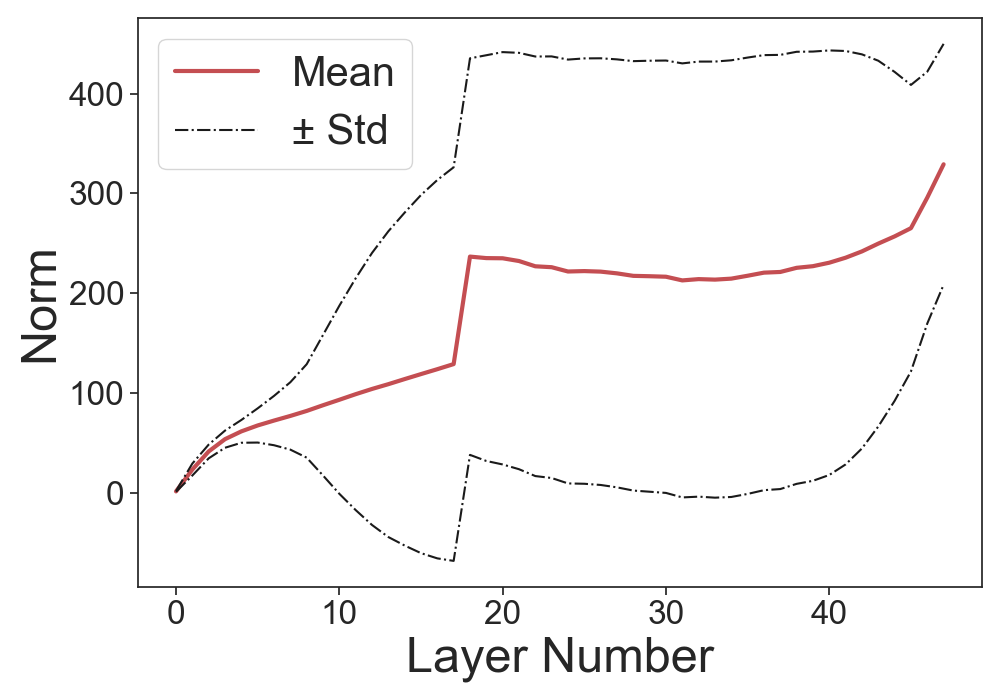}
        \caption{FT/500}
    \end{subfigure}
    \begin{subfigure}{0.22\textwidth}
        \includegraphics[width=\linewidth]{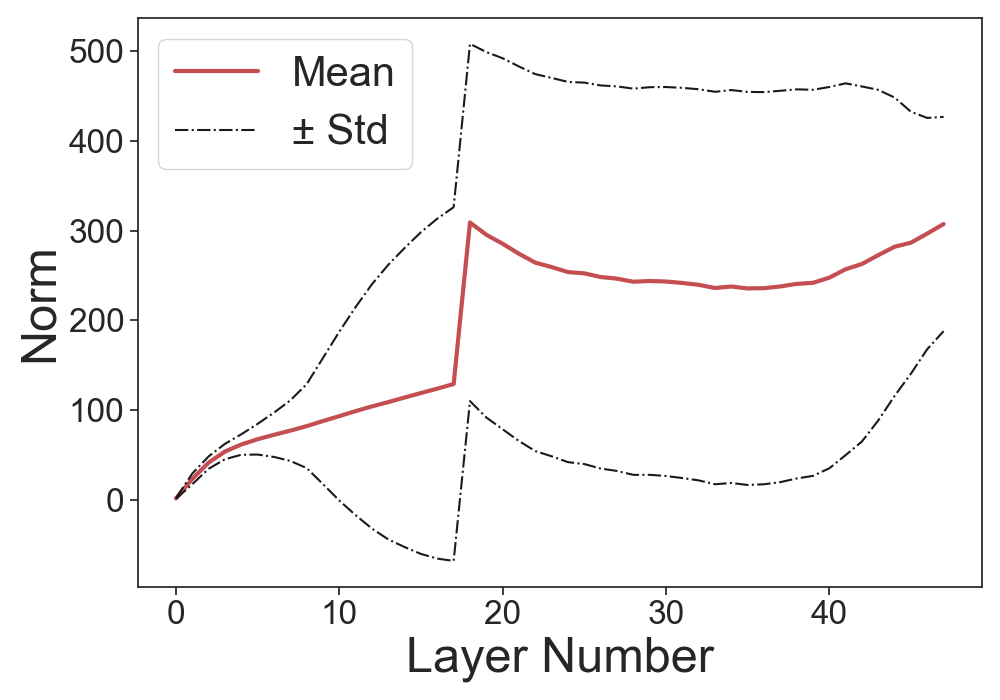}
        \caption{FT/2000}
    \end{subfigure}
    
    \caption{Activation Norms at different layers for edits 100, 500, 2000 for all editing methods for GPT2-XL.}
    \label{fig:activation-norm-growth-APPENDIX}
\end{figure*}

\end{document}